# Autonomous Driving with Deep Learning: A Survey of State-of-Art Technologies

Yu Huang
Autonomous Driving Research
Black Sesame Technology Inc.
Santa Clara, USA
yu.huang@bst.ai

Yue Chen
Corporate Technology Strategy
Futurewei Technology Inc.
Santa Clara, USA
yue.chen@futurewei.com

***Abstract*— Since DARPA's Grand Challenges (rural) in 2004/05 and Urban Challenges in 2007, autonomous driving has been the most active field of AI applications. Almost at the same time, deep learning has made breakthrough by several pioneers, three of them (also called fathers of deep learning), Hinton, Bengio and LeCun, won ACM Turin Award in 2019. This is a survey of autonomous driving technologies with deep learning methods. We investigate the major fields of self-driving systems, such as perception, mapping and localization, prediction, planning and control, simulation, V2X and safety etc. Due to the limited space, we focus the analysis on several key areas, i.e. 2D/3D object detection in perception, depth estimation from cameras, multiple sensor fusion on the data, feature and task level respectively, behaviour modelling and prediction of vehicle driving and pedestrian trajectories.**

## I. Introduction

Autonomous Driving has been active for more than 10 years [48, 51, 57, 58]. In 2004 and 2005, DARPA held the Grand Challenges in rural driving of driverless vehicles. In 2007, DAPRA also held the Urban Challenges for autonomous driving in street environments. Then professor S. Thrun at Stanford university, the first-place winner in 2005 and the second-place winner in 2007, joined Google and built Google X and the self-driving team.

Breakthroughs on deep learning have been achieved since Hinton published new deep structured learning architecture, called deep belief network (DBN) [5]. In 2019, the fathers of deep learning, Hinton, Bengio and LeCun, were nominated as the winners of ACM Turin Award. The past decade has seen rapid developments of deep learning techniques with significant impacts on signal and information processing. In the ImageNet Challenge 2002, the first place winner came from Hinton's group, using a novel Convolutional Neural Network (CNN) called AlexNet [6].

In this paper, we investigate how autonomous driving marries deep learning. Recently there are two survey papers in this area [1, 2]. Our survey work spans the state-of-art technology in major fields of self-driving technologies, such as perception, mapping and localization, prediction, planning and control, simulation, V2X and safety etc. Due to the limited space, we focus on some critical areas, i.e. 2D/3D object detection in perception based on different sensors (cameras, radar and LiDAR), depth estimation from cameras, sensor fusion in data, feature and task level respectively, behavior modeling and prediction for vehicle and pedestrian trajectories.

## II. Overview of Deep Learning

There are not a few overview papers [13, 21-40], even a well-organized technical book [14] about deep learning state-of-art theories and applications. In this section, we briefly review basic theories of deep learning, shown in Fig. 1, well as related issues, like distributed computing, model compression and acceleration.

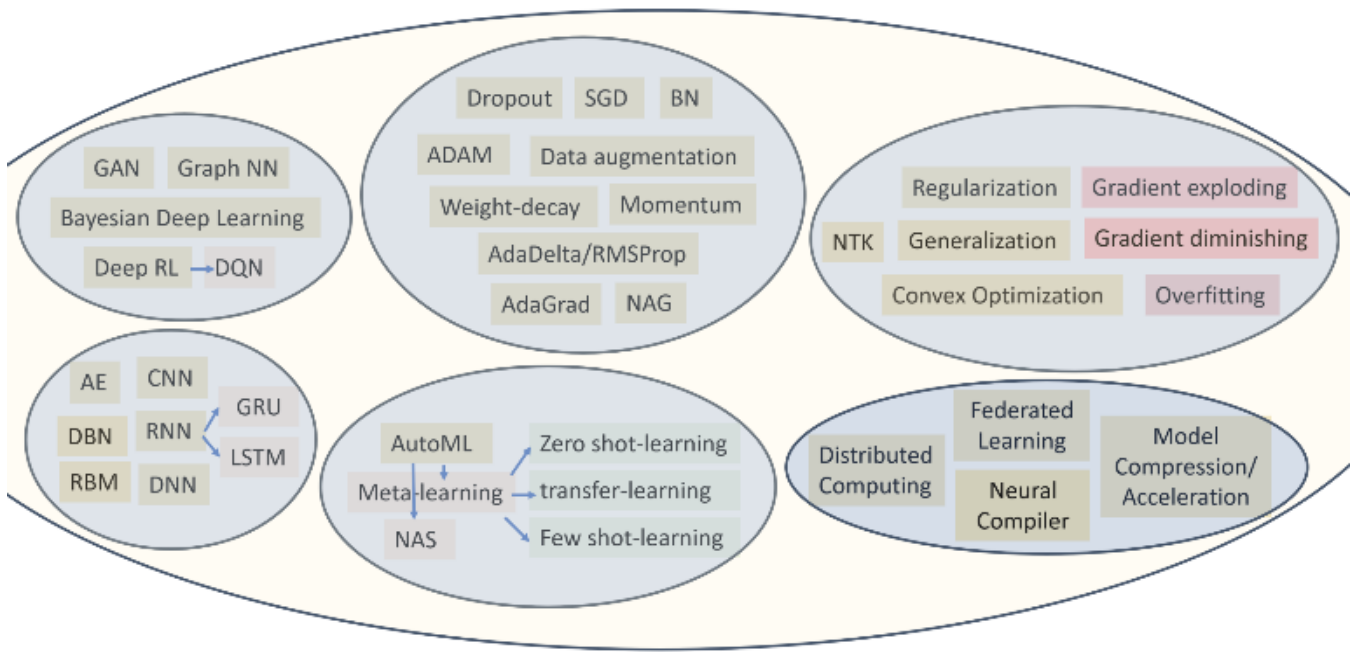

Fig. 1. Deep learning kingdom.

### A. Basic Theory

In contrary to the deep learning success era, the previous methods in machine learning are called *shallow learning*. Like machine learning, deep learning also follows the category as unsupervised, semi-supervised, supervised and reinforcement learning (RL) [23, 28].

In *supervised learning* domain, there are different deep leaning methods, including Deep Neural Networks (DNN), Convolutional Neural Networks (CNN), Recurrent Neural Networks (RNN) as Long Short-Term Memory (LSTM) [4] and Gated Recurrent Units (GRU) [12].

In *unsupervised learning* domain, there are several members for clustering and non-linear dimensionality reduction, including

Auto Encoders (AE), Deep Restricted Boltzmann Machines (RBM), and GAN (Generative Adversarial Networks). In addition, RNNs, such as LSTM and Deep RL, are also used for unsupervised learning in many application domains.

In *semi-supervised learning* domain, Deep RL and GAN are used as semi-supervised learning techniques. Additionally, RNN including LSTM and GRU are used as well.

*Deep reinforcement learning* [26] is the combination of RL and deep learning. The famous Deep RL method applied in AlphaGo proposed by DeepMind is Deep Q-Network (DQN) with Q-learning. About Deep RL methods, we'd like to refer comments from Yann LeCun: "*If intelligence was a cake, unsupervised learning would be the cake, supervised learning would be the icing, and reinforcement learning would be the carry*."

Below we briefly introduce important milestones in the deep learning history [23, 28].

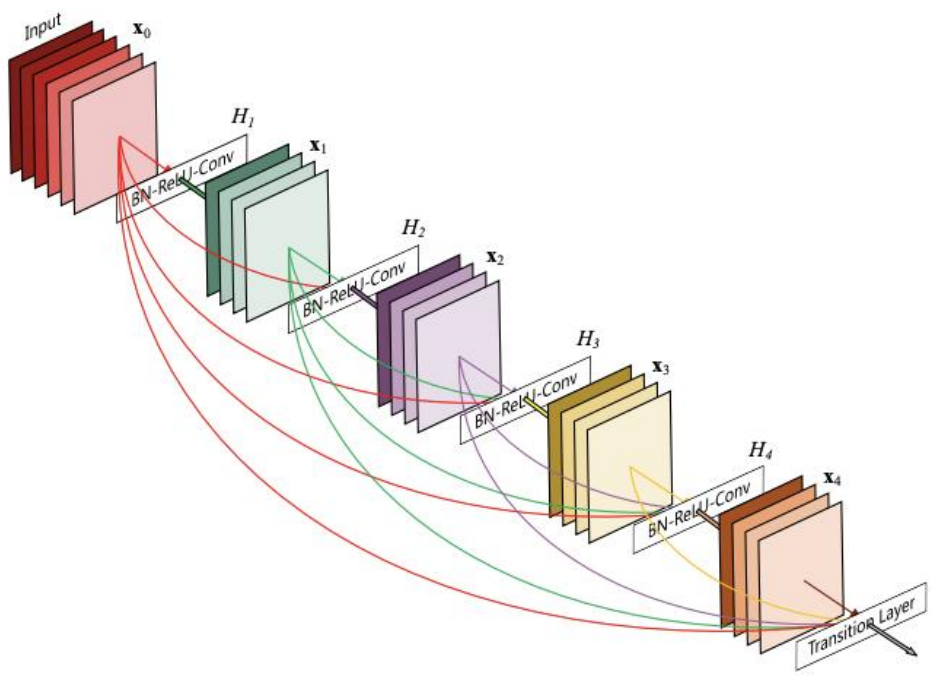

Fig. 2. DenseNet structure from reference [17].

There are some successful *CNNs* with convolution layers and pooling layers after the *DNNs* with the pure fully connected layers:1) the first CNN in image domain, the LeNet with Back propagation (BP) learning proposed by LeCun [3], 2) the first break though over traditional methods in image domain, the AlexNet with Local Response Normalization (LRN) and dropout, applied in ImageNet Challenge 2012 [6], 3) NiN (network in network) with multiple layer perception convolution [7], 4) VGG Net with ReLU (Rectified Linear Unit) activation function from Visual Geometry Group (VGG) at University of Oxford [8], 5) GoogleNet with Inception layers[9] and its modification with depth-wise separable convolutions [15], 6) ResNet (Residual Network) [16] from Microsoft Research Asia, later its several variations appeared, such as ResNeSt [20] with attention mechanism etc., 7) SE-Net (Squeeze and excitation) with global average pooling [18] from Momenta and its modification SK-Net (Selective Kernel) [19] with adaptive receptive field, 8) DenseNet with densely connected convolution layers [17] (shown in Fig. 2), and variations as PeleeNet [72] and VoVNet [73], and 9) EfficientNet [44] uniformly scales all dimensions of depth/width/resolution of the CNN.

Different from CNN, *RNN* allows operations over the sequence of vectors through time. One of better models was introduced by Gers and Schmidhuber, named *Long Short-Term Memory* (LSTM) [4]. The key idea of LSTMs is the cell state called gates: input gate, forget gate and output gate. The *Gated Recurrent Unit* (GRU) [12] came from LSTMs with slightly more variation. It combines the forget and input gates into a single "update gate" and merges the cell state and hidden state along with some other changes.

GANs [11], proposed by Goodfellow in 2014, are an unsupervised approach where the generator and the discriminator compete against each other in a zero-sum game. Each of the two networks gets better at its given task with each iteration. Typically, the generator starts with Gaussian noise to generate images and the discriminator determines how good the generated images are. This process continues until outputs of the generator become close to actual input samples. There are some GANs proposed in various applications [40], such as Style GAN, Info-GAN, Laplacian Pyramid of Adversarial Networks (LAPGAN), Deep Convolution GAN (DCGAN) and Wasserstein GAN etc.

There are some limitations of using simple GAN: 1) images are generated from input noise, 2) GANs differentiate between 'real' and 'fake' objects. An AE is a DNN approach used for unsupervised feature learning with the encoder and the decoder. A special AE called *Variational Autoencoder* (VAE) [10] is proposed to overcome those limitation of basic GANs, where the latent vector space is used to represent the images which follow a unit Gaussian distribution. In VAE, there are two losses, i.e. a mean squared error that determines how good the network is doing for reconstructing image and a latent loss (the Kullback-Leibler divergence) that determines how closely the latent variable match is with unit Gaussian distribution.

To make machine learning techniques easier to apply and reduce the demand for experienced human experts, *automated machine learning* (AutoML) has emerged as a hot topic [32]. ML experts can profit from AutoML by automating tedious tasks like *hyperparameter optimization* (HPO) leading to a higher efficiency, for example, *Bayesian optimization*. The famous AutoML in deep learning is *neural architecture search* (NAS) proposed by Google [31]. Besides that, meta-learning is a popular technique used in AutoML. *Meta-learning* is the science of learning how ML algorithms learn [25], in which a set of approaches learn from the prior models, like *transfer leaning* [24], *few-shot learning* [37] and even *zero-shot learning* [38].

*Graph neural networks* (GNNs) capture the dependence of graphs via message passing between the nodes of graphs [33]. Unlike standard neural networks, GNNs retain a state that can represent information from its neighbourhood with arbitrary depth. Like CNN, GNNs also include recurrent GNN, convolutional GNN, graph Auto Encoder, and Spatial-temporal GNN.

Optimization in training a deep learning model is critical, to avoid overfitting, gradient exploding or diminishing and to accelerate the training process. There have been many methods and "tricks" proposed [29, 30], such as pre-training and Xavier initialization, data augmentation, *stochastic gradient descent* (SGD), momentum, weight decay, adaptive learning rate, Dropout, *Batch Normalization* (BN), *Nesterov Accelerated Gradient* (NAG), ADAM, AdaGrad, AdaDelta or RMSProp etc.

There are open deep learning platforms for researchers and engineers to design and develop models [23, 28], such as PyTorch, Tensorflow, MxNet, Caffe and CNTK etc.

### B. *Distributed Learning*

Accelerating deep learning training is a major challenge and techniques range from distributed algorithms to low-level circuit design, where a main effort is to exploit their inherent *parallelism*. Most of the operations in learning can be modelled as operations on tensors as a parallel programming model. They are highly data-parallel and only summations introduce dependencies [27].

Every computation on a computer can be modelled as a *directed acyclic graph* (DAG). The vertices of the DAG are the computations and the edges are the data dependencies (or data flow). In deep learning, one often needs to reduce (sum) large tables of multiple independent parameters and return the result to all processes. This is called *AllReduce* in the MPI specification.

Deep learning usually applies minibatches, which make it possible to partition both the forward inference flow and the backpropagation training flow among parallel processors. There are three kinds of partitioning methods, one by input samples (*data parallelism*), one by network structure (*model parallelism*), and one by layer (*pipelining*). A hybrid method can combine them together to overcome each one's drawbacks.

In distributed environments, there may be multiple training agents running SGD instances independently. Then the distribution schemes for deep learning can be categorized along three axes: *model consistency* (synchronous, stale-synchronous or asynchronous), *parameter distribution and communication* (centralized such as *parameter server*, or decentralized), and *training distribution* (model consolidation and optimization).

It is seen existing cloud computing is unable to manage distributed learning tasks, *edge computing* [34, 35] turns out to be an alternative. Compared with cloud computing only, edge computing combined with cloud computing can achieve backbone network alleviation, agile service response and powerful cloud backup.

Dividing the DL model horizontally or vertically in distributed computation and pushing part of DL tasks from the cloud to the edge can improve the throughput of the cloud. Meanwhile, the edge architecture can alleviate the pressure of networks by processing the data or training at themselves, where some distributed learning strategies are adopted, like *gradient descent compression* (GDC) divided into gradient sparsification and gradient quantization, *gossip-type algorithms* for decentralized training.

*Federated learning* is a special machine learning setting proposed by Google, where not a few clients train a model collaboratively under the orchestration of a central server, while preserving the training data decentralized [36]. Federated learning allows focused data collection and minimization, so it alleviates the systemic privacy risks and costs.

Federated learning can deal with some challenges in edge computing networks, like Non-IID training data, limited communication, unbalanced contribution, and privacy/security.

### C. *Model Compression and Acceleration*

Deep neural network models are computationally expensive and memory intensive, prohibiting their deployment in devices with small memory resources or in applications with low latency requirements. A solution is to perform model compression and acceleration without significantly decreasing the model performance. So far there are some techniques proposed for use [21, 22], roughly categorized into four categories: *parameter pruning and sharing, low-rank factorization, transferred/compact convolutional filters, and knowledge distillation*.

MobileNets, put forward by Google, are based on a streamlined architecture that uses depth-wise separable convolutions to build light weight deep neural networks [41]. In its version 2, the inverted residual structure with shortcut connections between the thin bottleneck layers is proposed [42]. In version 3 (shown in Fig. 3), a combination of hardware aware NAS, complemented by the NetAdapt algorithm and Squeeze-and-Excite (SE), tunes the model to the mobile device [43].

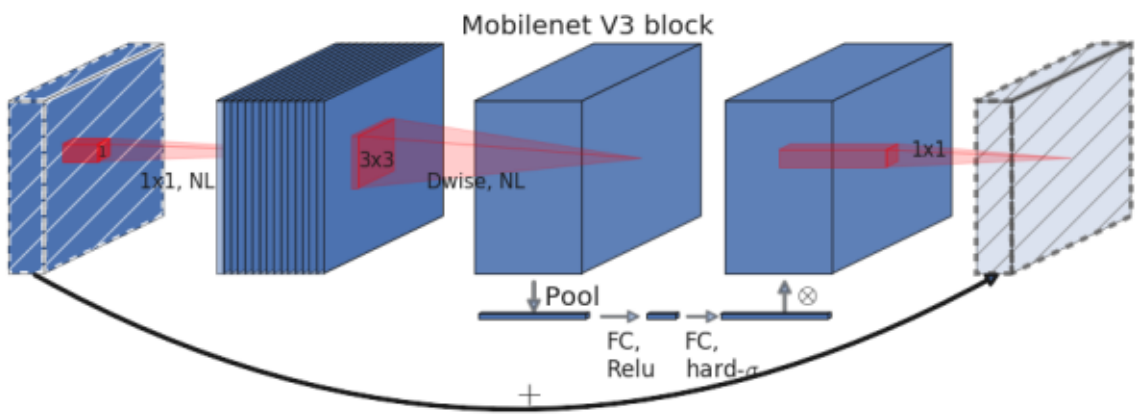

Fig. 3. MobileNets v3 with SE, from reference [43].

Due to the difficulty of deploying various deep learning models on diverse DL hardware, to develop the *deep learning compilers* gets important. Several compilers have been proposed such as Tensorflow XLA/MLIR and the open source TVM [39]. The deep learning compilers generate efficient code implementations on various DL hardware from the model definitions in the DL frameworks. Aiming at the model specification and hardware architecture, the transformation between model definition and specific code implementation are highly optimized. The compiler design can be separated into frontend, multi-level *Intermediate Representations* (IRs) and backend.

## III. Overview of Autonomous Driving

There have been some survey papers about the self-driving technologies, from the whole system/platform to individual modules/functions [45-58]. In this section, we briefly review the basic autonomous driving functions and modules, shown in Fig. 4, hardware and software architecture, perception, prediction, planning, control, safety, simulation, and V2X etc.

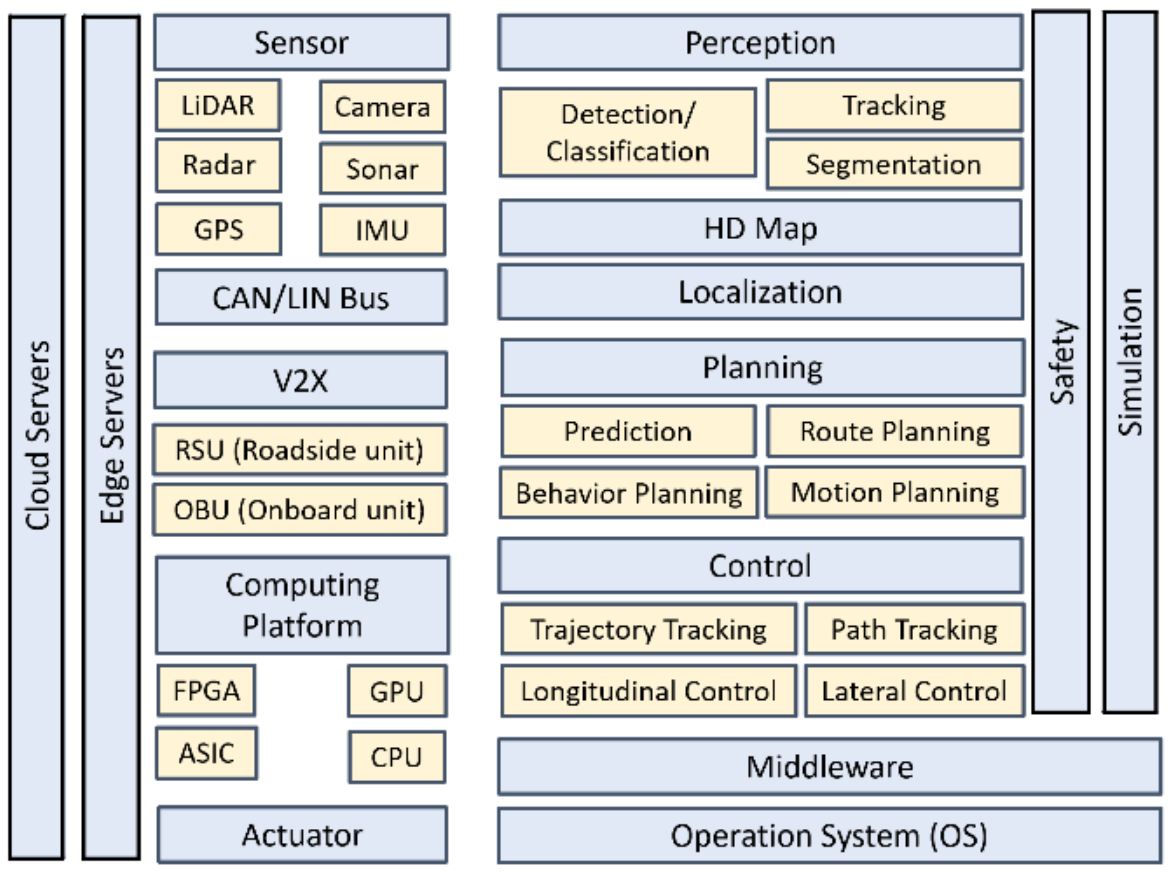

Fig. 4. HW and SW of the autonomous driving platform.

### A. Automation Levels

The US Department of Transportation and the *National Highway Traffic Safety Administration* (NHTSA) had adopted the *Society of Automotive Engineers* (SAE) international standard for automation levels which define autonomous vehicles from Level 0 (the human driver has full control) to Level 5 (the vehicle completely drives itself).

In level 1, the driver and the automated system control the vehicle together. In level 2 the automated system takes full control of the vehicle, but the driver must be prepared to intervene immediately at any time. In level 3, the driver can be free from the driving tasks and the vehicle will call for an immediate response, so the driver must still be prepared to intervene within some limited time. In level 4, it is the same to level 3, but no driver attention is ever required for safety, e.g. the driver may safely go to sleep or leave the driver's seat.

### B. Hardware

Autonomous driving vehicle test platforms should be capable of realizing real-time communication, such as in *controller area network* (CAN) buses, and can accurately complete and control the directions, throttles, and brakes of vehicles in real time. Vehicle sensor configurations are conducted to meet the reliability requirements of environmental perception and to minimize production cost.

Sensing of autonomous driving vehicles falls into three main categories: *self-sensing, localization and surrounding sensing*. Self-sensing measures the current vehicle state, i.e. velocity, acceleration, yaw, and steering angle etc. with *proprioceptive sensors*. *Proprioceptive sensors* include odometers, *inertial measurement units* (IMUs), gyroscopes, and the CAN bus. *Localization*, using external sensors such as *global positioning system* (GPS) or dead reckoning by IMU readings, determines the vehicle’ s global and local position. *Surrounding sensing* uses *exteroceptive sensors* to perceive road markings, road slope, traffic signs, weather conditions and obstacles.

Proprioceptive and exteroceptive sensors can be categorized as either active or passive sensors. *Active* sensors emit energy in the form of electromagnetic waves and measure the return time to determine parameters such as distance. Examples include sonar, radar, and *Light Detection And Ranging* (LiDAR) sensors. *Passive* sensors do not emit signals, but rather perceive electromagnetic waves already in the environment (e.g., light-based and infrared cameras).

Another important issue is the computing platform, which supports sensor data processing to recognize the environments and make the real-time control of the vehicles through those computationally intensive algorithms of optimization, computer vision and machine learning. There are different computing platforms, from CPUs, GPUs, ASIC to FPGAs etc. To support AI-based autonomous driving, cloud servers are required to support big data service, such as large-scale machine learning and large size data storage (for example, HD Map). To support vehicle-road collaboration, edge communication and computing devices are required from both the vehicle side and the roadside.

### C. Software

A software platform of autonomous driving is classified multiple layers, from bottom to top as the *real time operating system* (RTOS), middleware, function software and application software. The software architecture could be end-to-end or modular style.

*End-to-end* systems generate control signals directly from sensory inputs. Control signals can be operation of steering wheel and pedals (throttles and brakes) for acceleration/deceleration (even stop) and turn left/right. There are three main approaches for end-to-end driving: direct supervised deep learning, neuro evolution and deep reinforcement learning.

*Modular* systems are built as a pipeline of multiple components connecting sensory inputs to actuator outputs. Key functions of a modular autonomous driving system (ADS) are regularly summarized as: perception, localization and mapping, prediction, planning and decision making, and vehicle control etc.

- *Perception* collects information from sensors and discovers relevant knowledge from the environment. It develops a contextual understanding of driving environment, such as detection, tracking and segmentation of obstacles, road signs/marking and free space drivable areas. Based on the sensors implemented, the environment perception task can be tackled by using LIDARs, cameras, radars or a fusion between these three kinds of devices.
- At the highest level, perception methods can fall into three categories: *mediated perception, behaviour reflex perception, and direct perception*. Mediated perception develops detailed maps of the surroundings as vehicles, pedestrians, trees, road markings, etc. Behaviour reflex perception maps sensor data (image, point cloud, GPS location) directly to driving manoeuvres. Direct perception combines behaviour reflex perception with the metric collection of the mediated perception approach.
- *Mapping* refers to building the map with information of roads, lanes, signs/markings and traffic rules etc. In general, there are two main categories of maps: planar which refers to maps that rely on layers or planes on a Geographic Information System (GIS), e.g. High Definition (HD) maps, and point-cloud which refers to maps based on a set of data points in the GIS.
- *Localization* determines its position with respect to the driving environment. *Global Navigation Satellite Systems* (GNSS) such as GPS, GLONASS, BeiDou, and Galileo rely on at least four satellites to estimate global position at a relatively low cost. GPS accuracy can be improved upon by using Differential GPS. GPS is often integrated with IMU to design a low-cost vehicle localisation system. IMUs have been used to estimate vehicle position relative to its initial position, in a method known as Dead Reckoning.
- Recently, many studies have emerged on self-contained odometry methods and *simultaneous localization and mapping* (SLAM). Usually the SLAM techniques apply an odometry algorithm to obtain the pose where later fed into a global map optimization algorithm. Visual SLAM is still a challenging direction due to drawbacks of image-based computer vision algorithms, like feature extraction and matching, camera motion estimation, 3-D reconstruction (triangulation) and optimization (bundle adjustment).
- *Prediction* refers to estimating the obstacles' trajectories based on their kinematics, behaviours and long-term/short-term histories.
- *Planning* makes decisions on taking the vehicle to the destination while avoiding obstacles, which generates a reference path or trajectory. Planning can be classified as route (mission) planning, behaviour planning and motion planning at different levels.
- *Route planning* is referred as finding the point-to-point shortest path in a directed graph, and conventional methods are examined under four categories as goal-directed, separator-based, hierarchical and bounded-hop techniques.
- *Behavioural planning* decides on a local driving task that progresses the vehicle towards the destination and abides by traffic rules, traditionally defined by a finite state machine (FSM).
- *Motion planning* then picks up a continuous path through the environment to accomplish a local driving task, for example RRT and Lattice planning.
- *Control* executes the planned actions by selecting appropriate actuator inputs. Usually control could be split into lateral and longitudinal control. Mostly the control design is decoupled into two stages, trajectory/path generation and tracking, for example the pure pursuit method. However, it can generate the trajectory/path and track simultaneously.
- *V2X (vehicle to everything)* is a vehicular technology system that enables vehicles to communicate with the traffic and the environment around them, including vehicle-to-vehicle communication (V2V) and vehicle-to-infrastructure (V2I). From mobile devices of pedestrians to stationary sensors on a traffic light, an enormous amount of data can be accessed by the vehicle with V2X. By accumulating detailed information from other peers, drawbacks of the ego vehicle such as sensing range, blind spots and insufficient planning may be alleviated. The V2X helps in increasing safety and traffic efficiency.

It is worth to mention, the *ISO* (International Organization for Standardization) *26262* standard [50] for *functional safety* of driving vehicles defines a comprehensive set of requirements for assuring safety in vehicle software development. It recommends the use of a Hazard Analysis and Risk Assessment (*HARA*) method to recognize hazardous events and to define safety goals that mitigate the hazards. Automotive Safety Integrity Level (*ASIL*) is a risk classification scheme defined in ISO 26262 in an automotive system.

Besides of either modular or end-to-end system, there is an important platform "*simulation*" in ADS development. Since the driving of an experimental vehicle on the road still costs highly and experiments on existing human driving road networks are restricted, a simulation environment is beneficial for developing certain algorithms/modules before real road tests.

A *simulation* system consists of the following core components: sensor modelling (cameras, radar, LiDAR and sonar), vehicle dynamics and kinematic, shape and kinematic modelling of pedestrians, motorists and cyclists, road network and traffic network, 3-D virtual environment (urban and rural scenes) and driving behaviour modelling (age, culture, race etc.)

## IV. PERCEPTION

There are two survey papers related to the perception module [51, 52]. In this section, we focus on the detection,

reconstruction (depth) and sensor fusion, while mentioning other fields as image processing (denoising and super-resolution), segmentation, motion estimation, tracking and human pose estimation (used for pedestrian movement analysis). The detection part is split into 2-D and 3-D. The 3-D method is classified as camera-based, LiDAR-based, radar-based and sensor fusion-based. Similarly, depth estimation is categorized as monocular image-based, stereo-based and sensor fusion-based.

### A. *Image Processing*

Image quality and resolution are requested in the perception. The existing denoising methods can fall into end-to-end CNN and combination of CCN with prior knowledge. A survey of image denoising by deep learning is given in [60].

The super-resolution methods based on deep learning can be roughly categorized into supervised and unsupervised. The supervised manner is split into pre-upsampling, post-upsampling, progressive-upsampling and iterative up-and-down sampling. The unsupervised manner could be zero-shot learning, weak supervised and prior knowledge-based. An overview of image super-resolution refers to [59].

### B. *2-D Detection*

There are good survey papers in this domain [76, 77]. Here we only briefly introduce some important methods in the short history.

The object detection by deep learning are roughly named as *one-stage* and *two-stage* methods. The first two stage method is R-CNN (region-based) [61] with feature extraction by CNN; fast R-CNN [63] improves it with *SPP* (spatial pyramid pooling) [62] layers that generate a fixed-length representation without rescaling; faster RCNN [64] realizes end-to-end detection with the introduced *RPN* (region proposal network); *FPN* (feature pyramid network) [67] proposes a top-down architecture with lateral connections to build the feature pyramid for detection with a wide variety of scales; a GAN version of fast RCNN is achieved [69].

*YOLO* (You Only Look Once) is the first one-stage detector in deep learning era, which divides the image into regions and predicts bounding boxes and probabilities for each region simultaneously [66]. *SSD* (single shot detector) was the second one-stage detector with the introduction of the multi-reference and multi-resolution [65], which can detect objects of different scales on different layers of the network. *PeleeNet* [72] is used for detection as the SSD backbone, a combination of DenseNet and MobileNet.

YOLO already has 4 versions [68, 70-71, 75], which further improve the detection accuracy while keeps a very high detection speed:

- YOLO v2[68] replaces dropout and VGG Net with BN and GoogleNet respectively, introduces anchor boxes as prior in training, takes images with different sizes by removing fully connected layers, apply DarkNet for acceleration and WordTree for 9000 classes in object detection;

- YOLO v3 [71] uses multi-label classification, replaces the softmax function with independent logistic classifiers, applies feature pyramid like FPN, replace DarkNet-19 with DarkNet-53 (skip connection);

- YOLO v4 [75] uses Weighted-Residual-Connections (WRC) and Cross-Stage-Partial-Connections (CSP), takes Mish-activation, DropBlock regularization and Cross mini-Batch Normalization (CmBN), runs Self-adversarial-training (SAT) and Mosaic data augmentation in training, and CIoU loss in bounding box regression.

*RetinaNet* is a method with a new loss function named “focal loss” to handle the extreme foreground-background class imbalance [70]. Focal Loss enables the one-stage detectors to achieve comparable accuracy of two-stage detectors while maintaining very high detection speed. *VoVNet* [73] is another variation of DenseNet comprised of One-Shot Aggregation (OSA) applied for both one-stage and two-stage efficient object detection. *EfficientDet* [74] applies a weighted bi-directional FPN (BiFPN) and EfficientNet backbones.

Recently **anchor-free** methods get more noticed due to the proposal of FPN and Focal Loss [78-90]. Anchors are introduced to refine to the final detection location, first occurred in SSD, then in faster R-CNN and YOLO v2. The anchors are defined as the grid on the image coordinates at all possible locations, with different scale and aspect ratio. However, it is found fewer anchors result in better speed but deteriorate the accuracy.

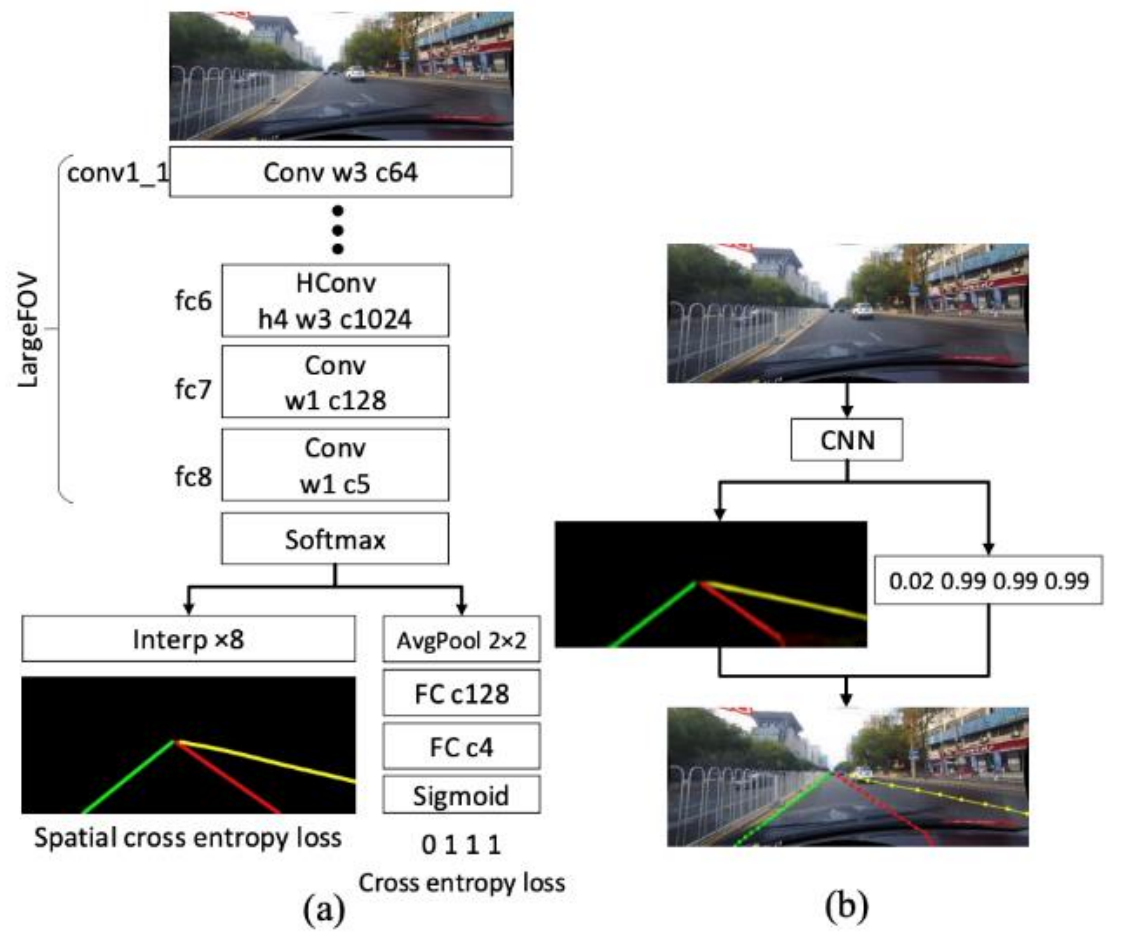

Fig. 5. Spatial CNN for lane detection, from reference [94].

There are not a few anchor-free methods proposed [78-90], such as CornerNet [78], FCOS (fully convolutional one-stage) [81], FoveaBox [82], Objects as Points [84], CenterNet with Keypoint Triplets [85], CornerNet-lite [86], WSMA-Seg (weakly supervised multimodal annotation segmentation) [88] and CentripetalNet [90] etc. Recently a noticeable work [89] finds the gap between anchor-based and anchor free methods to discriminate them on how to define positive and negative training samples, which proposes an *Adaptive Training Sample Selection* (ATSS) to automatically select positive and negative samples according to statistical characteristics of object.

Besides, there are special objects for autonomous driving to detect/classify, i.e. lane and road markings, traffic sign and traffic light. For instances, a fully convolutional network [91] is designed for traffic sign detection and classification based on the built Tencent traffic sign database; *VPGNet* (Vanishing Point Guided Network) [92] is proposed for lane and road marking detection and recognition; SSD is applied for traffic light detection in [93]; *Spatial CNN* [94] is proposed (shown in Fig. 5) for lane detection, which generalizes traditional deep layer-by-layer convolutions to slice-by-slice convolutions within feature maps, thus enabling message passing between pixels across rows and columns in a layer.

### C. 3-D Detection

Two survey papers [152-153] investigate 3-d detection from both camera and LiDAR data, individually or jointly (in a fusion manner). Two survey papers [124-125] focus on 3-D point clouds for 3D shape classification, 3D object detection and tracking, and 3D point cloud segmentation.

For 3-D sensors, like LiDAR and depth sensor (RGB-D), 3-D object detection is direct by finding 3-D bounding box. For single camera, the 3-D object detection needs extensive inference beyond the simple 2-D bound box, to estimate the 3-D bounding box and 3-D pose. Radar can find the object information limited to the scan plane.

Below we start investigation from LiDAR-based methods to more challenging camera-based methods (monocular and stereo), and eventually the radar-based methods in a concise way.

#### 1) LiDAR-based

LiDAR sensors obtain the point cloud data from the surroundings, so the detection methods could be roughly categorized as 3-D volume-based and projection-based (mapping 3-D data onto 2-D planes), shown in Table I. Like 2-D detection, the algorithms can fall into one stage and two stage methods too.

First projection-based neural networks are analysed. Li, Zhang and Xia [95] project the point cloud to a 2D point map analogous to cylindric images, while a kind of FCN (fully convolution network) is designed to regress the 3-D bounding box and classify the objectness with two branches. Based on a grid map environment representation, Wirget et al. [99] realize free-space estimation, detect and classify objects using deep CNNs, in which the range sensor measurements are transformed to a multi-layer top view grid map to infer rotated 3D bounding boxes together with semantic classes.

Real-time 3-dimensional (*RT3D*) vehicle detection method [100] also projects the point cloud onto the bird view first, and then a pose-sensitive feature map design is activated by the relative poses of vehicles for estimation of location, orientation, and size of vehicles. Similarly, *BirdNet* [101] also projects the 3-D LiDAR information into a cell encoding for Bird's Eye View (BEV) projection. In *LMNet* [102], the input consists of 5 perspective representations of the unorganized point cloud data as reflection, range, and the position on each of the three axes. *HDNet* [103] is a one stage detector to extract geometric and semantic features from the HD maps in a BEV (bird-eye view) representation, in which a map prediction module estimates the map on the fly.

TABLE I. LIDAR-BASED 3-D OBJECT DETECTION METHODS

| 3-D Volume-based | Projection-based | |
|---|---|---|
| | BEV | Frontal view |
| VoxelNet [96]<br>PointNet[97]/PointNet++ [98]<br>IPOD [104]<br>SECOND [107]<br>PointRCNN [113]/Fast Point RCNN [119]<br>PointPillars [114]<br>Part A2-Net [116]<br>Voxel FPN [117]<br>STD [118]<br>StarNet [120]<br>Sparse 3D CNN [121]<br>VoteNet [122] | RT3D [100]<br>BirdNet [101]<br>HDNet [103]<br>PIXOR [105]<br>YOLO3D [108]<br>Complex-YOLO [109]<br>YOLO4D [110]<br>FaF [112] | Frontal view FCN [95]<br>Grid map CNN [99]<br>LMNet [102]<br>DepthCN [106]<br>Deconvolutional Network [111]<br>FVNet [115]<br>LaserNet [123] |

*PIXOR* (ORiented 3D object detection from PIXel-wise NN predictions) [105] is a proposal-free, single-stage detector by also representing the scene from the BEV. Instead, DepthCN [106] transforms the point cloud to a Dense-depth Map (DM) followed by ground points removal (segmentation) and obstacle (object hypotheses) projection onto the DM, then the bounding boxes are fitted to the segmented objects as vehicle hypotheses used for verification by a ConvNet.

An extension of YOLO v2 to 3-D space is proposed called *YOLO3D* [108], which the loss function is extended to include the yaw angle, the 3D box centre and the height as a direct regression problem after projection of 3-D point cloud to two BEV grid maps (density and height). Similar to YOLO3D, *Complex-YOLO* is proposed [109] where the novel point is a specific complex regression strategy to estimate multi-class 3D boxes in Cartesian space and a *Euler-Region-Proposal Network* (E-RPN) is designed to estimate the object pose. *YOLO4D* [110] is extension of YOLO3D in the spatial-temporal domain with a convolutional LSTM.

Vaquero et al. extend the detection with tracking jointly by MH-EKF [111], where the 3D point cloud data is projected to a featured image-like representation containing ranges and reflectivity information (frontal view). Instead Uber's FaF (*Fast and Furious*) [112] realizes the detection, tracking and motion broadcasting based on the projected BEV, and specially within a single neural network (3-D CNN).

*FVNet* [115] is a two-stage method, consisting of two stages: generation of front-view (cylindrical surface) proposals and estimation of 3D bounding box parameters. LaserNet [123] projects LiDAR data onto the frontal image plane to get a small and dense range image, then uses a fully convolutional network to predict a multimodal distribution over 3D boxes for each point, which are fused to generate a prediction for each object.

Next, the volume-based methods follow. *VoxelNet* is proposed [96], which divides a point cloud into equally spaced 3D voxels and transforms a group of points within each voxel into a unified feature representation through the voxel feature encoding (VFE) layer that is connected to a RPN to generate bounding box prediction within a single stage, end-to-end trainable deep network. *PointNet* [97] is a deep net architecture that learns both global and local point features, providing a simple, efficient and effective approach for 3-D recognition and segmentation.

However, PointNet does not capture local structures induced by the metric space points live in, so *PointNet++* [98] is designed, which applies PointNet recursively on a nested partitioning of the input point set with a hierarchical NN and is able to learn local features with increasing contextual scales. IPOD (Intensive Point-based Object Detector) is a two-stage method [104], where features of all points within a proposal are extracted from the backbone network and PointNet++ is applied to build 3-D feature representation for final bounding inference.

*SECOND* (Sparsely Embedded Convolutional Detection) with sparse convolution [107] (shown in Fig. 6) introduces a new form of angle loss regression to improve the orientation estimation performance and a new data augmentation approach to enhance the convergence speed and performance within a two-stage voxel feature encoding (VFE) detection network. *PointRCNN* is a two-stage method [113]: in stage-1 the proposal generation sub-network directly generates a small number of high-quality 3D proposals from point cloud in a bottom-up manner via FG-BG segmentation, and in stage-2 the refinement sub-network transforms the pooled points of each proposal to canonical coordinates to learn local spatial features, which is combined with global semantic features of each point learned in stage-1.

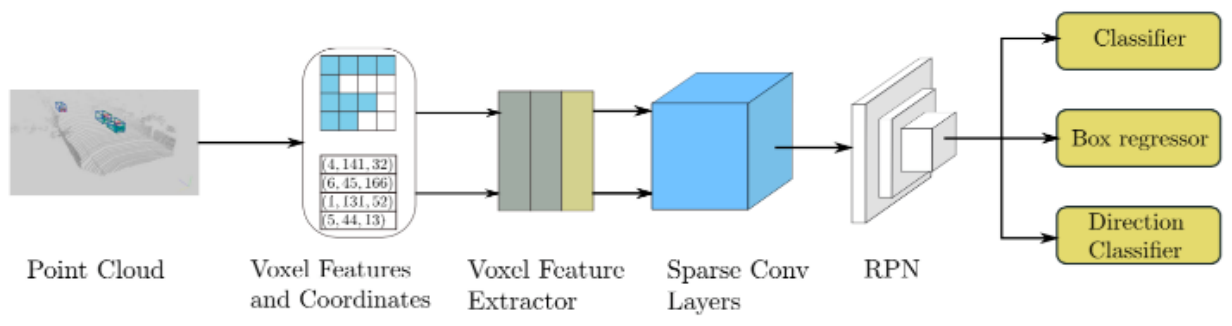

Fig. 6. SECOND for LiDAR-based object detection, from reference [107].

*PointPillars* is an encoder which utilizes PointNet to learn a representation of point clouds organized in vertical columns (pillars) [114], in which the raw point cloud is converted to a stacked pillar tensor and pillar index tensor and then the encoded features can be used with any standard 2D convolutional detection architecture. A two-stage method with the designed *part-aware and aggregation neural network* (Part-$A^2$ Net) [116] is proposed, where the part-aware stage learns to simultaneously predict coarse 3D proposals as well as intra-object part locations and the part-aggregation stage with RoI (region of interest)-aware point cloud pooling learns to re-score the box and refine the location.

*Voxel-FPN* is a one-stage object detector [117], in which the encoder extracts multi-scale voxel information in a bottom-up manner while the decoder fuses multiple feature maps from various scales in a top-down way. The *STD* (sparse-to-dense) Detector is a two-stage method instead [118], where the first stage generates proposals by seeding each point with a new spherical anchor and transforming interior point features from sparse expression to compact representation by *PointsPool* (based on PointNet), and the second stage implements a parallel intersection-over-union (IoU) branch to increase awareness of localization accuracy.

*Fast Point RCNN* [119] is similar to Point RCNN, where coordinate and indexed convolutional feature of each point in initial prediction are fused with the attention mechanism after *VoxelRPN*, preserving both accurate localization and context information. Based on the observation of point cloud data that objects are sparse and vary widely in location without exhibition of scale distortions, *StarNet* proposed [120] the data-driven object proposals to maximize spatial coverage or match the observed densities of point cloud data.

In nuScenes' 3D Detection Challenge, the winner method [121] utilizes sparse 3D convolution to extract semantic features fed into a class-balanced multi-head network and designs a class-balanced sampling and augmentation strategy to generate a more balanced data distribution. *VoteNet* is an end-to-end network which votes to object centres, and then groups and aggregates them to predict 3D bounding boxes and semantic classes of objects [122], with the backbone network constructed by PointNet++ layers.

Shown in Table X and Table XII (from reference [124]), the LiDAR-based 3D detection performance of different methods is compared on the Kitti dataset's 3D and BEV benchmark respectively, with respect to the object types as cars, pedestrians and cyclists etc.

In summary, volume-based methods are more accurate in detection but computationally expensive; instead projection-based methods are more efficient, though less accurate. In projection-based methods, BEV shows more potentials to achieve better detection results than frontal-view.

### 2) *Camera-based*

The camera-based 3-D detection methods can be classified as proposal-based, 3D shape-based, 2D-3D geometry-based, depth map-based and transform-based, shown in Table II.

Chen et al. [126] propose a CNN pipeline to obtain 3-D object detections by generating a set of candidate class-specific object proposals, *Mono3D*, in which assumption of object candidates in 3D on the ground-plane can score each candidate box projected to the image plane via semantic and instance segmentation, context, location priors and shape. Deep *MANTA* (Many-Tasks) is proposed [127] applies a coarse-to-fine 2D object proposal that boosts the vehicle detection, where the vehicle is represented by a wireframe model of parts for building 3D multiple templates and 2D-3D matching (EPnP) is called to refine 3D bounding boxes from 2D bounding boxes.

TABLE II. CAMERA-BASED 3-D OBJECT DETECTION METHODS

| proposal-based | 3D shape-based | 2D-3D geometry |
|---|---|---|
| Mono3D [126]<br>Deep MANTA [127]<br>Multiple fusion [131]<br>MonoPSR [137]<br>M3D-RPN [142] | Deep MANTA [127]<br>Mono3D++ [133]<br>MonoDIS [138] | Mousavian et al. [128]<br>Joint detection and tracking [129]<br>MonoGRNet [132]<br>GS3D [135]<br>MonoDIS [138]<br>Shift R-CNN [139]<br>Two stage [140]<br>SS3D [141]<br>RTM3D [146]<br>Polygon-Cuboid [147] |
| **Depth map-based** | **Transform-based** | **Stereo-based** |
| Pseudo-LiDAR [134]<br>Pseudo-LiDRA Color [136]<br>ForeSeE [143]<br>RefinedMPL [144] | OFT [130]<br>MoVi-3D [145] | TLNet [149]<br>Stereo R-CNN [148]<br>Pseudo-LiDAR stereo [150]<br>Object centric [151] |

Mousavian et al.'s work [128] first regresses relatively stable 3D object properties using a deep CNN and then combines these estimates with geometric constraints provided by a 2D bounding box to produce a complete 3D bounding box. A joint detection-and-tracking framework [129] extends Mousavian et al.'s method [128] by leveraging 3D box depth-ordering matching for instance association and utilizing 3D trajectory prediction for re-identification of occluded vehicles with a motion learning module based on an LSTM for more accurate long-term motion extrapolation.

Roddick, Kendall and Cipolla [130] proposes the orthographic feature transform (*OFT*) to map image-based features into an orthographic 3D space, which can reason holistically about the spatial configuration of the scene with size and distance information, in an E2E deep learning architecture for 3D detection. Xu and Chen [131] implement an e2e multi-level fusion-based framework with the help of depth estimation composed of two parts: one for 2D region proposal generation and another for simultaneously predictions of objects' 2D locations, orientations, dimensions, and 3D locations.

Qin, Wang and Lu's [132] *MonoGRNet* is amodal 3D object localization via geometric reasoning in both the observed 2D projection and the unobserved depth dimension, consisting of four subnetworks, 2D object detection, instance depth estimation (IDE), 3D localization and local corner regression. Mono3D++ [133] uses a morphable wireframe model to generate a fine-scaled representation of vehicle shape and pose, plus unsupervised monocular depth and a ground plane constraint, to optimize two-scale projection consistency between the generated 3D hypotheses and their 2D pseudo-measurements.

Weng and Kitani [134] enhance LiDAR-based algorithms by performing monocular depth estimation, lifted to a point cloud representation, called *Pseudo-LiDAR point cloud*, and then detects 2D object proposals in the input image (2D-3D consistency and instance mask) and extracts a point cloud frustum from the pseudo-LiDAR for each proposal, later an oriented 3D bounding box is detected for each frustum. Ma et al. [136] use the same idea and perform the 3D detection using PointNet backbone net to obtain objects' 3D locations, dimensions and orientations with a multi-modal features fusion module to embed the complementary RGB cue into the generated point clouds.

Similary, MonoPSR [137] generates a 3D proposal per object in a scene from 2D object detector, then a point cloud is predicted in an object cantered coordinate system (instance reconstruction) to learn local scale and shape information with a projection alignment loss. *GS3D* [135] is an approach to obtain a coarse cuboid for each predicted 2D box and then guide determining the 3D box of the object by refinement by employing the visual features of visible surfaces.

Barabanau et al.'s method, *MonoDIS*, [138] build the multi-branch model around 2D keypoint detection with a conceptually simple geometric reasoning method based on a set of five 3D CAD models and instance depth clues. *Shift R-CNN* [139], adapts a Faster R-CNN network for regressing initial 2D and 3D object properties, which is passed through a final ShiftNet network that refines the result using the proposed *Volume Displacement Loss*.

Simonelli et al. [140] design a two-stage architecture consisting of a single-stage 2D detector (FPN as backbone) with an additional 3D detection head constructed on top of features pooled from the detected 2D bounding boxes. Joergensen, Zach and Kahl [141] propose a single-stage monocular 3D object detector, *SS3D*, which consists of a 2D object detector with uncertainty estimates and a 3D bounding box optimizer by a nonlinear LS method.

A standalone *3D region proposal network*, called M3D-RPN [142], leverages the geometric relationship of 2D and 3D perspectives, with global convolution and local depth-aware convolution to predict multi-class 3D bounding boxes. Wang et al. [143] first run foreground-background separated monocular depth estimation (ForeSeE) and then apply depth of foreground objects (like LiDAR) in 3D object detection and localization.

*RefinedMPL* [144] is a refined Pseudo-LiDAR method, which performs structured sparsification of foreground points identified with supervised 2D object detection-based method and unsupervised keypoint clustering-based method, before 3-D detection. Simonelli et al.'s work [145], a single-stage deep architecture called *MoVi-3D*, leverages geometrical information to generate virtual views, which present cropped and scaled version of the original image that preserves the scale of objects, reducing the visual appearance variability associated to objects placed at different distances from the camera.

Cai et al. [147] propose to decompose the detection problem into a structured polygon prediction task, a height-guided depth recovery task and a fine-grained 3D box refinement, where the structured polygon in the 2D image consists of several projected surfaces of the target object, inversely projected to a cuboid in the 3D physical world. A keypoint FPN-based method, called RTM3D [146], predicts the nine perspective keypoints of a 3D bounding box in image space, and then utilize the geometric relationship of 3D and 2D perspectives to recover the dimension, location, and orientation in 3D space.

There are some stereo images-based methods.

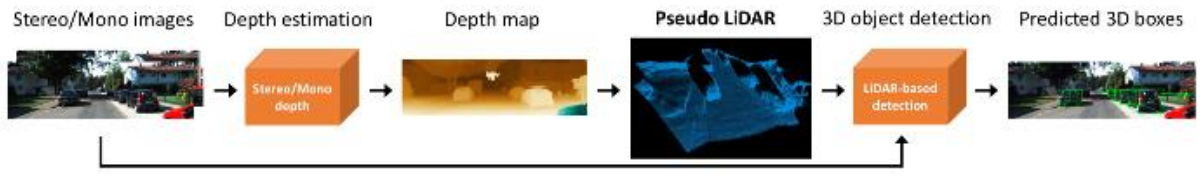

Fig. 7. Pseudo-LiDAR, from reference [150].

Wang et al. [150] propose the pseudo-LiDAR based on stereo images (shown in Fig. 7), which convert image domain to point cloud domain via depth map estimation. Stereo R-CNN by Li, Chen and Shen [148] extends faster RCNN to stereo images, fully exploiting the sparse and dense, semantic and geometry information with the proposed stereo Region Proposal Network (RPN) and region-based photometric alignment using left and right RoIs.

Qin, Zhang and Lu [149] employ 3D anchors to explicitly construct object-level correspondences between the ROI in stereo images, from which DNN learns to detect and triangulate the targeted object in 3D space, called TLNet (Triangulation Learning Network). Pon et al. [151] propose an object-centric stereo matching method that only estimates the disparities of the objects of interest.

Shown in Table XI and Table XIII (from reference [146]), the camera-based 3D detection performance of different methods is compared on the Kitti dataset's 3D and BEV benchmark respectively, with respect to the different scenarios as easy, moderate and hard etc.

Comparison of these methods shows that, depth map-based and stereo-based approaches work in the volume space, transform-based approaches work in the projected space, while 3D shape-based, proposal-based and 2D-3D geometry approaches work in both the volume and projection space. It looks working in both 3D volume and 2D projection space is more efficient with 2D-3D correspondence consistency constraints, in which proposal-based methods run in two stages, either 2D or 3D, and 3D shape-based methods consider 3-D structure constraints. The methods working only in the volume space may appear weak due to depth estimation artefacts and long tail on reconstructed object boundary.

#### 3) *Radar-based*

Recently, deep learning has been applied in radar-based object detection. Though the radar signal could be regarded as 3-D data, the specific scanning plane results in 2D signal processing for object detection.

Denzer et al. [154] detect 2D objects solely depending on sparse radar data using PointNets (with PointNet and Frustum PointNet), which are adjusted for radar data performing 2D object classification with segmentation, and 2D bounding box regression in order to estimate an amodal 2D bounding box.

RRPN [155], a radar- based region proposal algorithm (faster R-CNN), generates object proposals for object detection by mapping radar detections to the image coordinate system and generating pre-defined anchor boxes for each mapped radar detection point, which are transformed and scaled based on the object's distance from the vehicle.

Major et al. [156] work on range-azimuth-doppler (RAD) tensors instead of point cloud data for object detection with an encoder-decoder LSTM framework, where the feature extractor is a FPN architecture and the detector performs SSD with focal loss.

### *D. Tracking*

There are some survey papers about object tracking [237-239]. The tracking problem is typically solved in two different directions, single object tracking (SOT) [239] and multiple object tracking (MOT) [238]. It is also seen the tracking methods have been extended from camera-based visual solutions [239] to LiDAR point cloud-based solutions [237].

The SOT methods are categorized based on architecture (CNN, SNN, RNN, GAN, and custom networks), network exploitation (off-the-shelf deep features and deep features for visual tracking), network training for visual tracking (i.e., only offline training, only online training, both offline and online training), network objective (i.e., regression-based, classification-based, and both classification and regression-based), and exploitation of correlation filter advantages (i.e., DCF framework and utilizing correlation filter/layer/functions).

The MOT methods can apply deep learning in feature learning (similar to SOT), data association, end-to-end learning and state prediction etc.

Relatively radar-based object tracking is easily formulated due to its range sensing mechanism, compared to camera-based and LiDAR-based.

### E. *Segmentation*

Segmentation can be classified [232] as *semantic segmentation*, *instance segmentation* and *panoptic segmentation*. From the point view of deep learning models, they may fall into FCN (full convolution network), graph convolution network, encoder-decoder, multiple scale and pyramid network, dilated convolution, R-CNN, RNN, attention model and GAN etc. There are two survey papers about deep learning-based image segmentation methods [234].

Recently some new methods from image domain are put forward, such as 4D tensor-based TensorMask [228] for instance segmentation, PointRend (Point-based Rendering) [229] for both instance and semantic segmentation, a single-shot method CenterMask [230] for instance segmentation and EfficientNet-based EfficientPS (*Efficient Panoptic Segmentation*) [231].

LiDAR-based segmentation methods are overviewed in the respective part of the two survey papers [124-125]. Similarly, the segmentation methods are separated into semantic and instance segmentation groups. For methods of *semantic segmentation*, they are classified into projection-based or point-based. The projection-based methods are further split into representation as multi-view, spherical, volumetric, permutohedral lattice and hybrid. The point-based methods fall into pointwise MP, point convolution, RNN-based and graph-based. For methods of *instance segmentation*, they are grouped as proposal-based and proposal free, like 2-D object detection.

### F. *Depth Estimation*

Depth estimation from images is a reconstruction task in computer vision. Stereo matching could be categorized as bottom-up or top-down, 2-D feature-based or 3-D cost volume-based. A survey about reconstruction from images is given in [185]. Depth estimation from monocular image is more challenging than from stereo images. The methods in this domain fall into supervised or unsupervised, with different constraints from edge, surface normal, segment, pose and flow (videos), shown in Table III. A survey of monocular image-based methods is given in [186].

If we project LiDAR's 3-d point cloud to the image plane, it is seen that the generated depth map looks sparse and even with "holes" due to drawbacks of the LiDAR sensor itself. So, depth up-sampling, completion or inpainting from the LiDAR data are also useful. Fusion image information with depth map from LiDAR data can improve the depth resolution and accuracy.

We will introduce the typical deep learning-based methods in stereo-based disparity/depth estimation and then pay more attention to noticeable work of depth estimation from the monocular image only, at last some work of depth completion is also analysed (note: depth fusion sees the following session "sensor fusion").

TABLE III. MONOCULAR CAMERA-BASED DEPTH ESTIMATION METHODS

| Unsupervised/semi-supervised/self-supervised Methods | | | | |
|---|---|---|---|---|
| Stereo | Camera motion/pose | Object motion/flow | Normal/Edge | |
| Garg [161]<br>Godard [162]<br>Kuznietsov [163]<br>Luo [165]<br>Zhou, Fang & Liu [177] | Zhou [164]<br>G Wang [167]<br>Casser [168]<br>Yin & Shi [170]<br>DF-Net [171]<br>Andrahetti [176]<br>Shi [178] | G Wang [167]<br>Yin & Shi [170]<br>DF-Net [171]<br>Casser [174] | LEGO [172] | |
| Supervised Methods | | | | |
| Brute-force | Segment/Attention | Flow | Pose/VO | Normal |
| Eigen, Puhrsch & Fergus [157] | Eigen&Fergus [158]<br>Liu [160]<br>P Wang [161]<br>Jiao [166]<br>CC [173]<br>Bian [175] | CC [173]<br>Casser [174] | C Wang [159]<br>Casser [174] | Eigen&Fergus [158]<br>GeoNet [169] |

#### 1) *Stereo images-based*

An end-to-end method [180] for stereo matching forms a cost volume using deep feature representations is proposed, called *GC-Net* (Geometry and Context Network), in which 3-D convolution is performed through the volume.

*PSM-Net* [182], a pyramid stereo matching network, consists of two main modules: spatial pyramid pooling to use global context information by aggregating context in different scales and locations to form a cost volume, and 3D CNN to regularize the cost volume using stacked multiple hourglass networks.

A unsupervised stereo matching method [181] consists of four parts: disparity prediction, confidence map estimation, training data selection and network training, which starts with a randomly initialized network with two branches (the first for computing the cost volume and the other for jointly filtering the volume) and then left-right check is adopted to guide the training.

*SegStereo* [183] conducts semantic feature embedding from segmentation and regularizes semantic cues as the loss term to improve learning disparity. Similarly, *DispSegNet* [184] is an encoder-decoder network, which generates the initial disparity estimates from the feature map-based cost volume with 3-D CNN and refines with an embedding learned from the semantic segmentation branch.

#### 2) *Monocular image-based*

Eigen, Puhrsch and Fergus [157] define two deep network stacks for depth prediction: one that makes a coarse global prediction based on the entire image, and another that refines this prediction locally. Later Eigen and Fergus [158] propose a modified method, which progressively refines predictions using a sequence of scales and captures many image details on semantic labels and surface normals. Wang et al. [159] also propose a unified framework for joint depth estimation and semantic segmentation, which formulates the inference problem in a two-layer Hierarchical Conditional Random Field (*HCRF*) to produce the final depth and semantic map.

Liu et al. [160] propose a deep structured learning scheme to learn the unary and pairwise potentials of continuous CRF in a unified deep CNN framework for depth estimation, where the input image is first over-segmented into super-pixels. Garg et al. [161] propose an unsupervised framework to learn a deep CNN by training the network like an autoencoder, with stereo images. Godard et al.'s work [162] is also an unsupervised method, which exploits epipolar geometry constraints and then generates disparity images by training the network with an image reconstruction loss and a left-right consistency loss.

Kuznietsov et al. [163] propose a semi-supervised depth map prediction method, which uses sparse ground-truth depth from LiDAR for supervised learning and enforces the deep network to produce photo consistent dense depth maps in a stereo setup using a direct image alignment loss. Zhou et al. [164] also run an unsupervised learning framework for the task of monocular depth and camera motion estimation, combining of multi-view pose networks and a depth network.

Luo et al. [165] formulate the depth estimation as two sub-problems, a view synthesis procedure followed by stereo matching. Jiao et al. [166] propose a multi-task deep CNN consisting of four parts: the depth prediction sub-network, semantic labelling sub-network, knowledge sharing unit/connection, and the attention-driven loss. A direct method [167] argue that the depth CNN predictor can be learned without a pose CNN predictor, then it incorporates a differentiable implementation of DVO (direct visual odometry), along with a depth normalization strategy.

Casser et al. [168] discuss unsupervised learning of scene depth and ego-motion where supervision is provided by monocular videos, by modelling the scene and the individual objects with camera ego-motion and object motions estimation separately. Qi et al. [169] design Geometric Neural Network (*GeoNet*) to jointly predict depth and surface normal maps by incorporating geometric relation between depth and surface normal via the depth-to-normal and normal-to-depth networks.

Yin and Shi [170] propose a jointly unsupervised learning framework for monocular depth, optical flow and ego-motion estimation from videos, where geometric relationships are extracted over the predictions of individual estimation modules (pose net, depth net and flow net) combined in an image reconstruction loss and geometric consistency loss. Another unsupervised learning framework [171], called *DF-Net* (shown in Fig. 8), simultaneously train single-view depth prediction and optical flow estimation models, which can use the predicted scene depth and camera motion to synthesize 2D optical flow by back-projecting the induced 3D scene flow.

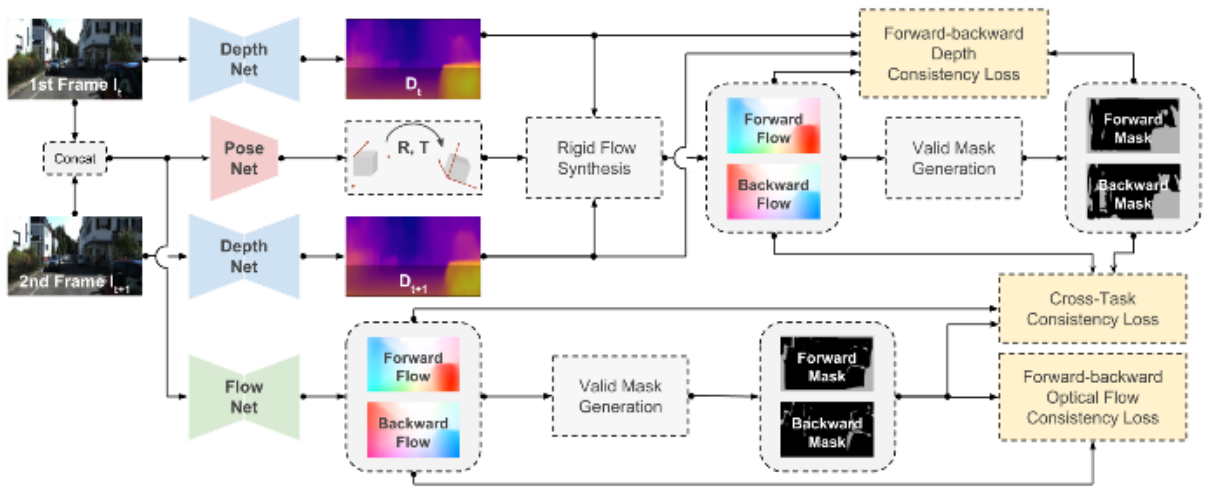

Fig. 8. DF-Net for depth estimation and pose estimation, from [171].

Yang et al. [172] design an unsupervised framework that Learns Edges and Geometry (depth, normal) all at Once (*LEGO*), which introduces a "3D as-smooth-as-possible (3D-ASAP)" prior inside the pipeline by requiring that any two points recovered in 3D from an image should lie on an existing planar surface. Nvidia researchers introduce *Competitive Collaboration* (CC) [173], a framework that facilitates the coordinated training of multiple specialized NNs to solve complex problems with NNs that act as both competitors to explain pixels that correspond to static or moving regions, and as collaborators through a moderator that assigns pixels to be either static or independently moving.

Casser et al. [174] extend their previous work [168] by modelling the motion of individual objects and learn their 3D motion vector jointly with depth and egomotion. To handle violation of the static scene assumption and scale-inconsistency, Bian et al. [175] propose a geometry consistency loss for scale-consistent predictions and an induced self-discovered mask for handling moving objects and occlusions.

Andraghetti et al. [176] improve monocular depth estimation by integrating into existing self-supervised networks a geometrical prior to process the output of conventional visual odometry algorithms working in synergy with depth-from-mono networks. Zhou, Fang and Liu [177] proposes an unsupervised method by incorporating left-right binocular images reconstruction loss based on the estimated disparities and edge aware smooth L2 regularization to smooth the depth map.

Shi et al. [178] propose a self-supervised method, which jointly optimizes the scene depth and camera motion via incorporating differentiable Bundle Adjustment (BA) layer by minimization of the feature-metric error and the photometric consistency loss with view synthesis as the final supervisory signal. Wang et al. [179] from Tsinghua U. apply joint unsupervised training of depth, pose and flow, which segments the occluded region explicitly, which framework is similar to Zhan et al.'s VO method [256] (introduced in "mapping and localization" session later).

Shown in Table XV and Table XVII (from reference [185]), the stereo matching performance on the Kitti 2015 benchmark and monocular depth regression performance on three benchmarks are compared respectively. Shown in Table XI (from reference [186]), the semi-supervised and unsupervised depth estimation performance on the Kitti benchmark is compared.

In summary, the depth estimation methods evolve from supervised to semi-supervised/unsupervised. It is shown a multiple task learning framework is stronger which is flexible to add supervising clues, such as stereo matching, camera pose, optic flow and normal/edge. In supervised methods, segmentation and attention mechanism are also used as additional constraints in depth estimation.

### 3) *LiDAR only*

Sparse invariant CNN [187] is proposed to handle sparsity of depth map from LiDAR data, which explicitly considers the location of missing data during the convolution operation. Eldesokey et al. [188] design an algebraically constrained convolution layer for CNNs with sparse input for depth regression, which produces a continuous pixel-wise confidence map enabling information fusion, state inference, and decision support.

A sparsity-invariant hierarchical multi-scale encoder-decoder network (HMS-Net) [189] is proposed for depth completion with three sparsity-invariant operations (addition, concatenation and convolution) to handle sparse inputs and sparse feature maps.

## *G. 2-D Optic Flow and 3-D Scene Flow*

Optic flow is pixel-level motion, from which the local object motion and global camera motion are estimated. The flow estimation from camera images is similar to disparity/depth estimation, either feature-based or regression-based, except the epipolar constraints of multiple view geometry. This kind of deep learning-based work is investigated in the survey paper [236], also classified as supervised, unsupervised or semi-supervised.

3-D scene flow estimation from point cloud data can see an overview in a session of the survey paper [124].

## *H. Sensor Fusion*

In the survey paper [235], three questions in sensor fusion are addressed. 1) What sensing modalities should be fused; 2) How to fuse with operations (average, concatenation, ensemble and mixture experts etc.); 3) When to fuse at given stages of feature representation in a NN.

The sensor fusion could be realized in data level and task level. A prerequisite work is calibration of multiple sensors, to determine transform of aligning the data from different sensors. First, we will introduce the deep learning method to camera-LiDAR calibration, then we will survey the methods of depth fusion from camera and LiDAR data, and eventually we will investigate the 3-D object detection methods with camera, LiDAR and/or radar.

### 1) *Calibration of LiDAR and Camera*

*RegNet* [190] casts all three conventional calibration steps (feature extraction, feature matching and global regression) into a single real-time capable CNN. *CalibNet* [191] is a self-supervised deep network capable of automatically estimating the 6-DoF rigid body transformation between a 3D LiDAR and a 2D camera in real-time, which trains the network to predict calibration parameters that maximize the geometric and photometric consistency of the input images and point clouds.

### 2) *Depth Fusion*

Similar to depth estimation from images, depth fusion methods with camera images and LiDAR are also categorized roughly as supervised and unsupervised, with different constraints from pose, flow, edge and surface normal etc., shown in Table IV.

Ma et al. [192] use a single deep regression network to learn directly from the RGB-D raw data, which explore the impact of number of depth samples on prediction accuracy. Later, they propose an unsupervised method [196] which utilizes the RGB image sequence to guide the training along with PnP and image warping. A depth completion method [193] is proposed which estimates surface normal and boundary before optimization of depth prediction.

TABLE IV. DEPTH FUSION METHODS WITH LIDAR AND CAMERA

| Sparsity-based | Segmentation/Attention | Stereo |
|---|---|---|
| Sparse-to-dense [192]<br>Deep Depth Completion [193]<br>Deep Depth Densification [194]<br>Eldesokey, Felsberg and Khan [205] | Jaritz [197]<br>Morphological [198]<br>CFCNet [204] | Park [195]<br>Gansbeke [202]<br>Wang [203] |
| **Image guided** | **Motion/Pose** | **Normal** |
| DPP [200]<br>DFuseNet [201]<br>Tang [206] | DFineNet [207]<br>PLIN [208] | Ma [196]<br>DeepLiDAR [199]<br>Xu [209] |

A method, called *Deep Depth Densification*, is proposed [194] with a parametrization of the sparse depth. Fusing of LiDAR and stereo is handled [195] with disparity estimation and RGB information in a deep CNN. Jariz et al. [197] propose to handle sparse depth data with optional dense RGB and accomplish depth completion and semantic segmentation changing only the last layer.

A method for completing sparse depth images in a semantically accurate manner by training a novel morphological NN is proposed [198], which applies an early fusion U-Net architecture to combine dilated depth channels and RGB and approximates morphological operations by Contra-harmonic Mean Filter layers trained in a contemporary NN framework. Qiu et al. design a modified encoder-decoder structure, called *DeepLiDAR* [199], which estimates surface normals as the intermediate representation to produce dense depth, trained end-to-end.

Yang, Wong and Soatto [200] design a deep learning system to infer the posterior distribution of a dense depth map associated with an image by exploiting sparse range measurements, for instance from a LiDAR, in which a Conditional Prior Network associates a probability to each depth value given an image and combines it with a likelihood term that uses the sparse measurements. *DFuseNet* [201] is a Spatial Pyramid Pooling (SPP)-based architecture that seeks to pull contextual cues separately from the intensity image and the depth features, and then fuse them later in the network.

Gansbeke et al. [202] propose LiDAR sparse depth completion with RGB guidance in a framework, which fuse information from two parts: the global branch to output a guidance map, a global depth map and a confidence map, and the local branch to predict a confidence map and a local map. Wang et al. [203] work on fusing LiDAR with stereo, in which a stereo matching network is improved with two enhanced techniques: *Input Fusion* to incorporate the LiDAR depth's geometric info with the RGB images and *Conditional Cost Volume Normalization* (CCVNorm) to adaptively regularize cost volume optimization.

Zhong et al. [204] design an encoder-decoder network, called Correlation For Completion Network (*CFCNet*) with defined Sparsity-aware Attentional Convolutions (*SAConv*), which decomposes RGB image into two parts by the mask, sparse RGB concatenated with sparse depth map and complementary RGB to complete the missing depth information. Eldesokey, Felsberg and Khan [205] apply an algebraically constrained normalized convolution layer for fusion of RGB image and sparse depth in CNNs, which determine the confidence from the convolution operation and propagating it to consecutive layers.

Inspired by the guided image filtering, Tang et al. [206] define a guided network to predict kernel weights from the guidance image, which are applied to extract the depth image features for dense depth generation by a depth estimation network. *DFineNet* [207] (shown in Fig. 9), an end-to-end learning algorithm that is capable of using sparse, noisy input depth for refinement and depth completion with RGB guidance, also produces the camera pose as a byproduct.

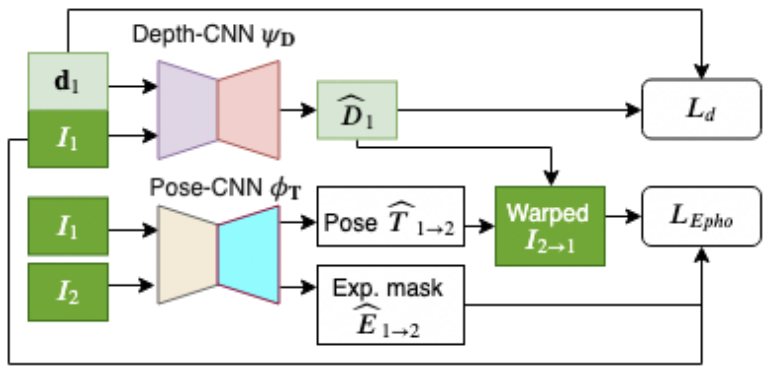

Fig. 9. DFineNet, from reference [207].

Different from depth interpolation, *PLIN* (Pseudo-LiDAR interpolation network) [208] increases the frequency of Pseudo LiDAR sensors by temporally and spatially generating high-quality point cloud sequences to match the high frequency of cameras by a coarse-to-fine cascade structure, which consists of three modules, i.e. motion guidance module, scene guidance module and transformation module. Xu et al. [209] propose a unified CNN framework which models the geometric constraints between depth and surface normal in a diffusion module and predicts the confidence of sparse LiDAR measurements to mitigate the impact of noise.

Shown in Table XIV (from reference [198]), the depth fusion performance of different methods with LIDAR and camera is compared on the Kitti benchmark.

Similar to depth estimation from monocular image, depth fusion methods could be supervised or unsupervised. On the one hand, special convolution layers with sparsity handling achieve inherently depth completion from sparse LiDAR point cloud. On the other hand, the guidance of the RGB image provides new information channel to depth completion. The depth fusion is also solved efficiently with multi-task learning, with supervising signals as stereo, motion, pose, normal, segmentation and attention etc.

### 3) 3-D Object Detection

Similarly, object detection methods with LiDAR and camera are also classified as volume-based, proposal-based, transform-based and projection-based, shown in Table V.

Chen et al. propose the LiDAR-and-camera-based detection method [210], called *MV3D*, which is composed of two subnetworks: one for 3D object proposal generation from the bird's eye view representation of 3D point cloud and another for multi-view feature fusion with interactions between intermediate layers of different paths. Matti, Ekenel and Thiran [211] work on pedestrian detection, where LiDAR data is utilized to generate region proposals by processing the 3-D point cloud, projected on the image space to provide a region of interest.

TABLE V. 3-D OBJECT DETECTION METHODS WITH CAMERA AND LIDAR

| Volume-based | Projection-based |
|---|---|
| LiDAR space clustering [211]<br>PointFusion [213]<br>Du [214]<br>Extension of LaserNet [222]<br>Pseudo-LiDAR++ [225] | MV3D [210]<br>Extension of MV3D by tracking [219]<br>Virtual Multi-View Synthesis [223] |
| **Proposal-based** | **Transform-based** |
| AVOD [216]<br>Frustum PointNet [217]<br>SIFRNet [220]<br>CAP [224]<br>MLOD [226]<br>3D Refinement [227] | Sparse-Nonhomogeneous pooling layer [212]<br>RoarNet [215]<br>Continuous Fusion Layer [218]<br>MVX-Net [221] |

Wang, Zhan and Tomizuka [212] construct a layer called *sparse non-homogeneous pooling layer* to **transform** features between bird's eye view and front view based on point cloud, where the encoder backbone for RGB image and LiDAR data is MSCNN and VoxelNet respectively. PointFusion [213] (shown in Fig. 10) consists of a CNN that extracts appearance and geometry features from input RGB image crops, a variant of PointNet that processes the raw 3D point cloud, and a fusion sub-network that combines the two outputs to predict 3D bounding boxes by prediction of multiple 3D box hypotheses prediction and their confidences using the 3D points as spatial anchors.

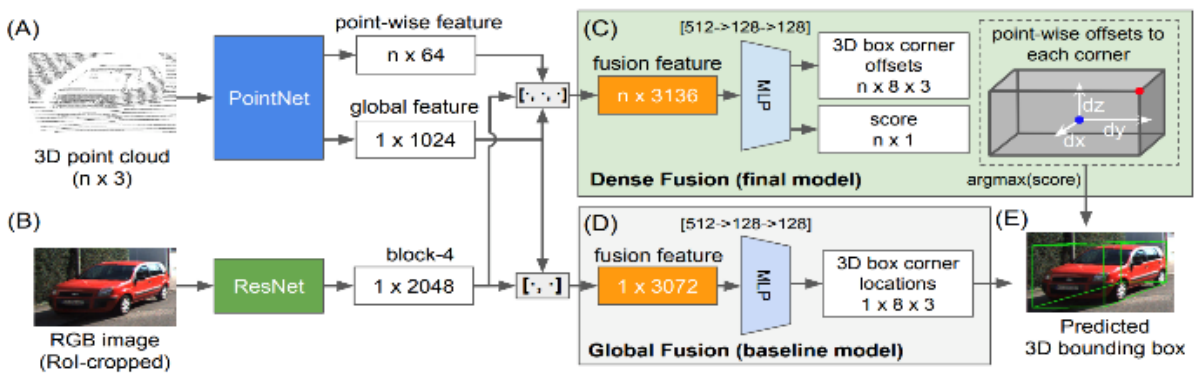

Fig. 10. PointFusion, from reference [213].

Du et al. [214] design a pipeline to adopt any 2D detection network and fuse it with a 3D point cloud to generate 3D information with minimum changes of the 2D detection networks, where 3D bounding box is identified by model fitting based on generalized car models and refined with a two stage CNN method. *RoarNet* [215], proposed by Shin et al., consists of two part, one is RoarNet 2D that estimates the 3D poses of objects and the other one is RoarNet 3D that takes the candidate regions and conducts in-depth inferences to conclude final poses in a recursive manner.

Ku et al. [216] propose a two-stage fusion network, called *AVOD* (Aggregate View Object Detection), in which an RPN is capable of performing multimodal feature fusion on feature maps to generate 3D object proposals and a detection network performs oriented 3D bounding box regression and category classification using these proposals. *Frustum PointNet* [217] applies 2-D object detectors to generate 2D proposals and each 2D region is extruded to a 3D viewing frustum in which to get a point cloud from depth data for 3D bounding box prediction with a light-weight regression PointNet (T-Net).

Liang et al. [218] propose an E2E learnable architecture, exploiting the *continuous fusion layer* to encode both discrete-state image features and continuous geometric information in two streams, which reason in BEV and fuses image features by learning to project them into BEV space. Frossard and Urtasum [219] solve the tracking-by-detection problem with both camera and LIDAR data, formulated as inference in a deep structured model, where detection is achieved by MV3D and association of two detections exploits both appearance and motion via a Siamese network.

*Scale Invariant and Feature Reweighting Network* (SIFRNet) [220] is a 3-D detection method with camera and LiDAR data, which consists of 3D instance segmentation network (Point-UNet) to capture the orientation information, T-Net to extract the feature (reflectivity intensity of the LiDAR) and 3D bounding box estimation network (Point-SENet) to learn the global feature.

*Multimodal VoxelNet* (MVX-Net) [221] can run in two different ways: one is *PointFusion* where points from the LiDAR sensor are projected onto the image plane, followed by image feature extraction from a pre-trained 2D detector and the concatenation of image features and the corresponding points are then jointly processed by VoxelNet; the other is *VoxelFusion* where non-empty 3D voxels created by VoxelNet are projected to the image, followed by extracting image features for every projected voxel using a CNN, which are pooled and appended to the VFE feature encoding for every voxel and further used by the 3D RPN to produce 3D bounding boxes.

A sensor fusion method [222] is extension of LaserNet [123], associates LiDAR points with camera pixels without the requirement of image labels, where the LiDAR as well as camera features are extracted and then concatenated to LaserNet. Ku et al. [223] develop a pedestrian detection method with LiDAR and camera, where a flexible Virtual Multi-View Synthesis module is adopted to improve orientation estimation in an Orientation Prediction Module after depth completion/colorization and 3D object detection.

Raffiee and Irshad [224] extend *AVOD* by proposing a Class-specific Anchoring Proposal (*CAP*) strategy based on object-sizes-and-aspect-ratios-based clustering of anchors in LiDAR and camera. You et al. [225] extend *Pseudo-LiDAR* with stereo camera and extremely sparse LiDAR sensor, called Pseudo-LiDAR++, in which a graph-based depth correction algorithm (GDC) refines the depth map by leveraging sparser LiDAR signal.

*Multi-view Labelling Object Detector* (MLOD) [226] follow the two-stage object detection framework, where the first stage (a RPN) generates 3D proposals in a BEV projection of the point cloud and the second stage projects the 3D proposal bounding boxes to the image as well as BEV feature maps and sends the corresponding map crops to a detection header for classification and bounding-box regression. Li, Liu and Shen [227] use the monocular camera as the fundamental sensor for 2D object proposal and initial 3D bounding box prediction, while the stereo cameras and LiDAR are used to refine the 3D box localization by the stereo photometric alignment or point cloud alignment.

Shown in Table XVIII (from reference [235]), the fusion-based 3D detection performance of different methods with LIDAR and camera is compared on the different benchmarks.

In summary, fusion-based detection methods can work in the volume space or the projection space, apparently the former ones achieve better detection accuracy the latter ones are more computationally efficient. Proposal-based methods can generate either 2-D proposals or 3-D proposals in the two-stage framework. Transform-based methods utilize the relation between BEV and frontal view.

*4) 3-D Segmentation*

Methods of semantic segmentation by multimodal sensors refer to the survey paper [235].

### *I. Human Pose Estimation*

Human pose estimation with deep learning falls into bottom-up and top-down categories，where the bottom-up methods are split into single stage or multi-stage. The deep learning-based methods are also classified as regression-based and detection-based. A survey of these methods is given in [240] (Note: LiDAR is too sparse to analysis the human motion).

## V. MAPPING AND LOCALIZATION

In a SLAM survey paper by Cadena et al. [241], *semantic SLAM* is investigated. Semantic mapping consists in associating semantic concepts to geometric entities in robot's surroundings, where deep learning is applied for semantic object detection and classification.

Milz et al. [242] give an overview of deep learning applications in visual SLAM, from depth estimation, optic flow estimation, feature extraction and matching, loop closure detection/re-localization, semantic segmentation and camera pose estimation. Huang, Zhao and Liu [243] publish an overview of SLAM with LiDAR, camera, IMU and their fusions, in which deep learning methods are investigated in respective sessions too, like feature extraction, object detection, segmentation, moving object removal, pose estimation and localization.

Below we introduce some noticeable work in the SLAM.

*PointVLAD* [244] is a combination/modification of the existing PointNet and NetVLAD, which allows end-to-end training and inference to extract the global descriptor from a given 3D point cloud, solving point cloud-based retrieval for place recognition. *L3-Net* [245] is a learning-based LiDAR localization system, which learns local descriptors with PointNet specifically optimized for matching, constructs cost volumes over the solution space and applies CNNs and RNNs to estimate the optimal pose.

*LF-Net* (Local Feature Network) [246] is proposed by EPFL, where a detector network generates a scale-space score map along with dense orientation estimates to select the keypoints and image patches around the chosen keypoints are cropped with a differentiable sampler (STN) and fed to the descriptor network to generate a descriptor. *SuperPoint* [247], proposed by Magic Leap, is a fully convolutional network trained to detect interest points and compute their accompanying descriptors, which jointly computes pixel-level interest point locations and associated descriptors in one forward pass.

*DeMoN* (Depth and Motion Network) [248] formulates SfM as a learning problem and trains the stacked encoder-decoder networks end-to-end to compute depth and camera motion from successive, unconstrained image pairs, with the surface normal and optic flow output as well. SfM-Net [252] is a geometry-aware NN for motion estimation in videos that decomposes frame pixel motion in terms of scene and object depth, camera motion and 3D object rotations and translations. VINNet [251] is on-manifold sequence-to-sequence learning approach to motion estimation using visual and inertial sensors, which performs fusion of the data at an intermediate feature-representation level in a two stream LSTM (camera and IMU) network.

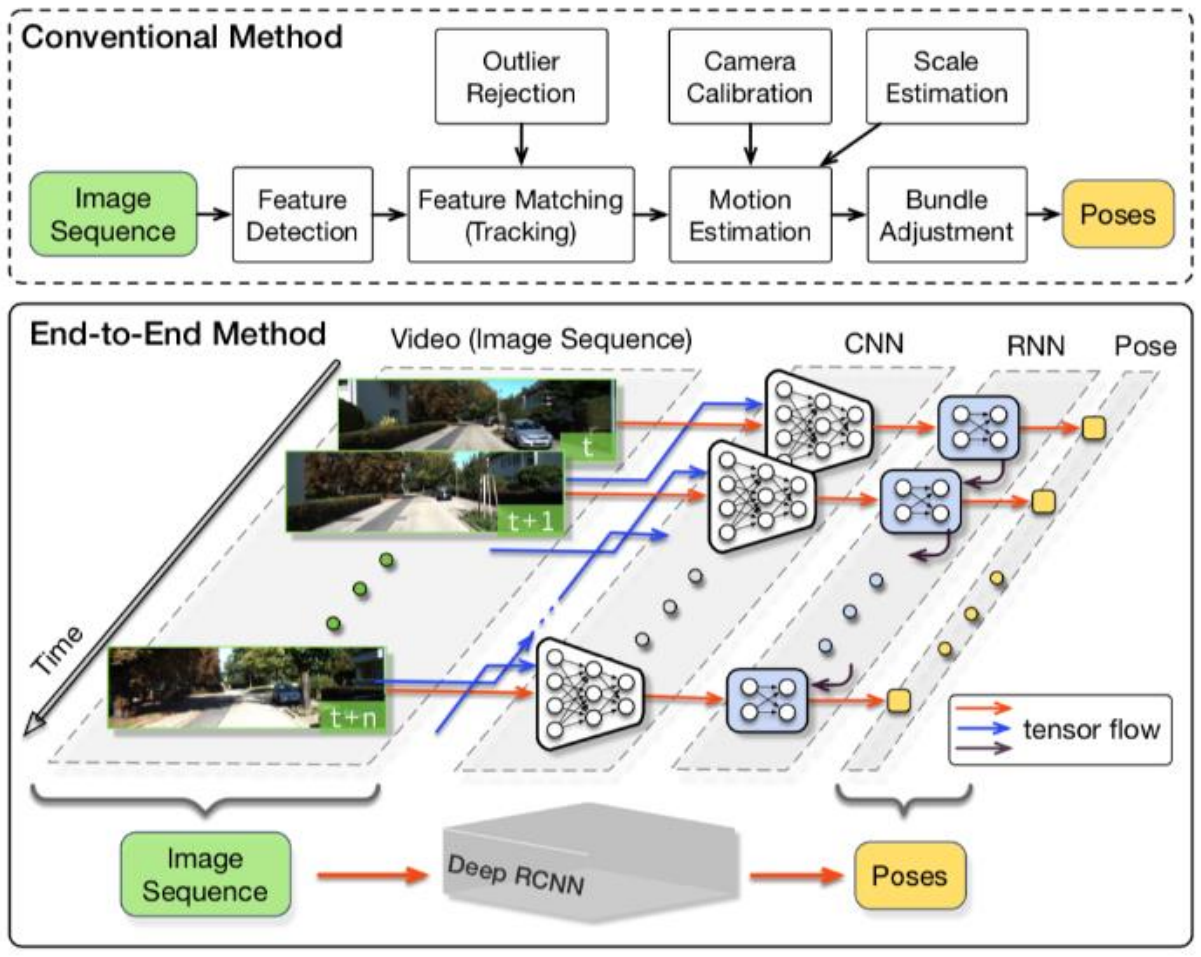

Fig. 11. DeepVO, from reference [250].

UnDeepVO [249] is able to estimate the 6-DoF pose of a monocular camera and the depth of its view by training deep neural networks in an unsupervised manner. DeepVO [250] (shown in Fig. 11) is an end-to-end framework for monocular VO by using *Recurrent Convolutional Neural Networks* (RCNNs), which not only automatically learns effective feature representation through CNN but also implicitly models sequential dynamics and relations using RNN.

CNN-SLAM [263] fuses CNN-predicted dense depth maps naturally together with depth measurements obtained from direct monocular SLAM. Similarly, Deep Virtual Stereo Odometry [254] incorporates deep depth predictions into Direct Sparse Odometry (DSO) as direct virtual stereo measurements. Li et al. [255] propose a self-supervised learning framework for visual odometry (VO) that incorporates correlation of consecutive frames and takes advantage of adversarial learning, where the generator learns to estimate depth and pose to generate a warped target image and the discriminator evaluates the quality of generated image with high-level structural perception.

Zhan et al. [256] explore a way to integrate deep learning with epipolar geometry and Perspective-n-Point (PnP) method, which trains two CNNs for estimating single-view depths and two-view optical flows as intermediate outputs. Zhao et al. [257] propose a self-supervised network architecture for effectively predicting 6-DOF pose, which incorporates the pose prediction into DSO for robust initialization and tracking process.

## VI. PREDICTION, PLANNING AND DECISION MAKING

There are survey papers about human motion prediction and vehicle behaviour prediction [274, 301].

### A. *Human Behaviour Modeling/Prediction*

Pedestrian behaviour modelling can be typically classified as *physics*-based, *pattern*-based and *planning*-based. Most of deep learning-based methods and GAN-based methods are pattern-based, while deep reinforcement learning-based methods are planning-based, shown in Table VI.

TABLE VI. PEDESTRAIN BEHAVIOR PREDICTION METHODS

| RNN/LSTM | GAN | Attention |
|---|---|---|
| Social LSTM [258]<br>Interaction aware motion [259]<br>VRNN [263]<br>SR-LSTM [266]<br>StarNet [267]<br>MATF [268]<br>TraPHic [271]<br>RGM [272]<br>Google Next [273] | Social GAN [260]<br>SoPhie [261]<br>Social Ways [269]<br>IDL [270] | SoPhie [261]<br>Social Attention [262]<br>LVA LSTM[264]<br>ST Attention [265] |

*Social LSTM* [258] regards trajectory prediction as a sequence generation task and proposes an LSTM model to learn general human movement with S-pooling and predict their future trajectories. Another LSTM-based method is proposed by Pfeiffer et al, which incorporates both static obstacles and surrounding pedestrians for trajectory forecasting [259] with interaction aware motion modelling. To emphasize the social characteristics of humans, *Social GAN* [260] predicts socially plausible futures by training adversarially against a recurrent discriminator and encourages diverse predictions with a variety loss, in which LSTM models the temporal pattern.

Another GAN-based method is *SoPhie*, which leverages two sources of information about both physical and social attention, the path history of all the agents in a scene and the scene context information to predict paths [261], based on LSTM as well. Vemula, Muelling and Oh learn the *social attention* model [262] based on a feedforward, fully differentiable, and jointly trained *RNN* mixture which can predict future locations and capture the human-human interactions in a spatial-temporal graph (*ST-graph*). Another RNN framework, proposed by Hoy et al [263], realizes class- specific object tracking and short-term path prediction based on a variant of a *Variational RNN* (VRNN), which incorporates *latent* variables corresponding to a dynamic state space model.

A joint Location-Velocity Attention (LVA) LSTM (location LSTM layer and velocity LSTM layer) is used to model the attention mechanism and then predict trajectories [264]. Another LSTM-based attention modelling method, proposed by Haddad [265], takes into account the interaction with static (physical objects) and dynamic (other pedestrians) elements in the scene and then also applies ST-graph (edge LSTM and node LSTM) for trajectory prediction. A data-driven state refinement module for LSTM network (SR- LSTM) is proposed [266], which activates the utilization of the current intention of neighbours and refines the current states of all participants in the crowd through a message passing mechanism.

*StarNet*, a star topology designed by Meituan researchers [267], includes a unique hub network and multiple host networks based on LSTM, where the hub network generates a description of the interpersonal interactions and the host networks predict future trajectories. A trajectory prediction method, called *Social Ways* [269], applies info-GAN to sample plausible predictions for any agent in the scene with the interaction aware capability. A LSTM-based Multi-Agent Tensor Fusion (MATF) network [268], applied for both pedestrians and vehicles, models the social interaction and scene context constraint jointly.

Li reported his work [270] on Imitative Decision Learning (IDL) with GAN, trying to understand and imitate the underlying human decision-making process to anticipate future paths in dynamic scenes, which first infers the distribution of latent decisions of multimodality by learning from moving histories and then generats a policy by taking the sampled latent decision into account to predict the future. His method integrates spatial and temporal dependencies into one single *encoder-decoder* framework, in contrast to handling them with two-step settings.

*TraPHic* [271] models the interactions between different road agents using a LSTM-CNN hybrid network for trajectory prediction, considering heterogeneous interactions that account for the varying shapes, dynamics, and behaviours of different road agents. Choi and Dariush work on a relation-aware framework for future trajectory forecast [272], which aims to infer relational information from the interactions of road users with each other and with environments by a designed relation gate module (*RGM*).

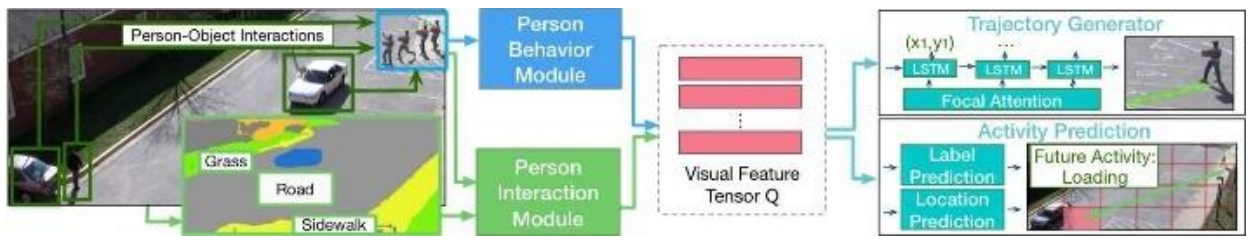

Fig. 12. Google's multi-task learning system Next, from reference [273].

Google proposes a E2E multi-task learning system, called *Next* (shown in Fig. 12), utilizing rich visual features about human behavioural information and interaction with their surroundings [273]; it encodes a person through semantic features about visual appearance, body movement and interaction with the surroundings consisting of person behaviour module, person interaction module, trajectory generator and activity prediction based on CNN and LSTM.

Shown in Table XIX (from reference [267]), the human trajectory prediction performance of different methods is compared on the different benchmarks.

It is seen that pedestrian behaviour modelling methods mostly employ the RNN/LSTM to formulate the temporal pattern, while GAN-based methods boost the prediction model training with adversarial samples. Attention mechanism is modelled in the prediction framework by some methods, which enhance the representation capability of social interaction in the crowded pedestrian activities.

### B. *Vehicle Driving Behaviour Modeling and Decision Making*

Vehicle behaviour prediction models are categorized to *physics*-based, *manoeuvre*-based, and *interaction*-aware models. Roughly the deep learning-based methods are classified based on the model types as CNN, RNN (LSTM, GRU), GAN, GNN and Deep RL/IRL, shown in Table VII.

TABLE VII. VEHICLE BEHAVIOR MODELING AND DECISION MAKING

| CNN-LSTM | GAN/VAE | RL | GNN |
|---|---|---|---|
| Baidu rank-based IRL [275]<br>Berkeley IRL [277]<br>Cui et al. (Deep CNN) [279]<br>TrafficPredict [280]<br>ChauffeurNet [281]<br>CIL [282]<br>INFER [286]<br>Multi-Path [294] | DESIRE [285]<br>AGen (PS-GAIL) [287]<br>Multi-agent [288]<br>CTPS (C-BCGAN) [289]<br>CGNS [290]<br>CVAE + IRL [291]<br>CVAE + STL [292] | Deep RL (DQN) [276]<br>RL-RC [278]<br>MCTC [283]<br>Multi-modal [284] | GRIP [293]<br>Graph conv LSTM [295]<br>Interaction graph [296]<br>Social-WaGDAT [297]<br>EvolveGraph [298]<br>Semantic graph 299]<br>VectorNet [300] |

Baidu Apollo team proposes a rank-based conditional *inverse reinforcement learning* (IRL) algorithm [275] to tune an autonomous driving motion planner, in which the reward functional learning includes two key parts: conditional comparison and rank-based learning based on Siamese network including value networks of both the human and the sampled trajectories. Sun, Zhan and Tomizuka [277] propose a probabilistic prediction approach based on hierarchical IRL method to achieve interaction-aware predictions, where first the distribution over all future trajectories is formulated as a mixture of distributions partitioned by the discrete decisions, then IRL is applied hierarchically to learn the distributions from demonstrations.

Deep Reinforcement Learning (RL) with DQN is used by Wolf et al. [276] to learn manoeuvre decisions based on a compact semantic state representation of all the scene objects, such as vehicles, pedestrians, lane segments, signs and traffic lights, where the state & reward are extended by a behaviour adaptation function and a parameterization respectively. Another deep RL method, proposed by Xu, Tang and Tomizuka, designs a robust-control-based (RC) generic transfer architecture [278], called RL-RC, incorporates a transferable hierarchical RL trajectory planner and a robust tracking controller based on disturbance observer (DOB).

A method [279], proposed by Cui et al., encodes each surrounding traffic actor's surrounding context into a raster BEV image, which is used as input by deep convolutional networks to automatically derive relevant features for predicting multiple possible trajectories. A LSTM-based real-time traffic prediction algorithm on the frontal view, called *TrafficPredict*, is proposed [280], which uses an instance layer to learn instances (vehicles, bicycles, and pedestrians)' movements and interactions and has a category layer to learn the similarities of instances belonging to the same type to refine the prediction.

Google WayMo proposes a CNN-RNN based network, called *ChauffeurNet* [281], to train a policy for autonomous driving via imitation learning, in which it renders the BEV image with information of ego vehicle, obstacles, and traffic rules (traffic light/sign, speed limit and navigation route) and augments the imitation loss with additional losses that penalize undesirable events and encourage progress. Another behaviour cloning method uses an expanded formulation, called *Conditional Imitation Learning* (CIL), where a ResNet perception module processes an input image to a latent space followed by two prediction heads: one for steering and one for speed [282].

Hoel et al. [283] formulate the decision-making problem for the two highway driving cases was formulated as a *POMDP* (partially observable Markov decision process) and then combine the concepts of planning and learning in the form of *Monte Carlo tree search* (MCTC) and deep RL, like AlphaGo. Moghadam and Elkaim [284] develop a multi-modal architecture that includes the environmental modelling of ego vehicle surrounding in stochastic highway driving scenarios and train a deep RL agent in decision making.

Lee et al. 's work [285] is called DESIRE, DEep Stochastic *Inverse optimal control* (IOC) RNN Encoder- decoder, in which *conditional variational auto-encoder* (CVAE) is employed to obtain hypothetical future prediction samples, ranked and refined by the following RNN *scoring-regression* module, and a RNN *scene context fusion* module jointly captures past motion histories, the semantic scene context and interactions among multiple agents in dynamic scenes.

*INFER* (INtermediate representations for FuturE pRediction) is defined by Srikanth et al. [286], which relies on semantics (by joint stereo-LiDAR segmentation) other than texture information and train an AR model to predict future trajectories of traffic participants (vehicles) based on encoder-decoder and LSTM. *AGen* (An adaptable Generative prediction framework) is put forward by Si, Wei and Liu [287], to perform online adaptation of the offline learned models to recover individual differences for better prediction, where the *recursive LS parameter adaptation algorithm* (RLS-PAA) is combined with the offline learned model from the imitation learning method, *parameter sharing generative adversarial imitation learning* (PS-GAIL).

Based on GAN, Li, Ma and Tomizuka from UC Berkeley propose a generic multi-agent probabilistic prediction and tracking framework [288] which takes the interactions among multiple entities into account, in which all the entities are treated as a whole. In order to tackle the task of probabilistic prediction for multiple, interactive entities, soon later they propose a *coordination and trajectory prediction system* (CTPS) [289], which has a hierarchical structure including a macro-level coordination recognition module based on a *variational RNN* followed by a probabilistic classifier and a micro-level subtle pattern prediction module which solves a probabilistic generation task, based on a Coordination-Bayesian Conditional Generative Adversarial Network (C-BCGAN). They also propose a *conditional generative neural system* (CGNS) in [290] for probabilistic trajectory prediction to approximate the data distribution, with which realistic, feasible and diverse future trajectory hypotheses can be sampled.

Two prediction models are typically used: learning-based model and planning-based model. Hu, Sun and Tomizuka from Berkeley [291] leverage the advantages from both prediction models and aim at predicting the behaviour of the selected vehicle while considering the potential influence of its future behaviour from its own vehicle, in which the learning-based method applies conditional variational autoencoder (CVAE) and the planning-based method stems from *Theory of Mind* employing continuous domain maximum entropy IRL.

Cho et al. publish their work [292] on jointly reasoning both a future trajectories of vehicles and degree of satisfaction of each traffic rule in the deep learning framework consisting of 4 modules as encoder (LSTM-based), interaction (CVAE-based), prediction (LSTM-based) and control, while a rule is represented as a *signal temporal logic* (STL) formula and a robustness slackness, a margin to the satisfaction of the rule.

A method of Graph-based Interaction-aware Trajectory Prediction (*GRIP*) is proposed [293], which uses a graph to represent the interactions of close objects, applies several graph convolutional blocks to extract features, and subsequently uses an encoder-decoder LSTM model to make predictions.

An improved work on imitation learning over ChauffeurNet from Google WayMo, called MultiPath [294], leverages a fixed set of future state-sequence anchors that correspond to modes of the trajectory distribution, where at inference the model predicts a discrete distribution over the anchors and, for each anchor, regresses offsets from anchor waypoints along with uncertainties, yielding a Gaussian mixture at each time step.

Lee et al. [295] propose a GNN that jointly predicts the discrete interaction modes and 5-second future trajectories for all traffic agents in the scene, which infers an interaction graph with nodes as agents and edges to get the long-term interaction intents among the agents. Another traffic forecasting method [296], proposed by Chandra et al., applies a two-stream *graph convolutional LSTM network* with dynamic weighted traffic-graphs to model the proximity between the road agents, where the first stream predicts the spatial coordinates of road-agents, and the second stream predicts whether a road-agent is going to exhibit aggressive, conservative, or normal behaviour.

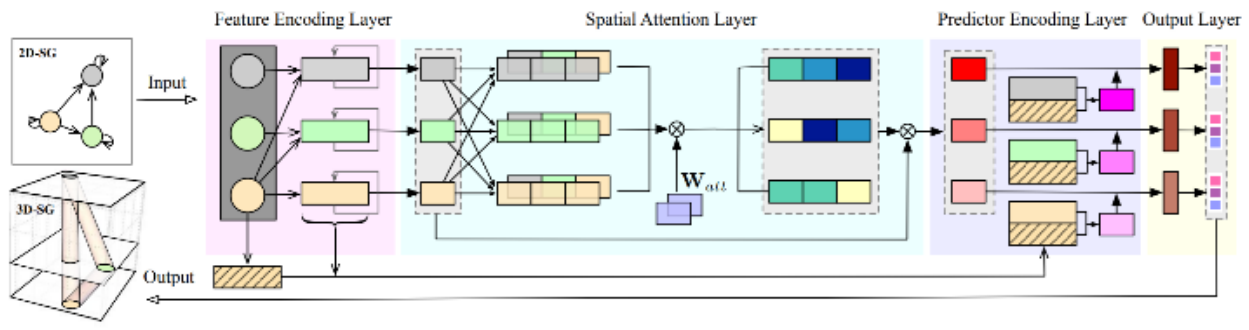

Fig. 13. Berkeley's Semantic Graph Network, from reference [299].

Three papers from UC Berkeley apply GNN to make trajectory prediction: 1) a generic generative neural system (called *Social-WaGDAT*) [297], which makes a step forward to explicit interaction modelling by incorporating relational inductive biases with a dynamic graph representation and leverages both trajectory and scene context information; 2) a generic framework (named *EvolveGraph*) [298] with explicit interaction modelling via a latent interaction graph among multiple heterogeneous, interactive agents, which is trained with static graph learning stage and dynamic graph learning stage respectively; 3) a scenario-transferable and interaction-aware probabilistic prediction algorithm based on *semantic graph reasoning* [299] (shown in Fig. 13), where generic representations for various environment information considering road topological elements, traffic regulations and the defined *dynamic insertion areas* (DIA) as building blocks to construct their spatio-temporal structural relations.

The recent work from Google WayMo, called VectorNet [300], is a hierarchical GNN that exploits the spatial locality of individual road components represented by vectors and then models the high-order interactions among all components, by operating on the vectorized high definition (HD) maps and agent trajectories.

Shown in Table XX (from reference [285]), the vehicle prediction performance of different methods is compared on the Kitti and SDD datasets.

In summary, CNN-LSTM is still the dominant model for modelling the driving behaviour spatial-temporal pattern, while GAN/VAE utilizes the generative learning to capture the interaction behaviour globally. Deep RL methods show the online adaptation capability to jointly learn the prediction and planning. Recently graph neural network become noticeable in modelling interaction efficiently.

## VII. Control

There is a survey paper [302] about vehicle control by deep learning. These methods are roughly classified as reinforcement learning and supervised learning (simulation learning and inverse reinforcement learning), and the control styles fall into lateral control, longitudinal control and their joint control.

## VIII. End-to-End System

Besides of modular autonomous driving systems, there are some platform working in an end-to-end manner. They are either the entire loop from perception to control, loop from planning to control (without perception), or loop from perception to planning (without control).

Bojarski et al. [303] design an e2e trainable architecture for learning to predict a distribution over future vehicle ego-motion from monocular camera observations and previous vehicle state with FCN and LSTM.

Santana and Hotz [304] from comma.ai report work on learning to clone driver behaviours and planning manoeuvres by simulating future events in the road, by variational autoencoders (VAE) and RNN.

*DeepTest* [305] is a systematic testing tool for automatically detecting erroneous behaviours of DNN-driven vehicles that can potentially lead to fatal crashes, which is designed to automatically generated test cases leveraging real-world changes in driving conditions.

Chen et al. [307] regard the model driving process as a continuous prediction task, i.e. to train a DNN-LSTM model that receives multiple perception information including video frames and point clouds within PointNet and Point Cloud Mapping, and to predict correct steering angles and vehicle speeds.

Hecker, Dai and Gool [306] from ETH propose a method that learns to predict the occurrence of driving failures with CNN-LSTM, where the prediction method is able to improve the overall safety of an automated driving model by alerting the human driver timely. They also propose to learn a driving model by integrating info. from the surround-cameras, the route planner and a CAN bus reader [308] (shown in Fig. 14), where the driving model consists of CNN networks for feature encoding, LSTM networks to integrate the outputs of the CNNs over time and FCN to integrate info. from multiple sensors to predict the driving manoeuvres.

Grigorescu et al. [309] run a neuroevolutionary approach, called *NeuroTrajectory*, to local state trajectory learning for autonomous vehicles, estimated over a finite prediction horizon by a perception-planning deep NN (CNN plus LSTM), different from the pure end-to-end method or the perception-planning-control method.

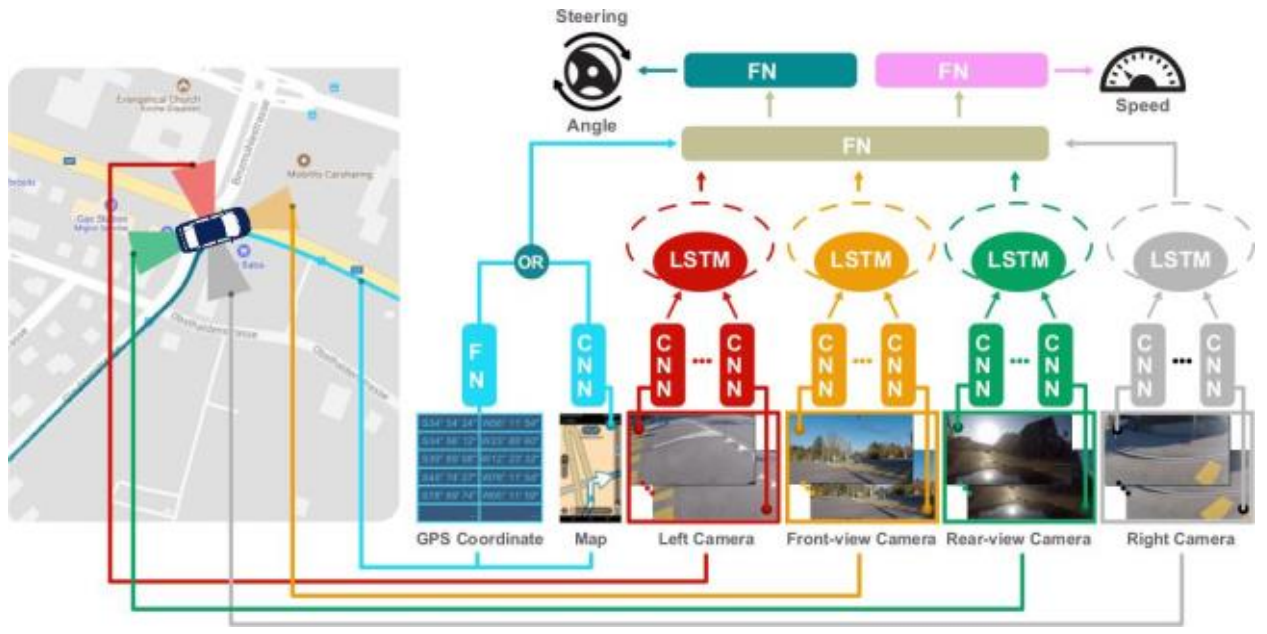

Fig. 14. ETH's E2E driving system, from reference [308].

## IX. SIMULATION

Deep learning applications for simulation for autonomous driving mostly fall into sensor modelling. In this session we will introduce some papers in radar modelling, LiDAR model and image/video synthesis.

Wheeler et al. [310] propose a methodology for the construction of stochastic automotive radar models based on deep learning with adversarial loss connected to real-world data, in which the inputs to the network are the spatial raster (a 3D tensor with two primary dimensions of range-azimuth and the third dimension consisting of layers that are dedicated to different types of information) and the object list (also represented as a 3D tensor), and the output is sensor reading of the virtual radar, represented by a normal distribution or a GMM.

Yue et al. [313] present a framework to rapidly create point clouds with accurate point-level labels from a computer game, in which point clouds from auto-driving scenes can be used as training data for deep learning algorithms, such as *SqueezeSeg* used for point cloud segmentation. Elmadawi et al. [314] propose a deep Learning-based LiDAR sensor model to learns the complex function mapping from physical environmental setup to LiDAR data physical properties, where a DNN models echo pulse widths learned from real data using Polar Grid Maps (PGM).

Alhaija et al. [311] propose a paradigm which augments real-world imagery with virtual objects of the target category, where an efficient procedure creates realistic composite images which exhibit both realistic background appearance and a large number of complex object arrangements.

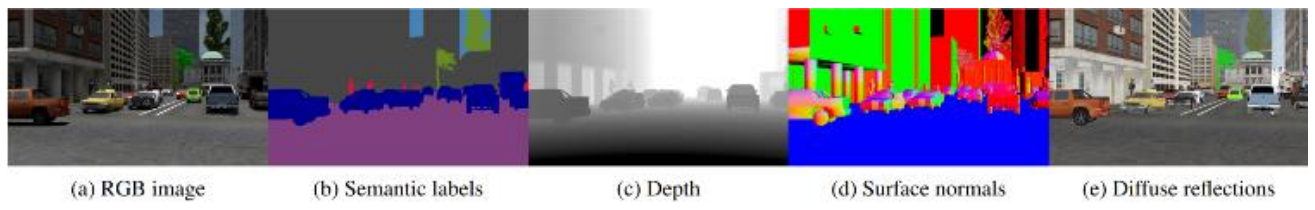

Fig. 15. Adversarially Tuned Scene Generation [312].

Veeravasarapu, Rothkopf and Visvanathan [312] combine recent advances in CG and generative adversarial training (shown in Fig. 15), using iterative estimation of the posterior density of prior distributions for a graphical model, as a scene generative model combined with static/dynamic objects, light source model, weather model, ground model, camera model and road network.

## X. V2X

There are two deep learning application survey papers, one for mobile and wireless networking [315], one for V2X [316]. In networking applications, deep learning can be used for network optimization, routing, scheduling, resource allocation and radio control etc. In V2X, deep learning could be used for network congestion control, security in VANETs, vehicular edge computing, content delivery/offloading and vehicle platoons etc.

For the vehicle side in V2X, deep learning methods are used for multi-agent SLAM, collaborative perception and planning.

## XI. SAFETY

Safety in AI applications is a new and challenging topic. The main issue could include the interpretability, uncertainty, verification and attack-defence etc.

McAllister et al. [317] investigate three under-explored themes for autonomous driving (AV) research: safety, interpretability, and compliance. *Safety* can be improved by quantifying the uncertainties of component outputs. *Interpretability* is concerned with explaining what the AV observes. *Compliance* refers to maintaining some control for humans. A principled approach to modelling uncertainty is Bayesian probability theory, so they propose to use Bayesian deep learning by either placing distributions over model weights, or by learning a direct mapping to probabilistic outputs.

VerifAI [318] is a software toolkit from UC Berkeley, for the formal design and analysis of systems that include AI and machine learning components. It particularly seeks to address challenges with applying formal methods to perception and machine learning components in the presence of environment *uncertainty*. VerifAI provides users with SCENIC, a probabilistic programming language for modelling environments, combined with a renderer or simulator for generating sensor data, can produce semantically consistent input for perception components. Therefore, like V-model in the software development process, VerifAI could design and develop the deep learning models to solve the perception problems with uncertainty handling.

Jha et al. present a machine learning-based fault injection engine [319], called *DriveFI* (shown in Fig. 16), which can mine situations and faults that maximally impact autonomous driving safety, as demonstrated on two industry-grade autonomous driving technology stacks (from NVIDIA and Baidu). *DriveFI* leverages the existing tools to simulate driving scenarios and control the vehicle in simulation by using an AI agent.

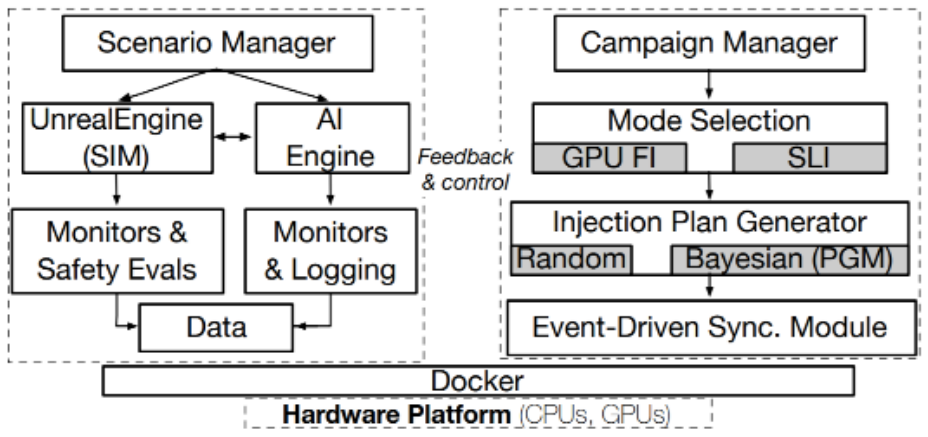

Fig. 16. DriveFI architecture, from reference [319].

Huang et al. [320] report a survey paper, which conducts a review of the current research effort into making DNNs safe and trustworthy, by focusing on four aspects: *verification, testing, adversarial attack and defence, and interpretability*. Existing approaches on the *verification* of networks largely fall into the following categories: constraint solving, search based approach, global optimisation, and over-approximation. They survey *testing* techniques from three aspects: coverage criteria, test case generation, and model level mutation testing. The goal of *attack* techniques is to give adversarial examples for the lack of robustness of a DNN. *Defence* techniques are dual to attack techniques, by either improving the robustness of the DNN, or differentiating the adversarial examples from the correct inputs. Eventually, they review three categories of *interpretability* methods: instance-wise explanation, model explanation, and information theoretical methods.

## XII. OPEN SOURCES AND DATASETS

There are some open sources of autonomous driving developments: Baidu Apollo [321], Autoware [322], OpenPilot [323] and Udacity [324]. Open source simulators to assist autonomous driving development include Intel Carla [326], MS AirSim [325] and LGSVL simulator [327]. There are some traffic network simulation tools, such as SUMO [328]. A framework is developed by UC Berkeley, called FLOW [329], for designing custom traffic scenarios and integration with deep reinforcement learning and traffic microsimulation libraries. Besides, some companies provide visualization tools, like GM Cruise.AI [330] and Uber [331].

There are a number of open data sources (i.e. sensor data, including cameras, LiDAR, radar, GPS/IMU, wheel encoder and so on) in autonomous driving communities, like Kitti [332], Udacity [336], NuScenes [335], Waymo [338], Lyft Level5 [339], BaiduScope [334], BDD (Berkeley) [333], ArgoVerse [337] and PandaSet [341] etc., whose comparison in scenes, sensors, weathers, locations, classes, time size (hours), object bounding boxes and map layers, are shown in Table VIII.

Some datasets are open for trajectory-based autonomous driving research (their comparisons in scenarios, behaviours, maps and interactions are shown in Table IX), like NGSim [342], HighD [343] and INTERACTION [344] (shown in Fig. 17).

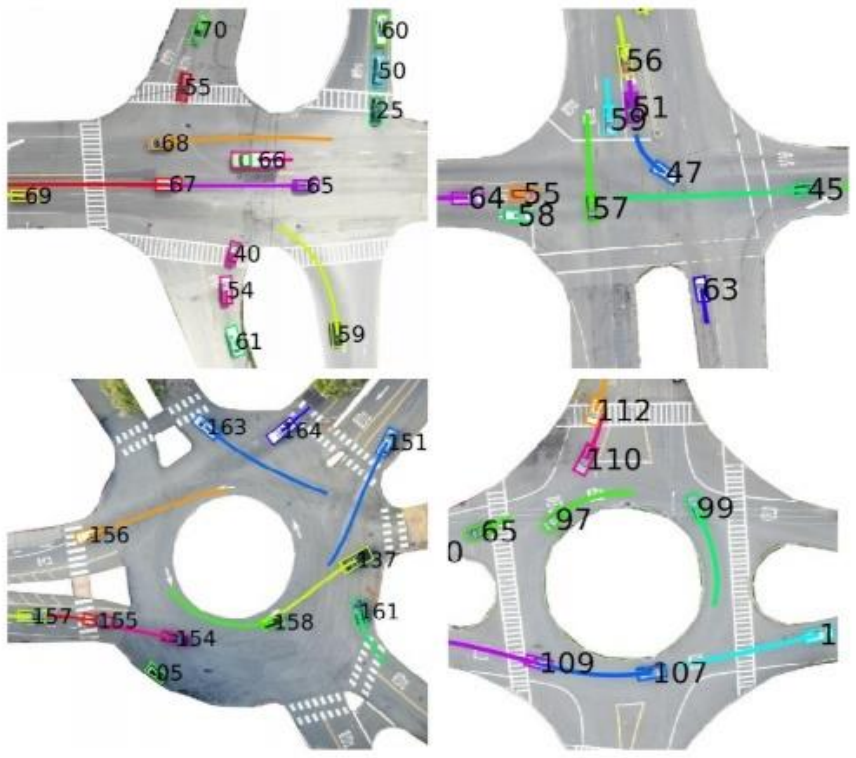

Fig. 17. Detection & tracking examples in the INTERACTION dataset [344].

## XIII. Conclusion

We have investigated state-of-art deep learning methods applied for autonomous driving in several major areas. It is seen the marriage of them has made impressive and promising accomplishments. However, there are still some challenges in this field, due to either the autonomous diving task itself or the deep learning shortcomings [345-346], listed as follows.

- *Perception*. The long tail effect is obvious and there are corner cases to find. To train a deep learning model still requires a lot of data, while model overfitting and sensitivity of image changes are still bother us [346-347]. Sensor fusion (including V2X as well) is the requisite for perception and localization, especially in bad weather conditions, but modelling of each sensor's characteristics (capacity and limitation) for these corner cases is not well defined.
- *Prediction*. The vehicle or pedestrian trajectory prediction needs more data to train the model, while behaviour and intention modelling is still lacking, for both short-term and long-term predictions. More clues are required to include into the model, extracted from the perception, like human's gaze and pose, drivers' emotion and hand gesture, and vehicles' turn light signal etc.
- *Planning and control*. Behaviour planning and motion planning are unmatured in deep learning's application, especially real time implementation in crowded and highly dynamic traffic situations. How to define the driving scenarios and imitate the driving actions from demonstrations are still challenging. Especially, collaborative planning based on V2X is still a complicated problem, either centralized or decentralized.
- *Safety*. Uncertainty of deep learning for autonomous driving is still an open problem [350-351], in either data uncertainty or model uncertainty. Based on GAN's application work [348-349], it is seen some adversarial cases are not easily handled. Though interpretability of deep learning is paid more attention to and simulation helps a lot in new algorithms' testing, how to find corner cases is not figured clearly. Fault detection for each module is critical for autonomous driving development, for fail-safe and fail-operational in L3/L4.
- *Computing platform*. It is not clear the computation power request for autonomous driving is calculated out, especially for the planning and control modules in L3-L4-L5, though there are not a few companies developing stronger SoCs and accelerators for autonomous driving vehicles. Meanwhile, the vehicle-road edge-cloud combination is still new to figure out a computing solution for autonomous driving.

## References

[1] S. Grigorescu et al., A Survey of Deep Learning Techniques for Autonomous Driving, arXiv 1910.07738, Mar. 2020

[2] B Ravi Kiran et al., Deep Reinforcement Learning for Autonomous Driving: A Survey, arXiv 2002.00444, Feb. 2020

[3] Y. LeCun, L. Bottou, Y. Bengio, P. Haffner. Gradient-based learning applied to document recognition. *Proceedings of the IEEE*, 1998

[4] Gers F A, and Schmidhuber J. Recurrent nets that time and count. Neural Networks, 2000.

[5] Hinton G, Osindero S, The Y, A fast learning algorithm for deep belief nets, Neural Computation, July 2006

[6] Krizhevsky, A., Sutskever, I. & Hinton, G. E. ImageNet Classification with Deep Convolutional Neural Networks. Adv. Neural Inf. Process. Syst. 2012.

[7] Lin M, Chen C, and Yan S. Network in network. arXiv 1312.4400, 2013.

[8] Chatfield K., Simonyan K., Vedaldi A., and Zisserman A. Return of the devil in the details: Delving deep into convolutional nets. arXiv 1405.3531, 2014

[9] Szegedy, C., Liu, W. & Jia, Y. Going deeper with convolutions. arXiv 1409.4842, 2014.

[10] Kingma D P. and Welling M. Stochastic gradient VB and the variational auto-encoder. Int. Conf. on Learning Representations, ICLR. 2014.

[11] Goodfellow I et al. Generative adversarial nets. Advances in neural information processing systems. 2014.

[12] Cho K et al., Learning Phrase Representations using RNN Encoder-Decoder for Statistical Machine Translation, arXiv 1406.1078, 2014.

[13] Y LeCun, Y Bengio, and G Hinton. Deep Learning, Nature 521, no. 7553, 2015.

[14] I Goodfellow et al., Deep Learning, *MIT press*, 2016.

[15] Chollet F. Xception: Deep Learning with Depthwise Separable Convolutions. arXiv 1610.02357, 2016

[16] He, K., Zhang, X., Ren, S. & Sun, J. Deep residual learning for image steganalysis. arXiv 1512.03385, 2017

[17] Huang, G., Liu, Z., Van Der Maaten, L. & Weinberger, K. Q. Densely connected convolutional networks. IEEE CVPR 2017

[18] Hu, J., Shen, L., Albanie, S., Sun, G. & Wu, E. Squeeze-and-Excitation Networks. arXiv 1709.01507, 2017.

[19] Li X et al., Selective Kernel Networks. arXiv 1903.06586, 2019

[20] Zhang H et al., ResNeSt: Split Attention Networks, arXiv 2004.08955, 2020

[21] Y Cheng et al., A Survey of Model Compression and Acceleration for Deep Neural Networks, arXiv 1710.09282, Oct. 2017

[22] V Sze et al., Efficient Processing of Deep Neural Networks: A Tutorial and Survey, *Proc. of IEEE*, 8, 2017

[23] M Alom et al., The History Began from AlexNet: A Comprehensive Survey on Deep Learning Approaches, arXiv 1803.01164, 2018

[24] Tan C et al., A Survey on Deep Transfer Learning, arXiv 1808.01974, 2018

[25] Vanshoren J, Meta-Learning: A Survey, arXiv 1810.03548, 2018

[26] Y Li, Deep Reinforcement Learning, arXiv 1810.06339, 2018

[27] T Ben-Nun, T Hoefler, Demystifying Parallel and Distributed Deep Learning: An In-Depth Concurrency Analysis, *ACM Computing Surveys*, 9, 2018

[28] S Pouyanfar et al., A Survey on Deep Learning: Algorithms, Techniques, and Applications, *ACM Computing Surveys*, Vol. 51, No. 5, 9, 2018

[29] Shorten C, Khoshgoftaar T M, A Survey on Image Data Augmentation for Deep Learning, *J. of Big Data*, 6, 60, 2019

[30] Sun R, Optimization for deep learning: theory and algorithms, arXiv 1912.08957, Dec. 2019

[31] Wistuba, M., Rawat, A., Pedapati, T. A Survey on Neural Architecture Search. arXiv 1905.01392, 2019

[32] He X, Zhao K, Chu X, AutoML: A Survey of State-of-Art. arXiv 1908.00709, 2019

[33] Z Wu et al., A Comprehensive Survey on Graph Neural Networks, arXiv 1901.00596, Dec. 2019

[34] Chen J, Ran X, Deep Learning with Edge Computing: A Review, Proc. IEEE, July 2019

[35] Wang X et al., Convergence of Edge Computing and Deep Learning: A Comprehensive Survey, arXiv 1907.08349, Jan. 2020

[36] Kairouz P et al., Advances and Open Problems in Federated Learning, arXiv 1912.04977, Dec. 2019

[37] Y Wang et al., Generalizing from a Few Examples: A Survey on Few-Shot Learning arXiv 1904.05046, May 2019

[38] W Wang et al., A Survey of Zero-Shot Learning: Settings, Methods, and Applications, *ACM Trans. Intell. Syst. Technol*. 10, 2, January 2019

[39] M Li et al., The Deep Learning Compiler: A Comprehensive Survey, arXiv 2002.03794, April 2020

[40] Wang Z, Shi Q, Ward T E, Generative Adversarial Networks in Computer Vision: A Survey and Taxonomy, arXiv 1906.01529, 2020.

[41] A Howard et al., MobileNets: Efficient Convolutional Neural Networks for Mobile Vision Applications, arXiv 1704.04861， 2017

[42] M Sandler et al., MobileNetV2: Inverted Residuals and Linear Bottlenecks, arXiv 1801.04381, 2018

[43] A Howard et al., Searching for MobileNetV3, arXiv 1905.02244, 2019

[44] Tan M, Le Q V., EfficientNet: Rethinking Model Scaling for Convolutional Neural Networks, arXiv 1905.11946, 2020

[45] D González et al. A Review of Motion Planning Techniques for Automated Vehicles, *IEEE Trans. Intelligent Transportation Systems* 17.4 (2016): 1135-1145.

[46] Paden, Brian, et al. A survey of motion planning and control techniques for self-driving urban vehicles. *IEEE Transactions on intelligent vehicles* 1.1 (2016): 33-55.

[47] G Bresson, Z Alsayed, L Yu, S Glaser. Simultaneous Localization and Mapping: A Survey of Current Trends in Autonomous Driving. *IEEE Transactions on Intelligent Vehicles*, 2017

[48] Pendleton S D et al., Perception, Planning, Control, and Coordination for Autonomous Vehicles, *Machines*, 5, 6, 2017

[49] J Rios-Torres, A Malikopollos, A Survey on the Coordination of Connected and Automated Vehicles at Intersections and Merging at Highway On-Ramps, *IEEE T-ITS*, May, 2017

[50] R Salay et al. An Analysis of ISO 26262: Using Machine Learning Safely in Automotive Software， arXiv 1709.02435， 2017

[51] Janai J et al., Computer Vision for Autonomous Vehicles: Problems, Datasets and State-of-the-Art, arXiv 1704.05519, April 2017

[52] Brummelen J V et al., Autonomous vehicle perception: The technology of today and tomorrow, *Transportation Research Part C*, 89, Elsevier Ltd, 2018

[53] Dixit et al., Trajectory planning and tracking for autonomous overtaking: State-of-the-art and future prospects， *Annual Reviews in Control*， March 2018

[54] W Schwarting, J Alonso-Mora, D Rus, Planning and Decision-Making for Autonomous Vehicles, *Annual Review of Control, Robotics and Autonomous Systems*, 1, 2018

[55] Kuutti S et al., A Survey of the State-of-the-Art Localisation Techniques and Their Potentials for Autonomous Vehicle Applications, IEEE Internet of Things Journal, 5(20), 2018

[56] F Jameel et al., Internet of Autonomous Vehicles: Architecture, Features, and Socio-Technological Challenges, arXiv 1906.09918, June 2019

[57] Badue C et al., Self-Driving Cars: A Survey, arXiv 1901.04407, 2019

[58] Yurtsever E, Lambert J, Carballo A, Takeda K. A Survey of Autonomous Driving: Common Practices and Emerging Technologies, arXiv 1906.05113, 2020

[59] Wang Z, Chen J, Hoi S C.H., Deep Learning for Image Super-resolution: A Survey, arXiv 1902.06068, 2019

[60] Tian C et al., Deep Learning on Image Denoising: An Overview, arXiv 1912.13171, 2020

[61] R. Girshick et al., Region based convolutional networks for accurate object detection and segmentation. *IEEE T-PAMI*, 2015

[62] He K. et al. Spatial pyramid pooling in deep convolutional networks for visual recognition. *European Conference on Computer Vision*. 2014

[63] R. Girshick. Fast R-CNN. *ICCV* 2015

[64] S. Ren et al., Faster R-CNN: Towards real-time object detection with region proposal networks. *NIPS*, 2015.

[65] W. Liu et al., SSD: Single shot multi-box detector. arXiv 1512.02325, 2015.

[66] J. Redmon et al., You only look once: Unified, real-time object detection. IEEE CVPR, 2016.

[67] T.-Y. Lin et al., Feature pyramid networks for object detection. IEEE CVPR, 2017.

[68] J. Redmon, A. Farhadi. YOLO9000: Better, faster, stronger. IEEE CVPR, 2017.

[69] X Wang et al., A-Fast-RCNN: Hard Positive Generation via Adversary for Object Detection, arXiv 1704.03414, 2017

[70] T-Y Lin et al., Focal Loss for Dense Object Detection, arXiv 1708.02002, 2017.

[71] J Redmon, A Farhadi, YOLOv3: An Incremental Improvement, arXiv 1804.02767, 2018

[72] Wang R., Li X, Ling C., Pelee: A Real-Time Object Detection System on Mobile Devices, arXiv 1804.06882, 2018

[73] Lee Y et al., An Energy and GPU-Computation Efficient Backbone Network for Real-Time Object Detection, arXiv1904.09730, 2019

[74] Tan M, Pang R, Le Q, EfficientDet: Scalable and Efficient Object Detection, arXiv 1911.09070, 2019

[75] Bochkovskiy A, Wang C-Y, Liao H-Y M. YOLO v4: Optimal Speed and Accuracy of Object Detection, arXiv 2004.10934, 2020

[76] Zou Z et al., Object Detection in 20 Years: A Survey. arXiv 1905.05055, 2019

[77] L Jiao et al., A Survey of Deep Learning-based Object Detection, arXiv 1907.09408, Oct. 2019

[78] H Law, J Deng, CornerNet: Detecting Objects as Paired Keypoints, arXiv 1808.01244, 2018

[79] X Zhou et al., Bottom-up Object Detection by Grouping Extreme and Center Points, arXiv 1901.080432, 2019

[80] C Zhu, Y He, M Savvides. Feature Selective Anchor-Free Module for Single-Shot Object Detection. arXiv 1903.00621, 2019.

[81] Z Tian et al., FCOS: Fully Convolutional One-Stage Object Detection. arXiv 1904.01355, 2019.

[82] K Tao et al., FoveaBox: Beyond Anchor-based Object Detector. arXiv 1904.03797, 2019.

[83] J Wang et al., Region proposal by guided anchoring. arXiv 1901.03278, 2019

[84] X Zhou, D Wang, P Krähenbühl, Objects as Points, arXiv 1904.07850, 2019

[85] K Duan et al., CenterNet: Object Detection with Keypoint Triplets, arXiv 1904.08189, 2019

[86] H Law et al., CornerNet-Lite: Efficient Keypoint Based Object Detection, arXiv 1904.08900, 2019

[87] W Liu et al., Center and Scale Prediction: A Box-free Approach for Object Detection, arXiv 1904.02948, 2019

[88] Cheng Z et al., Segmentation is All You Need, arXiv 1904.13300, 2019

[89] Zhang S et al., Bridging the Gap Between Anchor-based and Anchor-free Detection via Adaptive Training Sample Selection, arXiv 1912.02424, 2020

[90] Dong Z et al., CentripetalNet: Pursuing High-quality Keypoint Pairs for Object Detection, arXiv 2003.09119, 2020

[91] Zhu Z et al., Traffic-Sign Detection and Classification in the Wild, IEEE CVPR 2016.

[92] Lee S et al., VPGNet: Vanishing Point Guided Network for Lane and Road Marking Detection and Recognition, arXiv 1710.06288, Oct. 2017

[93] Mueller J and Dietmayer K, Detecting Traffic Lights by Single Shot Detection, arXiv 1805.02523, Oct. 2018

[94] Pan X, Shi J, Luo P, Wang X, Tang X. Spatial as Deep: Spatial CNN for Traffic Scene Understanding, *AAAI*, 2018

[95] Li B, Zhang T, Xia T, Vehicle Detection from 3D Lidar Using FCN, arXiv 1608.07916, 2016

[96] Zhou Y, Tuzel O, VoxelNet: End-to-End Learning for Point Cloud Based 3D Object Detection, arXiv 1711.06396, 2017

[97] Qi C et al., PointNet: Deep Learning on Point Sets for 3D Classification and Segmentation, arXiv 1612.00593, 2017

[98] Qi C et al., PointNet++: Deep Hierarchical Feature Learning on Point Sets in a Metric Space, arXiv 1706.02413, 2017

[99] Wirges S et al, Object Detection and Classification in Occupancy Grid Maps using Deep Convolutional Networks, arXiv 1805.08689, 2018

[100] Zeng Y et al., RT3D: Real-Time 3-D Vehicle Detection in LiDAR Point Cloud for Autonomous Driving, *IEEE RA Letters*, 2018

[101] Beltran J. et al., BirdNet: a 3D Object Detection Framework from LiDAR information, arXiv 1805.01195, 2018

[102] Minemura K et al., LMNet: Real-time Multiclass Object Detection on CPU using 3D LiDAR, arXiv 1805.04902, 2018

[103] Yang B, Liang M, Urtasun R, HDNET: Exploit HD Maps for 3D Object Detection, *CoRL* 2018.

[104] Yang Z et al., IPOD: Intensive Point-based Object Detector for Point Cloud, arXiv 1812.05276, 2018

[105] Yang B, Luo W, Urtasun R, PIXOR: Real-time 3D Object Detection from Point Clouds, arXiv 1902.06326, 2019

[106] Asvadi A et al., DepthCN: Vehicle Detection Using 3D-LIDAR and ConvNet, *IEEE ITSC*, 2017

[107] Yan Y, Mao Y, Li B, SECOND: Sparsely Embedded Convolutional Detection, *Sensors*, 18, 3337, 2018

[108] Ali W et al. YOLO3D: E2E RT 3D Oriented Object Bounding Box Detection from LiDAR Point Cloud, arXiv 1808.02350, 2018

[109] Simon M et al, Complex-YOLO: An Euler-Region-Proposal for Real-time 3D Object Detection on Point Clouds, arXiv 1803.06199, 2018

[110] Sallab A E et al.,YOLO4D: A ST Approach for RT Multi-object Detection and Classification from LiDAR Point Clouds, *NIPS Workshop*, 2018

[111] Vaquero V et al., Deconvolutional Networks for Point-Cloud Vehicle Detection and Tracking in Driving Scenarios, arXiv 1808.07935, 2018

[112] Luo W, Yang B, Urtasun R, Fast and Furious: Real Time E2E 3D Detection, Tracking and Motion Forecasting with a Single Convolutional Net, IEEE CVPR 2018

[113] Shi S, Wang X, Li H, PointRCNN: 3D Object Proposal Generation and Detection from Point Cloud, arXiv 1812.04244, 2018

[114] Lang A et al., PointPillars: Fast Encoders for Object Detection from Point Clouds, IEEE CVPR 2019

[115] Zhou J et al., FVNet: 3D Front-View Proposal Generation for Real-Time Object Detection from Point Cloud, arXiv 1903.10750, 2019

[116] Shi S et al., Part-A^2 Net: 3D Part-Aware and Aggregation Neural Network for Object Detection from Point Cloud, arXiv 1907.03670, 2019

[117] Wang B, An J, and Cao J, Voxel-FPN: multi-scale voxel feature aggregation in 3D object detection from point clouds, arXiv 1907.05286, 2019

[118] Yang Z et al., STD: Sparse-to-Dense 3D Object Detector for Point Cloud, arXiv 1907.10471,2019

[119] Chen Y et al., Fast Point RCNN, arXiv 1908.02990, 2019

[120] Ngiam J et al., StarNet: Targeted Computation for Object Detection in Point Clouds, arXiv 1908.11069, 2019

[121] Zhu B et al., Class-balanced Grouping and Sampling for Point Cloud 3D Object Detection, arXiv 1908.09492, 2019

[122] Qi C et al., Deep Hough Voting for 3D Object Detection in Point Clouds, arXiv 1904.09664, 2019

[123] Meyer G et al., LaserNet: An Efficient Probabilistic 3D Object Detector for Autonomous Driving, arXiv 1903.08701, 2019

[124] Guo Y et al., Deep Learning for 3D Point Clouds: A Survey, arXiv 1912.12033, 2019

[125] Bello Y, Yu S, Wang C, Review: Deep Learning on 3D Point Clouds, arXiv 2001.06280, 2020

[126] Chen X et al., Monocular 3D Object Detection for Autonomous Driving, IEEE CVPR 2016

[127] F Chabot et al., Deep MANTA: A Coarse-to-fine Many-Task Network for joint 2D and 3D vehicle analysis from monocular image， IEEE CVPR， 2017

[128] A Mousavian et al., 3D Bounding Box Estimation Using Deep Learning and Geometry， IEEE CVPR， 2017

[129] H-N Hu et al., Joint Monocular 3D Vehicle Detection and Tracking， arXiv 1811.10742, 11， 2018

[130] T Roddick, A Kendall, R Cipolla, Orthographic Feature Transform for Monocular 3D Object Detection， arXiv 1811.08188, 11， 2018

[131] B Xu, Z Chen， Multi-Level Fusion based 3D Object Detection from Monocular Images, IEEE CVPR, 2018

[132] Z Qin， J Wang， Y Lu， MonoGRNet: A Geometric Reasoning Network for Monocular 3D Object Localization, arXiv 1811.10247, 11, 2018

[133] T He, S Soatto， Mono3D++: Monocular 3D Vehicle Detection with Two-Scale 3D Hypotheses and Task Priors, arXiv 1901.03446, 1, 2019

[134] Weng X, Kitani K, Monocular 3D Object Detection with Pseudo-LiDAR Point Cloud, arXiv 1903.09847, 2019

[135] Li B et al., GS3D: An Efficient 3D Object Detection Framework for Autonomous Driving, arXiv 1903.10955, 2019

[136] Ma X et al., Accurate Monocular Object Detection via Color-Embedded 3D Reconstruction for Autonomous Driving, arXiv 1903.11444, 2019

[137] Ku J, Pon A D, Waslander S L, Monocular 3D Object Detection Leveraging Accurate Proposals and Shape Reconstruction, arXiv 1904.01690, 2019

[138] Barabanau I et al., Monocular 3D Object Detection via Geometric Reasoning on Keypoints, arXiv 1905.05618, 2019

[139] Naiden A et al., Shift R-CNN: Deep Monocular 3d Object Detection with Closed-Form Geometric Constraints, arXiv 1905.09970, 2019

[140] Simonelli A et al., Disentangling Monocular 3D Object Detection, arXiv 1905.12365, 2019

[141] Jörgensen E, Zach C, Kahl F. Monocular 3D Object Detection and Box Fitting Trained End-to-End Using Intersection-over-Union Loss, arXiv 1906.08070, 2019

[142] Brazil G, Liu X. M3D-RPN: Monocular 3D Region Proposal Network for Object Detection， arXiv 1907.06038, 2019

[143] Wang X et al., Task-Aware Monocular Depth Estimation for 3D Object Detection, arXiv 1909.07701, 2019

[144] Marie J et al., Refined MPL: Refined Monocular PseudoLiDAR for 3D Object Detection in Autonomous Driving, arXiv 1911.09712, 2019

[145] Simonelli1 A et al., Towards Generalization Across Depth for Monocular 3D Object Detection, arXiv 1912.08035, 2020

[146] Li P et al., RTM3D: Real-time Monocular 3D Detection from Object Keypoints for Autonomous Driving, arXiv 2001.03343, 2020

[147] Cai Y et al., Monocular 3D Object Detection with Decoupled Structured Polygon Estimation and Height-Guided Depth Estimation, arXiv 2002.01619, 2020

[148] Li P, Chen X, Shen S, Stereo R-CNN based 3D Object Detection for Autonomous Driving， arXiv 1902.09738, 2019

[149] Qin Z, Wang J, Lu Y, Triangulation Learning Network: from Monocular to Stereo 3D Object Detection， arXiv 1906.01193, 2019

[150] Wang Y et al., Pseudo-LiDAR from Visual Depth Estimation: Bridging the Gap in 3D Object Detection for Autonomous Driving, arXiv 1812.07179, 2019

[151] Pon A et al., Object-Centric Stereo Matching for 3D Object Detection, arXiv 1909.07566, 2019

[152] Arnold E et al., A Survey on 3D Object Detection Methods for Autonomous Driving Applications, IEEE T-ITS, 20(10), Oct. 2019

[153] Rahman M M et al., Recent Advances in 3D Object Detection in the Era of Deep Neural Networks: A Survey, IEEE T-IP, Nov. 2019

[154] Danzer A et al., 2D Car Detection in Radar Data with PointNets, arXiv 1904.08414, 2019

[155] Nabati R, Qi H, RRPN: Radar Region Proposal Network for Object Detection in Autonomous Vehicles, arXiv 1905.00526, 2019

[156] Major B et al., Vehicle Detection with Automotive Radar Using Deep Learning on Range-Azimuth-Doppler Tensors, IEEE ICCV workshop, 2019

[157] D Eigen, C Puhrsch and R Fergus, Depth Map Prediction from a Single Image using a Multi-Scale Deep Network, arXiv 1406.2283, 2014

[158] D Eigen, R Fergus, Predicting Depth, Surface Normals and Semantic Labels with a Common Multi-Scale Convolutional Architecture, IEEE ICCV 2015

[159] P Wang et al., Towards Unified Depth and Semantic Prediction From a Single Image, IEEE CVPR 2015

[160] F Liu et al., Deep Convolutional Neural Fields for Depth Estimation from a Single Image, IEEE CVPR 2015

[161] R Garg et al., Unsupervised CNN for Single View Depth Estimation: Geometry to the Rescue, ECCV 2016

[162] C Godard et al., Unsupervised Monocular Depth Estimation with Left-Right Consistency, IEEE CVPR 2017

[163] Y Kuznietsov et al., Semi-Supervised Deep Learning for Monocular Depth Map Prediction, IEEE CVPR 2017

[164] T Zhou et al., Unsupervised Learning of Depth and Ego-Motion from Video, IEEE CVPR 2017

[165] Y Luo et al., Single View Stereo Matching, arXiv 1803.02612, 2018

[166] J Jiao et al., Look Deeper into Depth: Monocular Depth Estimation with Semantic Booster and Attention-Driven Loss, ECCV, 2018

[167] C Wang et al., Learning Depth from Monocular Videos using Direct Methods, IEEE CVPR 2018

[168] V Casser et al., Depth Prediction Without the Sensors: Leveraging Structure for Unsupervised Learning from Monocular Videos, arXiv 1811.06152, 2018

[169] X Qi et al., GeoNet: Geometric Neural Network for Joint Depth and Surface Normal Estimation, IEEE CVPR 2018

[170] Z Yin, J Shi, GeoNet - Unsupervised Learning of Dense Depth, Optical Flow and Camera Pose, IEEE CVPR 2018

[171] Zou Y, Luo Z, Huang J., DF-Net: Unsupervised Joint Learning of Depth and Flow using Cross-Task Consistency, arXiv 1809.01649, 2018

[172] Z Yang et al., LEGO: Learning Edge with Geometry all at Once by Watching Videos, *AAAI*, 2019

[173] Ranjan A et al., Competitive Collaboration: Joint Unsupervised Learning of Depth, Camera Motion, Optical Flow and Motion Segmentation, arXiv 1805.09806, 2019

[174] Casser V et al., Unsupervised Monocular Depth and Ego-motion Learning with Structure and Semantics, arXiv 1906.05717, 2019

[175] Bian J et al., Unsupervised Scale-consistent Depth and Ego-motion Learning from Monocular Video, arXiv 1908.10553, 2019

[176] Andraghetti L et al., Enhancing self-supervised monocular depth estimation with traditional visual odometry, arXiv 1908.03127, 2019

[177] Zhou L, Fang J, Liu G, Unsupervised Video Depth Estimation Based on Ego-motion and Disparity Consensus, arXiv 1909.01028, 2019

[178] Shi Y et al., Self-Supervised Learning of Depth and Ego-motion with Differentiable Bundle Adjustment, arXiv 1909.13163, 2019

[179] Wang G et al., Unsupervised Learning of Depth, Optical Flow and Pose with Occlusion from 3D Geometry, arXiv 2003.00766, 2020

[180] A Kendall et al.，End-to-End Learning of Geometry and Context for Deep Stereo Regression, ICCV, 2017

[181] C Zhou et al., Unsupervised Learning of Stereo Matching, ICCV 2017

[182] J R Chang, Y S Chen, Pyramid Stereo Matching Network, arXiv 1803.08669, 2018

[183] G Yang et al., SegStereo: Exploiting Semantic Information for Disparity, ECCV 2018.

[184] J Zhang et al., DispSegNet: Leveraging Semantics for End-to-End Learning of Disparity Estimation from Stereo Imagery, arXiv 1809.04734, 2019

[185] H Laga, A Survey on Deep Learning Architectures for Image-based Depth Reconstruction, arXiv 1906.06113, June 2019

[186] C Zhao et al., Monocular Depth Estimation Based on Deep Learning: An Overview, arXiv 2003.06620, Mar. 2020

[187] Jonas Uhrig et al., Sparsity Invariant CNNs，*Int. Conf. on 3D Vision*，8，2017

[188] A Eldesokey et al., Propagating Confidences through CNNs for Sparse Data Regression，BMCV，5，2018

[189] HMS-Net: Hierarchical Multi-scale Sparsity-invariant Network for Sparse Depth Completion, arXiv 1808.08685, 2018

[190] Nick Schneider et. Al., RegNet: Multimodal Sensor Registration Using Deep Neural Networks, arXiv 1707.03167, 2017.

[191] G Iyer et al., CalibNet: Self-Supervised Extrinsic Calibration using 3D Spatial Transformer Networks, arXiv 1803.08181, 2018

[192] F Ma，S Karaman，Sparse-to-Dense: Depth Prediction from Sparse Depth Samples and a Single Image，arXiv 1709.07492, 9 2017

[193] Y Zhang, T Funkhouser，Deep Depth Completion of a RGB-D Image，IEEE CVPR，2018

[194] Z Chen et al., Estimating Depth from RGB and Sparse Sensing，ECCV，4，2018

[195] K Park, S Kim, K Sohn，High-precision Depth Estimation with the 3D LiDAR and Stereo Fusion，IEEE ICRA，5，2018

[196] F Ma et al., Self-Supervised Sparse-to-Dense: Depth Completion from LiDAR and Mono Camera，arXiv 1807.00275, 7，2018

[197] M Jaritz et al., Sparse and Dense Data with CNNs: Depth Completion and Semantic Segmentation，arXiv 1808.00769, 8，2018

[198] M Dimitrievski, P Veelaert, W Philips，Learn Morphological Operators for Depth Completion, *Int. Conf. on Advanced Concepts for Intelligent Vision Systems*, 7, 2018

[199] J Qiu et al., DeepLiDAR: Deep Surface Normal Guided Depth Prediction from LiDAR and Color Image, arXiv 1812.00488, 12, 2018

[200] Y Yang, A Wong, S Soatto，Dense Depth Posterior (DDP) from Single Image and Sparse Range, arXiv 1901.10034, 1, 2019

[201] S. Shivakumar et al., DFuseNet: Fusion of RGB and Sparse Depth for Image Guided Dense Depth Completion，arXiv 1902.00761, 2，2019

[202] Gansbeke W V et al, Sparse and noisy LiDAR completion with RGB guidance and uncertainty, arXiv 1902.05356, 2019

[203] T-H Wang et al., 3D LiDAR and Stereo Fusion using Stereo Matching Network with Conditional Cost Volume Normalization, arXiv 1904.02917, 2019

[204] Zhong Y et al., Deep RGB-D Canonical Correlation Analysis For Sparse Depth Completion, arXiv 1906.08967, 2019

[205] Eldesokey A, Felsberg M, Khan F S, Confidence Propagation through CNNs for Guided Sparse Depth Regression, arXiv 1811.01791, 2019

[206] Tang J et al., Learning Guided Convolutional Network for Depth Completion, arXiv 1908.01238, 2019

[207] Zhang Y et al., DFineNet: Ego-Motion Estimation and Depth Refinement from Sparse, Noisy Depth Input with RGB Guidance, arXiv 1903.06397, 2019

[208] Liu H et al., PLIN: A Network for Pseudo-LiDAR Point Cloud Interpolation, arXiv 1909.07137, 2019

[209] Xu Y et al., Depth Completion from Sparse LiDAR Data with Depth-Normal Constraints, arXiv 1910.06727, 2019

[210] X Chen et al., Multi-View 3D Object Detection Network for Autonomous Driving，arXiv 1611.07759, 2017.

[211] D Matti, H Kemal Ekenel, J-P Thiran, Combining LiDAR Space Clustering and Convolutional Neural Networks for Pedestrian Detection, arXiv 1710.06160, 2017

[212] Z Wang et al., Fusing Bird's Eye View LIDAR Point Cloud and Front View Camera Image for Deep Object Detection, arXiv 1711.06703, 2018

[213] D Xu et al., PointFusion: Deep Sensor Fusion for 3D Bounding Box Estimation, arXiv 1711.10871, 2018.

[214] X Du et al., A General Pipeline for 3D Detection of Vehicles，arXiv 1803.00387, 2018

[215] K Shin et al., RoarNet: A Robust 3D Object Detection based on RegiOn Approximation Refinement, arXiv 1811.03818, 2018

[216] J Ku et al., Joint 3D Proposal Generation and Object Detection from View Aggregation, arXiv 1712.02294, 2018

[217] C Qi et al., Frustum PointNets for 3D Object Detection from RGB-D Data, arXiv 1711.08488, 2018.

[218] M Liang et al., Deep Continuous Fusion for Multi-Sensor 3D Object Detection, ECCV, 2018

[219] D Frossard, R Urtasun, End-to-end Learning of Multi-sensor 3D Tracking by Detection, arXiv 1806.11534, 2018.

[220] Zhao X et al., 3D Object Detection Using Scale Invariant and Feature Reweighting Networks, arXiv 1901.02237, 2019

[221] Sindagi V A, Zhou Y, Tuzel O, MVX-Net: Multimodal VoxelNet for 3D Object Detection, arXiv 1904.01649, 2019

[222] Meyer G et al., Sensor Fusion for Joint 3D Object Detection and Semantic Segmentation, arXiv 1904.11466, 2019

[223] Ku J et al., Improving 3D Object Detection for Pedestrians with Virtual Multi-View Synthesis Orientation Estimation, arXiv 1907.06777, 2019

[224] Raffiee A, Irshad H, Class-specific Anchoring Proposal for 3D Object Recognition in LIDAR and RGB Images, arXiv 1907.09081, 2019

[225] You Y et al., Pseudo-LiDAR++: Accurate Depth for 3D Object Detection in Autonomous Driving, arXiv 1906.06310, 2019

[226] Deng J, Czarnecki K, MLOD: A multi-view 3D object detection based on robust feature fusion method, arXiv 1909.04163, 2019

[227] Li P, Liu S, Shen S, Multi-Sensor 3D Object Box Refinement for Autonomous Driving, arXiv 1909.04942, 2019

[228] Chen X, Girshick R, He K, Dollár P, TensorMask: A Foundation for Dense Object Segmentation, arXiv 1903.12174, 2019

[229] Kirillov A, Wu Y, He K, Girshick R, PointRend: Image Segmentation as Rendering, arXiv 1912.08193, 2019

[230] Lee Y, Park J, CenterMask: Real-Time Anchor-Free Instance Segmentation, IEEE CVPR, 2020

[231] Mohan R, Valada A, EfficientPS: Efficient Panoptic Segmentation, arXiv 2004.02307, 2020

[232] Lateef F, Ruicheck Y, Survey on semantic segmentation using deep learning techniques, *Neural Computing*, 338, 2019

[233] Ulku I, Akagunduz E, A Survey on Deep Learning-based Architectures for Semantic Segmentation on 2D images, arXiv 1912.10230, 2019

[234] S Minaee et al., Image Segmentation Using Deep Learning: A Survey, arXiv 2001.05566, April 2020

[235] Feng D et al., Deep Multi-modal Object Detection and Semantic Segmentation for Autonomous Driving: Datasets, Methods, and Challenges, arXiv 1902.07830, 2020

[236] J Hur, S Roth, Optical Flow Estimation in the Deep Learning Age, arXiv 2004.02853, April 2020

[237] S Krebs et al., Survey on Leveraging Deep Neural Networks for Object Tracking, *IEEE ITSC*, 10, 2017

[238] G Ciaparrone et al., Deep Learning in Video Multi-Object Tracking: A Survey, arXiv 1907.12740, Nov. 2019

[239] S M Marvasti-Zadehar et al., Deep Learning for Visual Tracking: A Comprehensive Survey, arXiv 1912.00535, 2019

[240] Y Chen, Y Tian, M He, Monocular human pose estimation: A survey of deep learning-based methods, *CVIU*, vol. 192, Mar. 2020

[241] Cadena C et al., Past, Present, and Future of Simultaneous Localization and Mapping: Toward the Robust-Perception Age, arXiv 1606.05830, 2016

[242] S Milz et al., Visual SLAM for Automated Driving: Exploring the Applications of Deep Learning, *IEEE CVPR Workshop*, 2018

[243] B Huang, J Zhao, J Liu, A Survey of Simultaneous Localization and Mapping, arXiv 1909.05214, Aug. 2019

[244] Uy M and Lee G. PointNetVLAD: deep point cloud-based retrieval for large-scale place recognition. IEEE CVPR, 2018

[245] Lu W et al. L3-Net: Towards learning based LiDAR localization for autonomous driving. IEEE CVPR, 2019.

[246] Ono Y et al., LF-Net: Learning Local Features from Images. arXiv 1805.09662, 2018

[247] DeTone D, Malisiewicz T, Rabinovich A. SuperPoint: Self-supervised interest point detection and description. IEEE CVPR Workshops, 2018

[248] B Ummenhofer et al., DeMoN: Depth and Motion Network for Learning Monocular Stereo, IEEE CVPR 2017

[249] R Li et al. UnDeepVO: Monocular Visual Odometry through Unsupervised Deep Learning. arXiv 1709.06841, 2017.

[250] S Wang et al., DeepVO: Towards End-to-End Visual Odometry with Deep Recurrent Convolutional Neural Networks, arXiv 1709.08429, 2017

[251] R Clark et al. VINet: Visual-Inertial Odometry as a Sequence-to-Sequence Learning Problem. *AAAI*. 2017

[252] S Vijayanarasimhan et al., SfM-Net: Learning of Structure and Motion from Video, arXiv 1704.07804, 2017

[253] K Tateno K et al. CNN-SLAM: Real-time dense monocular SLAM with learned depth prediction. arXiv 1704.03489, 2017.

[254] Yang N et al., Deep Virtual Stereo Odometry: Leveraging Deep Depth Prediction for Monocular Direct Sparse Odometry, arXiv 1807.02570, 2018

[255] Li S et al., Sequential Adversarial Learning for Self-Supervised Deep Visual Odometry, arXiv 1908.08704, 2019

[256] Zhan H et al., Visual Odometry Revisited: What Should Be Learnt? arXiv 1909.09803, 2019

[257] Zhao C, Tang Y, Sun Q, Deep Direct Visual Odometry, arXiv 1912.05101, 2019

[258] Alahi A et al., Social LSTM: Human Trajectory Prediction in Crowded Spaces, IEEE CVPR 2016

[259] Pfeiffer M et al., A Data-driven Model for Interaction-aware Pedestrian Motion Prediction in Object Cluttered Environments, arXiv 1709.08528, 2018

[260] Gupta A et al., Social GAN: Socially Acceptable Trajectories with Generative Adversarial Networks, CVPR 2018

[261] Sedegian A et al., SoPhie: An Attentive GAN for Predicting Paths Compliant to Social and Physical Constraints, arXiv 1806.01482, Sep. 2018

[262] Vemula A, Muelling K, Oh J, Social Attention: Modeling Attention in Human Crowds, ICRA, 2018

[263] Hoy M et al., Learning to Predict Pedestrian Intention via Variational Tracking Networks, *IEEE ITSC*, Nov. 2018

[264] Xue H, Huynh D, Reynolds M, Location-Velocity Attention for Pedestrian Trajectory Prediction, *IEEE WACV*, 2019

[265] Haddad S et al., Situation-Aware Pedestrian Trajectory Prediction with Spatio-Temporal Attention Model, *IEEE WACV*, 2019

[266] Zhang P et al., SR-LSTM: State Refinement for LSTM towards Pedestrian Trajectory Prediction, arXiv 1903.02793, Mar. 2019

[267] Zhu Y. et al., StarNet: Pedestrian Trajectory Prediction using Deep Neural Network in Star Topology, arXiv 1906.01797, June 2019

[268] Zhao T et al., Multi-Agent Tensor Fusion for Contextual Trajectory Prediction, arXiv 1904.04776, July 2019

[269] Amirian J, Hayet J-B, Pettre J, Social Ways: Learning Multi-Modal Distributions of Pedestrian Trajectories with GANs, *IEEE CVPR workshop*, 2019

[270] Li Y., Which Way Are You Going? Imitative Decision Learning for Path Forecasting in Dynamic Scenes, IEEE CVPR 2019

[271] Chandra R et al., TraPHic: Trajectory Prediction in Dense and Heterogeneous Traffic Using Weighted Interactions, IEEE CVPR 2019

[272] Choi C, Dariush B, Learning to Infer Relations for Future Trajectory Forecast, IEEE ICCV 2019

[273] Liang J et al., Peeking into the Future: Predicting Future Person Activities and Locations in Videos, IEEE CVPR 2019

[274] A Rudenko et al., Human Motion Trajectory Prediction: A Survey, arXiv 1905.06113, May 2019

[275] Fan H et al., An Auto-tuning Framework for Autonomous Vehicles, arXiv 1808.04913, 2018

[276] Wolf P et al., Adaptive Behavior Generation for Autonomous Driving using Deep Reinforcement Learning with Compact Semantic States, arXiv1809.03214, 2018

[277] Sun L, Zhan W, Tomizuka M, Probabilistic Prediction of Interactive Driving Behavior via Hierarchical Inverse Reinforcement Learning, arXiv 1809.02926, 2018

[278] Xu Z, Tang C, Tomizuka M, Zero-shot Deep Reinforcement Learning Driving Policy Transfer for Autonomous Vehicles based on Robust Control, arXiv 1812.03216, 2018

[279] Cui H et al., Multimodal Trajectory Predictions for Autonomous Driving using Deep Convolutional Networks, arXiv 1809.10732, 2018

[280] Ma Y et al., TrafficPredict: Trajectory Prediction for Heterogeneous Traffic-Agents, arXiv 1811.02146, 2018

[281] Bansal M, Krizhevsky A, Ogale A, ChauffeurNet: Learning to Drive by Imitating the Best and Synthesizing the Worst, arXiv 1812.03079, 2018

[282] Codevilla F et al., Exploring the Limitations of Behavior Cloning for Autonomous Driving, arXiv 1904.08980, 2019

[283] Hoel C-J et al., Combining Planning and Deep Reinforcement Learning in Tactical Decision Making for Autonomous Driving, arXiv 1905.02680, 2019

[284] Moghadam M, Elkaim G H, A Hierarchical Architecture for Sequential Decision-Making in Autonomous Driving using Deep Reinforcement Learning, 1906.08464, 2019

[285] Lee N et al., DESIRE: Distant Future Prediction in Dynamic Scenes with Interacting Agents, arXiv 1704.04394, 2017

[286] Srikanth S et al., INFER: INtermediate representations for FuturE pRediction, arXiv 1903.10641, 2019

[287] Si W, Wei T, Liu C, AGen: Adaptable Generative Prediction Networks for Autonomous Driving, *IEEE IV*, 2019

[288] Li J, Ma H, Tomizuka M, Conditional Generative Neural System for Probabilistic Trajectory Prediction, arXiv 1905.01631, 2019

[289] Li J, Ma H, Zhan W, Tomizuka M, Coordination and Trajectory Prediction for Vehicle Interactions via Bayesian Generative Modeling, arXiv 1905.00587, 2019

[290] Li J, Ma H, Tomizuka M, Interaction-aware Multi-agent Tracking and Probabilistic Behavior Prediction via Adversarial Learning, arXiv 1904.02390, 2019

[291] Hu Y, Sun L, Tomizuka M, Generic Prediction Architecture Considering both Rational and Irrational Driving Behaviors, arXiv 1907.10170, 2019

[292] Cho K et al., Deep Predictive Autonomous Driving Using Multi-Agent Joint Trajectory Prediction and Traffic Rules, IEEE IROS, 2019

[293] Li X, Ying X, Chuah M C, GRIP: Graph-based Interaction-aware Trajectory Prediction, arXiv 1907.07792, 2019

[294] Chai Y et al., Multi-Path: Multiple Probabilistic Anchor Trajectory Hypotheses for Behavior Prediction, arXiv 1910.05449, 2019

[295] Lee D et al., Joint Interaction and Trajectory Prediction for Autonomous Driving using Graph Neural Networks, arXiv 1912.07882, 2019

[296] Chandra R et al., Forecasting Trajectory and Behavior of Road-Agents Using Spectral Clustering in Graph-LSTMs, arXiv 1912.01118, 2019

[297] Li J et al., Social-WaGDAT: Interaction-aware Trajectory Prediction via Wasserstein Graph Double-Attention Network, arXiv 2002.06241, 2020

[298] Li J et al., EvolveGraph: Heterogeneous Multi-Agent Multi-Modal Trajectory Prediction with Evolving Interaction Graphs, arXiv 2003.13924, 2020

[299] Hu Y, Zhan W, Tomizuka M, Scenario-Transferable Semantic Graph Reasoning for Interaction-Aware Probabilistic Prediction, arXiv 2004.03053, 2020

[300] Gao J et al., VectorNet: Encoding HD Maps and Agent Dynamics from Vectorized Representation, arXiv 2005.04259, 2020

[301] S Mozaffari et al., Deep Learning-based Vehicle Behavior Prediction for Autonomous Driving Applications: A Review, arXiv 1912.11676, Dec. 2019.

[302] S Kuutti et al., A Survey of Deep Learning Applications to Autonomous Vehicle Control, arXiv 1912.10773, Dec. 2019

[303] M Bojarski et al., End to End Learning for Self-Driving Cars, arXiv 1604.07316, 2016

[304] E Santana, G Hotz, Learning a Driving Simulator, arXiv 1608.01230, 2016

[305] Y Tian et al., DeepTest: Auto Testing of DNN-driven Autonomous Cars, arXiv 1708.08559， 2017

[306] S Hecker, D Dai, L Gool, Failure Prediction for Autonomous Driving, arXiv 1805.01811， 2018

[307] Y Chen et al., LiDAR-Video Driving Dataset: Learning Driving Policies Effectively, IEEE CVPR 2018

[308] S Hecker, D Dai, L Gool, End-to-End Learning of Driving Models with Surround-View Cameras and Route Planners, ECCV 2018

[309] Grigorescu S et al., NeuroTrajectory: A Neuroevolutionary Approach to Local State Trajectory Learning for Autonomous Vehicles, arXiv 1906.10971, 2019

[310] T Wheeler et al., Deep Stochastic Radar Models, IEEE IV, 2017

[311] H Alhaija et al., Augmented Reality Meets Computer Vision: Efficient Data Generation for Urban Driving Scenes, arXiv 1708.01566， 2017

[312] VSR Veeravasarapu, C Rothkopf, R Visvanathan, Adversarially Tuned Scene Generation, arXiv 1701.00405, 2017

[313] X Yue et al， A LiDAR Point Cloud Generator: from a Virtual World to Autonomous Driving, arXiv 1804.00103, 2018

[314] K Elmadawi et al., End-to-end sensor modeling for LiDAR Point Cloud， arXiv 1907.07748v1， 2019， 7

[315] Wang T et al., Artificial Intelligence for Vehicle-to-Everything: A Survey, Special Section on Emerging Technologies on Vehicle to Everything (V2x), IEEE Access, 2019

[316] C Zhang, P Patras, H Haddadi, Deep Learning in Mobile and Wireless Networking, IEEE Communications Surveys and Tutorials, 21(3), 2019

[317] R McAllister et al. Concrete Problems for Autonomous Vehicle Safety: Advantages of Bayesian Deep Learning， *IJCAI*， 2017

[318] T Dreossi et al., VERIFAI: A Toolkit for the Formal Design and Analysis of Artificial Intelligence-Based Systems, *Int. Conf. on Computer Aided Verification* (CAV), 2019

[319] Jha S et al., ML-based Fault Injection for Autonomous Vehicles: A Case for Bayesian Fault Injection, arXiv 1907.01051, 2019

[320] Huang X et al., A Survey of Safety and Trustworthiness of Deep Neural Networks: Verification, Testing, Adversarial Attack and Defence, and Interpretability, arXiv 1812.08342, 2019

[321] Baidu Apollo：https://github.com/ApolloAuto/apollo

[322] Autoware：https://www.autoware.org/

[323] OpenPilot：https://github.com/commaai/research

[324] Udacity：https://github.com/udacity/self-driving-car

[325] Microsoft AirSim: https://github.com/Microsoft/AirSim

[326] Intel Carla: https://carla.org/

[327] LG LGSVL: https://www.lgsvlsimulator.com/

[328] SUMO: http://sumo.sourceforge.net/

[329] FLOW: https://flow-project.github.io/

[330] GM cruise.ai Visualization: https://webviz.io/

[331] Uber visualization: https://avs.auto/#/

[332] KITTI data: http://www.cvlibs.net/datasets/kitti/raw_data.php

[333] BDD: http://bdd-data.berkeley.edu/.

[334] Baidu ApolloScope: http://apolloscape.auto/

[335] NuScenes: https://www.nuscenes.org/

[336] Udacity: https://github.com/udacity/self-driving-car/tree/master/datasets

[337] Ford ArgoVerse: https://www.argoverse.org/data.html

[338] Google WayMo: https://waymo.com/open

[339] Lyft Level 5: https://level5.lyft.com/dataset/

[340] HONDA dataset: https://usa.honda-ri.com/H3D

[341] Scale and Hesai's PandaSet: https://scale.com/open-datasets/pandaset

[342] NGSIM data: https://ops.fhwa.dot.gov/trafficanalysistools/ngsim.htm

[343] HighD dataset: https://www.highd-dataset.com/

[344] INTERACTION dataset: https://interaction-dataset.com/

[345] Marcus G, Deep Learning: A Critical Appraisal, arXiv 1801.00631, 2018

[346] Yullie A, Liu C, Deep Nets: What have they ever done for Vision? arXiv 1805.04025, 2019

[347] Poggio T et al., Theoretical Issues in Deep Networks: Approximation, Optimization and Generalization, arXiv1908.09375, 2019

[348] Thys S, Ranst W, Goedeme T, Fooling Automated Surveillance Cameras: Adversarial Patches to Attack Person Detection, arXiv 1904.08653, 2019

[349] Jia Y et al., Fooling Detection Alone Is Not Enough: Adversarial Attack Against Multiple Object Tracking, ICLR, 2020

[350] Mohseni S et al., Practical Solutions for Machine Learning Safety in Autonomous Vehicles, arXiv, 2019

[351] Loquercio A, Segu M, Scaramuzza D, A General Framework for Uncertainty Estimation in Deep Learning, arXiv 1907.06890, 2020

TABLE VIII. COMPARISON OF AV DATASETS, FROM [335]

| **Dataset** | **Year** | **Sce-nes** | **Size (hr)** | **RGB imgs** | **PCs lidar††** | **PCs radar** | **Ann. frames** | **3D boxes** | **Night / Rain** | **Map layers** | **Clas-ses** | **Locations** |
|---|---|---|---|---|---|---|---|---|---|---|---|---|
| CamVid [ ] | 2008 | 4 | 0.4 | 18k | 0 | 0 | 700 | 0 | No/No | 0 | 32 | Cambridge |
| Cityscapes [ ] | 2016 | n/a | - | 25k | 0 | 0 | 25k | 0 | No/No | 0 | 30 | 50 cities |
| Vistas [ ] | 2017 | n/a | - | 25k | 0 | 0 | 25k | 0 | Yes/Yes | 0 | 152 | Global |
| BDD100K [ ] | 2017 | 100k | 1k | 100M | 0 | 0 | 100k | 0 | Yes/Yes | 0 | 10 | NY, SF |
| ApolloScape [ ] | 2018 | - | 100 | 144k | 0** | 0 | 144k | 70k | Yes/No | 0 | 8-35 | 4x China |
| $D^2$-City [ ] | 2019 | 1k† | - | 700k† | 0 | 0 | 700k† | 0 | No/Yes | 0 | 12 | 5x China |
| KITTI [ ] | 2012 | 22 | 1.5 | 15k | 15k | 0 | 15k | 200k | No/No | 0 | 8 | Karlsruhe |
| AS lidar [ ] | 2018 | - | 2 | 0 | 20k | 0 | 20k | 475k | -/- | 0 | 6 | China |
| KAIST [ ] | 2018 | - | - | 8.9k | 8.9k | 0 | 8.9k | 0 | **Yes**/No | 0 | 3 | Seoul |
| H3D [ ] | 2019 | 160 | 0.77 | 83k | 27k | 0 | 27k | 1.1M | No/No | 0 | 8 | SF |
| nuScenes | 2019 | **1k** | 5.5 | **1.4M** | **400k** | **1.3M** | **40k** | 1.4M | **Yes/Yes** | **11** | **23** | Boston, SG |
| Argoverse [ ] | 2019 | 113† | 0.6† | 490k† | 44k | 0 | 22k† | 993k† | **Yes/Yes** | 2 | 15 | Miami, PT |
| Lyft L5 [ ] | 2019 | 366 | 2.5 | 323k | 46k | 0 | **46k** | 1.3M | No/No | 7 | 9 | Palo Alto |
| Waymo Open [ ] | 2019 | **1k** | 5.5 | 1M | 200k | 0 | **200k‡** | **12M‡** | **Yes/Yes** | 0 | 4 | 3x USA |
| A*3D [ ] | 2019 | n/a | **55** | 39k | 39k | 0 | **39k** | 230k | **Yes/Yes** | 0 | 7 | SG |
| A2D2 [ ] | 2019 | n/a | - | - | - | 0 | 12k | - | -/- | 0 | 14 | 3x Germany |

TABLE IX. COMPARISON OF TRAJECTORY DATASETS, FROM [344]

| | highly interactive scenarios | complexity of scenarios | density of aggressive behavior | near-collision situations and collisions | HD maps with semantics | completeness of interaction entities & viewpoint |
|---|---|---|---|---|---|---|
| NGSIM [ ] | ramp merging, (double) lane change | structured roads, explicit right-of-way | low | very few near-collision | no | yes, bird's-eye-view from a building |
| highD [ ] | lane change | structured roads, explicit right-of-way | low | very few near-collision | no | yes, bird's-eye-view from a drone |
| Argoverse [ ] | unsignalized intersections, pedestrian crossing | unstructured roads, inexplicit right-of-way | low | no | yes, but partially | only for the ego data-collection vehicle |
| INTERACTION | roundabouts, ramp merging, double lane change unsignalized intersections | unstructured roads, inexplicit right-of-way | high | yes | yes | yes, bird's-eye-view from a drone |

TABLE X. COMPARISON OF 3D LIDAR DETECTION RESULTS ON THE KITTI TEST 3D DETECTION BENCHMARK, FROM [124]

| **Method** | | | **Modality** | **Speed (fps)** | **Cars** | | | **Pedestrians** | | | **Cyclists** | | |
|---|---|---|---|---|---|---|---|---|---|---|---|---|---|
| | | | | | **E** | **M** | **H** | **E** | **M** | **H** | **E** | **M** | **H** |
| Region Proposal -based Methods | Multi-view Methods | MV3D [ ] | L & I | 2.8 | 74.97 | 63.63 | 54.00 | - | - | - | - | - | - |
| | | AVOD [ ] | L & I | 12.5 | 76.39 | 66.47 | 60.23 | 36.10 | 27.86 | 25.76 | 57.19 | 42.08 | 38.29 |
| | | ContFuse [ ] | L & I | 16.7 | 83.68 | 68.78 | 61.67 | - | - | - | - | - | - |
| | | MMF [ ] | L & I | 12.5 | 88.40 | 77.43 | 70.22 | - | - | - | - | - | - |
| | | SCANet [ ] | L & I | 11.1 | 79.22 | 67.13 | 60.65 | - | - | - | - | - | - |
| | | RT3D [ ] | L & I | 11.1 | 23.74 | 19.14 | 18.86 | - | - | - | - | - | - |
| | Segmentation -based Methods | IPOD [ ] | L & I | 5.0 | 80.30 | 73.04 | 68.73 | 55.07 | 44.37 | 40.05 | 71.99 | 52.23 | 46.50 |
| | | PointRCNN [ ] | L | 10.0 | 86.96 | 75.64 | 70.70 | 47.98 | 39.37 | 36.01 | 74.96 | 58.82 | 52.53 |
| | | PointRGCN [ ] | L | 3.8 | 85.97 | 75.73 | 70.60 | - | - | - | - | - | - |
| | | PointPainting [ ] | L & I | 2.5 | 82.11 | 71.70 | 67.08 | 50.32 | 40.97 | 37.87 | 77.63 | 63.78 | 55.89 |
| | | STD [ ] | L | 12.5 | 87.95 | 79.71 | 75.09 | 53.29 | 42.47 | 38.35 | 78.69 | 61.59 | 55.30 |
| | Frustum -based Methods | F-PointNets [ ] | L & I | 5.9 | 82.19 | 69.79 | 60.59 | 50.53 | 42.15 | 38.08 | 72.27 | 56.12 | 49.01 |
| | | SIFRNet [ ] | L & I | - | - | - | - | - | - | - | - | - | - |
| | | PointFusion [ ] | L & I | - | 77.92 | 63.00 | 53.27 | 33.36 | 28.04 | 23.38 | 49.34 | 29.42 | 26.98 |
| | | RoarNet [ ] | L & I | 10.0 | 83.71 | 73.04 | 59.16 | - | - | - | - | - | - |
| | | F-ConvNet [ ] | L & I | 2.1 | 87.36 | 76.39 | 66.69 | 52.16 | 43.38 | 38.80 | 81.98 | 65.07 | 56.54 |
| | | Patch Refinement [ ] | L | 6.7 | 88.67 | 77.20 | 71.82 | - | - | - | - | - | - |
| | Other Methods | 3D IoU loss [ ] | L | 12.5 | 86.16 | 76.50 | 71.39 | - | - | - | - | - | - |
| | | Fast Point R-CNN [ ] | L | 16.7 | 84.80 | 74.59 | 67.27 | - | - | - | - | - | - |
| | | VoteNet [ ] | L | - | - | - | - | - | - | - | - | - | - |
| | | Feng et al. [ ] | L | - | - | - | - | - | - | - | - | - | - |
| | | Part-A^2 [ ] | L | 12.5 | 87.81 | 78.49 | 73.51 | - | - | - | - | - | - |
| Single Shot Methods | BEV-based Methods | PIXOR [ ] | L | 28.6 | - | - | - | - | - | - | - | - | - |
| | | HDNET [ ] | L | 20.0 | - | - | - | - | - | - | - | - | - |
| | | BirdNet [ ] | L | 9.1 | 13.53 | 9.47 | 8.49 | 12.25 | 8.99 | 8.06 | 16.63 | 10.46 | 9.53 |
| | Point Cloud -based Methods | VeloFCN [ ] | L | 1.0 | - | - | - | - | - | - | - | - | - |
| | | 3D FCN [ ] | L | <0.2 | - | - | - | - | - | - | - | - | - |
| | | Vote3Deep [ ] | L | - | - | - | - | - | - | - | - | - | - |
| | | 3DBN [ ] | L | 7.7 | 83.77 | 73.53 | 66.23 | - | - | - | - | - | - |
| | | VoxelNet [ ] | L | 2.0 | 77.47 | 65.11 | 57.73 | 39.48 | 33.69 | 31.51 | 61.22 | 48.36 | 44.37 |
| | | SECOND [ ] | L | 26.3 | 83.34 | 72.55 | 65.82 | 48.96 | 38.78 | 34.91 | 71.33 | 52.08 | 45.83 |
| | | MVX-Net [ ] | L & I | 16.7 | 84.99 | 71.95 | 64.88 | - | - | - | - | - | - |
| | | PointPillars [ ] | L | 62.0 | 82.58 | 74.31 | 68.99 | 51.45 | 41.92 | 38.89 | 77.10 | 58.65 | 51.92 |
| | Other Methods | LaserNet [ ] | L | 83.3 | - | - | - | - | - | - | - | - | - |
| | | LaserNet++ [ ] | L & I | 26.3 | - | - | - | - | - | - | - | - | - |

TABLE XI. COMPARISON OF 3D CAMERA DETECTION RESULTS ON THE KITTI TEST DETECTION BENCHMARK, FROM [146]

| Method | Extra | Time | $AP_{3D}$ (IoU=0.5) | | | $AP_{3D}$ (IoU=0.7) | | |
|---|---|---|---|---|---|---|---|---|
| | | | Easy | Moderate | Hard | Easy | Moderate | Hard |
| Mono3D [ ] | Mask | 4.2 s | 25.19/ - | 18.20/ - | 15.52/ - | 2.53 / - | 2.31 / - | 2.31 / - |
| 3DOP [ ] | Stereo | 3 s | 46.04/ - | 34.63/ - | 30.09/ - | 6.55 / - | 5.07 / - | 4.10 / - |
| MF3D [ ] | Depth | - | 47.88/45.57 | 29.48/30.03 | 26.44/23.95 | 10.53/ 7.85 | 5.69 / 5.39 | 5.39 / 4.73 |
| Mono3D++ [ ] | Depth+Shape | >0.6s | 42.00/ - | 29.80/ - | 24.20/ - | 10.60/ - | 7.90 / - | 5.70 / - |
| GS3D [ ] | None | 2.3s | 32.15/30.60 | 29.89/26.40 | 26.19/22.89 | 13.46/11.63 | 10.97/10.51 | 10.38/10.51 |
| M3D-RPN [ ] | None | 0.16s | 48.96/49.89 | 39.57/36.14 | 33.01/28.98 | 20.27/**20.40** | **17.06/16.48** | 15.21/13.34 |
| Deep3DBox [ ] | None | - | 27.04/ - | 20.55/ - | 15.88/ - | 5.85 / - | 4.10 / - | 3.84 / - |
| MonoGRNet [ ] | None | 0.06s | 50.51/ - | 36.97/ - | 30.82/ - | 13.88/ - | 10.19/ - | 7.62 / - |
| FQNet[ ] | None | 3.33s | 28.16/28.98 | 21.02/20.71 | 19.91/18.59 | 5.98 / 5.45 | 5.50 / 5.11 | 4.75 / 4.45 |
| Ours (ResNet18) | None | 0.035 s | 47.43/46.52 | 33.86/32.61 | 31.04/30.95 | 18.13/18.38 | 14.14/14.66 | 13.33/12.35 |
| Ours (DLA34) | None | 0.055 s | **54.36/52.59** | **41.90/40.96** | **35.84/34.95** | **20.77**/19.47 | 16.86/16.29 | **16.63/15.57** |

TABLE XII. COMPARATIVE 3D LIDAR DETECTION RESULTS ON THE KITTI TEST BEV DETECTION BENCHMARK, FROM [124]

| Method | | | Modality | Speed (fps) | Cars | | | Pedestrians | | | Cyclists | | |
|---|---|---|---|---|---|---|---|---|---|---|---|---|---|
| | | | | | E | M | H | E | M | H | E | M | H |
| Region Proposal -based Methods | Multi-view Methods | MV3D [ ] | L & I | 2.8 | 86.62 | 78.93 | 69.80 | - | - | - | - | - | - |
| | | AVOD [ ] | L & I | 12.5 | 89.75 | 84.95 | 78.32 | 42.58 | 33.57 | 30.14 | 64.11 | 48.15 | 42.37 |
| | | ContFuse [ ] | L & I | 16.7 | 94.07 | 85.35 | 75.88 | - | - | - | - | - | - |
| | | MMF [ ] | L & I | 12.5 | 93.67 | 88.21 | 81.99 | - | - | - | - | - | - |
| | | SCANet [ ] | L & I | 11.1 | 90.33 | 82.85 | 76.06 | - | - | - | - | - | - |
| | | RT3D [ ] | L & I | 11.1 | 56.44 | 44.00 | 42.34 | - | - | - | - | - | - |
| | Segmentation -based Methods | IPOD [ ] | L & I | 5.0 | 89.64 | 84.62 | 79.96 | 60.88 | 49.79 | 45.43 | 78.19 | 59.40 | 51.38 |
| | | PointRCNN [ ] | L | 10.0 | 92.13 | 87.39 | 82.72 | 54.77 | 46.13 | 42.84 | 82.56 | 67.24 | 60.28 |
| | | PointRGCN [ ] | L | 3.8 | 91.63 | 87.49 | 80.73 | - | - | - | - | - | - |
| | | PointPainting [ ] | L & I | 2.5 | 92.45 | 88.11 | 83.36 | 58.70 | 49.93 | 46.29 | 83.91 | 71.54 | 62.97 |
| | | STD [ ] | L | 12.5 | 94.74 | 89.19 | 86.42 | 60.02 | 48.72 | 44.55 | 81.36 | 67.23 | 59.35 |
| | Frustum -based Methods | F-PointNets [ ] | L & I | 5.9 | 91.17 | 84.67 | 74.77 | 57.13 | 49.57 | 45.48 | 77.26 | 61.37 | 53.78 |
| | | SIFRNet [ ] | L & I | - | - | - | - | - | - | - | - | - | - |
| | | PointFusion [ ] | L & I | - | - | - | - | - | - | - | - | - | - |
| | | RoarNet [ ] | L & I | 10.0 | 88.20 | 79.41 | 70.02 | - | - | - | - | - | - |
| | | F-ConvNet [ ] | L & I | 2.1 | 91.51 | 85.84 | 76.11 | 57.04 | 48.96 | 44.33 | 84.16 | 68.88 | 60.05 |
| | | Patch Refinement [ ] | L | 6.7 | 92.72 | 88.39 | 83.19 | - | - | - | - | - | - |
| | Other Methods | 3D IoU loss [ ] | L | 12.5 | 91.36 | 86.22 | 81.20 | - | - | - | - | - | - |
| | | Fast Point R-CNN [ ] | L | 16.7 | 90.76 | 85.61 | 79.99 | - | - | - | - | - | - |
| | | VoteNet [ ] | L | - | - | - | - | - | - | - | - | - | - |
| | | Feng et al. [ ] | L | - | - | - | - | - | - | - | - | - | - |
| | | Part-A^2 [ ] | L | 12.5 | 91.70 | 87.79 | 84.61 | - | - | - | 81.91 | 68.12 | 61.92 |
| Single Shot Methods | BEV-based Methods | PIXOR [ ] | L | 28.6 | 83.97 | 80.01 | 74.31 | - | - | - | - | - | - |
| | | HDNET [ ] | L | 20.0 | 89.14 | 86.57 | 78.32 | - | - | - | - | - | - |
| | | BirdNet [ ] | L | 9.1 | 76.88 | 51.51 | 50.27 | 20.73 | 15.80 | 14.59 | 36.01 | 23.78 | 21.09 |
| | Point Cloud -based Methods | VeloFCN [ ] | L | 1.0 | 0.02 | 0.14 | 0.21 | - | - | - | - | - | - |
| | | 3D FCN [ ] | L | <0.2 | 70.62 | 61.67 | 55.61 | - | - | - | - | - | - |
| | | Vote3Deep [ ] | L | - | - | - | - | - | - | - | - | - | - |
| | | 3DBN [ ] | L | 7.7 | 89.66 | 83.94 | 76.50 | - | - | - | - | - | - |
| | | VoxelNet [ ] | L | 2.0 | 89.35 | 79.26 | 77.39 | 46.13 | 40.74 | 38.11 | 66.70 | 54.76 | 50.55 |
| | | SECOND [ ] | L | 26.3 | 89.39 | 83.77 | 78.59 | 55.99 | 45.02 | 40.93 | 76.50 | 56.05 | 49.45 |
| | | MVX-Net [ ] | L & I | 16.7 | 92.13 | 86.05 | 78.68 | - | - | - | - | - | - |
| | | PointPillars [ ] | L | 62.0 | 90.07 | 86.56 | 82.81 | 57.60 | 48.64 | 45.78 | 79.90 | 62.73 | 55.58 |
| | Other Methods | LaserNet [ ] | L | 83.3 | 79.19 | 74.52 | 68.45 | - | - | - | - | - | - |
| | | LaserNet++ [ ] | L & I | 26.3 | - | - | - | - | - | - | - | - | - |

TABLE XIII. COMPARISON OF 3D CAMERA DETECTION RESULTS ON THE KITTI TEST BEV DETECTION BENCHMARK, FROM [146]

| Method | Extra | Time | $AP_{BEV}$ (IoU=0.5) | | | $AP_{BEV}$ (IoU=0.7) | | |
|---|---|---|---|---|---|---|---|---|
| | | | Easy | Moderate | Hard | Easy | Moderate | Hard |
| Mono3D [ ] | Mask | 4.2 s | 30.50/ - | 22.39/ - | 19.16/ - | 5.22 / - | 5.19 / - | 4.13 / - |
| 3DOP [ ] | Stereo | 3 s | 55.04/ - | 41.25/ - | 34.55/ - | 12.63/ - | 9.49 / - | 7.59 / - |
| MF3D [ ] | Depth | - | 55.02/54.18 | 36.73/38.06 | 31.27/31.46 | 22.03/19.20 | 13.63/12.17 | 11.60/10.89 |
| Mono3D++ [ ] | Depth+Shape | >0.6s | 46.70/ - | 34.30/ - | 28.10/ - | 16.70/ - | 11.50/ - | 10.10/ - |
| GS3D [ ] | None | 2.3s | - / - | - / - | - / - | - / - | - / - | - / - |
| M3D-RPN [ ] | None | 0.16s | 55.37/55.87 | 42.49/41.36 | 35.29/34.08 | **25.94/26.86** | 21.18/21.15 | 17.90/17.14 |
| Deep3DBox [ ] | None | - | 30.02/ - | 23.77/ - | 18.83/ - | 9.99 / - | 7.71 / - | 5.30 / - |
| MonoGRNet [ ] | None | 0.06s | - / - | - / - | - / - | - / - | - / - | - / - |
| FQNet[ ] | None | 3.33s | 32.57/33.37 | 24.60/26.29 | 21.25/21.57 | 9.50 /10.45 | 8.02 / 8.59 | 7.71 / 7.43 |
| Ours(ResNet18) | None | 0.035s | 52.79/41.91 | 35.92/34.28 | 33.02/28.88 | 20.81/10.84 | 16.60/16.48 | 15.80/15.45 |
| Ours (DLA34) | None | 0.055 s | **57.47/56.90** | **44.16/44.69** | **42.31/41.75** | 25.56/24.74 | **22.12/22.03** | **20.91/18.05** |

TABLE XIV. COMPARISON OF DEPTH COMPLETION RESULTS ON THE KITTI BENCHMARK, FROM [198]

| Method | iRMSE | iMAE | RMSE | MAE | Runtime | Source |
|---|---|---|---|---|---|---|
| HMS-Net_v2 | 3.90 | 1.90 | 911.49 | 310.14 | 0.02 s / GPU | n/a |
| Sparse-to-Dense-2 | 3.21 | 1.35 | 954.36 | 288.64 | 0.04 s / GPU | n/a |
| HMS-Net | 3.25 | 1.27 | 976.22 | 283.76 | 0.02 s / GPU | n/a |
| **Morph-Net** | **3.84** | **1.57** | **1045.45** | **310.49** | **0.17 s / GPU** | **Proposed** |
| IP-Basic | 3.78 | 1.29 | 1288.46 | 302.60 | 0.011 s / 1 core | Ku [[redacted]] |
| ADNN | 59.39 | 3.19 | 1325.37 | 439.48 | 0.04 s / GPU | n/a |
| NN+CNN | 3.25 | 1.29 | 1419.75 | 416.14 | 0.02 s / | Uhrig [[redacted]] |
| SparseConvs | 4.94 | 1.78 | 1601.33 | 481.27 | 0.01 s / | Uhrig [[redacted]] |
| NadarayaW | 6.34 | 1.84 | 1852.60 | 416.77 | 0.05 s / 1 core | Uhrig [[redacted]] |
| SGDU | 7.38 | 2.05 | 2312.57 | 605.47 | 0.2 s / 4 cores | Schneider [[redacted]] |
| NiN CNN | 4.60 | 2.15 | 2378.79 | 685.53 | 0.01 s / | n/a |
| NiN+Mask CNN | 4.63 | 2.40 | 2534.26 | 848.25 | 0.01 s / GPU | n/a |

TABLE XV. COMPARISON OF STEREO MATCHING RESULTS ON KITTI 2015 BENCHMARK, FROM [185]

| Method | Description | Training | | | Avg D1 Non-Occ / Est | | | Avg D1 All / Est | | | Avg D1 Non-Occ / All | | | Avg D1 All / All | | | Time (s) | Environment |
|---|---|---|---|---|---|---|---|---|---|---|---|---|---|---|---|---|---|---|
| | | 3D Sup | Loss | End-to-end | D1-fg | D1-bg | D1-all | D1-fg | D1-bg | D1-all | D1-fg | D1-bg | D1-all | D1-fg | D1-bg | D1-all | | |
| MC-CNN Accr [[redacted]] | Raw disparity + classic refinement | ✓ | $\mathcal{L}_1^{13}$ | × | 7.64 | 2.48 | 3.33 | 8.88 | 2.89 | 3.89 | 7.64 | 2.48 | 3.33 | 8.88 | 2.89 | 3.89 | 67 | Nvidia GTX Titan X (CUDA, Lua/Torch7) |
| Luo *et al.* [[redacted]] | Raw disparity + classic refinement | ✓ | $\mathcal{L}_1^{6}$ | × | 7.44 | 3.32 | 4.00 | 8.58 | 3.73 | 4.54 | 7.44 | 3.32 | 4.00 | 8.58 | 3.73 | 4.54 | 1 | Nvidia GTX Titan X (Torch) |
| Chen *et al.* [[redacted]] | Raw disparity + classic refinement | ✓ | $\mathcal{L}_1^{13}$ | × | – | – | – | – | – | – | – | – | – | – | – | – | – | |
| L-ResMatch [[redacted]] | Raw disparity + confidence score + classic refinement | ✓ | | ○ | 5.74 | 2.35 | 2.91 | 6.95 | 2.72 | 3.42 | 5.74 | 2.35 | 2.91 | 6.95 | 2.72 | 3.42 | 48 | Nvidia Titan-X |
| Han *et al.* [[redacted]] | Matching network | ✓ | $\mathcal{L}_1^{6}$ | × | – | – | – | – | – | – | – | – | – | – | – | – | – | Nvidia GTX Titan Xp |
| Tulyakov *et al.* [[redacted]] | MC-CNN fast [[redacted]] + weakly-supervised learning | | – | × | 9.42 | 3.06 | 4.11 | 10.93 | 3.78 | 4.93 | 9.42 | 3.06 | 4.11 | 10.93 | 3.78 | 4.93 | 1.35 | 1 core 2.5 Ghz + K40 NVIDIA, Lua-Torch |
| FlowNetCorr [[redacted]] | Refined disparity | ✓ | $\mathcal{L}_1^{1}$ | ✓ | – | – | – | – | – | – | – | – | – | – | – | – | 1.12 | Nvidia GTX Titan |
| DispNetCorr [[redacted]] | Raw disparity | ✓ | | ○ | 3.72 | 4.11 | 4.05 | 4.41 | 4.32 | 4.34 | 3.72 | 4.11 | 4.05 | 4.41 | 4.32 | 4.34 | 0.06 | Nvidia Titan-X |
| Pang *et al.* [[redacted]] | Raw disparity | ✓ | $\mathcal{L}_1^{1}$ | ○ | 3.12 | 2.32 | 2.45 | 3.59 | 2.48 | 2.67 | 3.12 | 2.32 | 2.45 | 3.59 | 2.48 | 2.67 | 0.47 | – |
| Yu *et al.* [[redacted]] | Raw disparity map | ✓ | $\mathcal{L}_1^{1}$ | ○ | – | – | – | – | – | – | 5.32 | 2.06 | 2.32 | 5.46 | 2.17 | 2.79 | 1.13 | Nvidia 1080Ti |
| Yang *et al.* [[redacted]] - supp. | Raw disparity | ✓ | $\mathcal{L}_1^{1} + \mathcal{L}_1^{13} + \mathcal{L}_2^{1}$ | ✓ | 3.70 | 1.76 | 2.08 | 4.07 | 1.88 | 2.25 | 3.70 | 1.76 | 2.08 | 4.07 | 1.88 | 2.25 | 0.6 | |
| Yang *et al.* [[redacted]] - unsup. | Raw disparity | × | $\mathcal{L}_1^{9} + \mathcal{L}_1^{13} + \mathcal{L}_2^{1}$ | ✓ | – | – | – | – | – | – | – | – | 7.70 | – | – | 8.79 | 0.6 | |
| Liang *et al.* [[redacted]] | Refined disparity | ✓ | $\mathcal{L}_1^{1}$ | ✓ | – | – | – | – | – | – | 2.76 | 2.07 | 2.19 | 3.40 | 2.25 | 2.44 | 0.12 | Nvidia Titan-X |
| Khamis *et al.* [[redacted]] | Raw disparity + hierarchical refinement | ✓ | $\mathcal{L}_1^{4}$ with $\alpha = 1, c = 2$ | ✓ | – | – | – | – | – | – | – | – | – | 7.45 | 4.30 | 4.83 | 0.015 | Nvidia Titan-X |
| Gidaris & Komodakis [[redacted]] | Refinement only | ✓ | $\mathcal{L}_1^{1}$ | ○ | 4.87 | 2.34 | 2.76 | 6.04 | 2.58 | 3.16 | 4.87 | 2.34 | 2.76 | 6.04 | 2.58 | 3.16 | 0.4 | Nvidia Titan-X |
| Chang & Chen [[redacted]] | Raw disparity | ✓ | $\mathcal{L}_1^{5}$ | ○ | 4.31 | 1.71 | 2.14 | 4.62 | 1.86 | 2.32 | 4.31 | 1.71 | 2.14 | 4.62 | 1.86 | 2.32 | 0.41 | Nvidia GTX Titan Xp |
| Zhong *et al.* [[redacted]] | Raw disparity map | × | $\alpha_1\mathcal{L}_1^{12} + \alpha_2\mathcal{L}_1^{9} + \alpha_3\mathcal{L}_1^{7} + \alpha_4\mathcal{L}_2^{3} + \alpha_5\mathcal{L}_2^{5} + \alpha_6\mathcal{L}_2^{7}$ | ○ | 6.13 | 2.46 | 3.06 | 7.12 | 2.86 | 3.57 | 6.13 | 2.46 | 3.06 | 7.12 | 2.86 | 3.57 | 0.8 | P100 |
| Kendall *et al.* [[redacted]] | Refiend disparity map without refinement module | ✓ | $\mathcal{L}_1^{1}$ | ✓ | 5.58 | 2.02 | 2.61 | 6.16 | 2.21 | 2.87 | 5.58 | 2.02 | 2.61 | 6.16 | 2.21 | 2.87 | 0.9 | Nvidia GTX Titan X |
| Standard SGM-Net [[redacted]] | Refinement with CNN-based SGM | ✓ | Weighted sum of path cost and neighbor cost | ○ | – | – | – | – | – | – | 7.44 | 2.23 | 3.09 | – | – | – | 67 | Nvidia Titan-X |
| Signed SGM-Net [[redacted]] | Refinement with CNN-based SGM | ✓ | Weighted sum of path cost and neighbor cost | ○ | 7.43 | 2.23 | 3.09 | 8.64 | 2.66 | 3.66 | 7.43 | 2.23 | 3.09 | 8.64 | 2.66 | 3.66 | 67 | Nvidia Titan-X |
| Cheng *et al.* [[redacted]] | Refinement | ✓ | $\mathcal{L}_1^{1}$ | ○ | 2.67 | 1.40 | 1.61 | 2.88 | 1.51 | 1.74 | 2.67 | 1.40 | 1.61 | 2.88 | 1.51 | 1.74 | 0.5 | GPU @ 2.5 Ghz (C/C++) |
| EdgeStereo [[redacted]] | Raw disparity | ✓ | $\sum_{sc=1}^{nscales} (\mathcal{L}_1^{1} + \alpha\mathcal{L}_1^{14})_{sc}$ | × | 3.04 | 1.70 | 1.92 | 3.39 | 1.85 | 2.10 | 3.04 | 1.70 | 1.92 | 3.39 | 1.85 | 2.10 | 0.32 | Nvidia GTX Titan Xp |
| Tulyakov *et al.* [[redacted]] | Disparity with sub-pixel accuracy | ✓ | $\mathcal{L}_1^{7}$ | ○ | 3.63 | 2.09 | 2.36 | 4.05 | 2.25 | 2.58 | 3.63 | 2.09 | 2.36 | 4.05 | 2.25 | 2.58 | 0.5 | 1 core @ 2.5 Ghz (Python) |
| Jie *et al.* [[redacted]] | Refined disparity with DL | ✓ | $\mathcal{L}_1^{1}$ | ✓ | 4.19 | 2.23 | 2.55 | 5.42 | 2.55 | 3.03 | 4.19 | 2.23 | 2.55 | 5.42 | 2.55 | 3.03 | 49.2 | Nvidia GTX Titan X |
| Seki *et al.* [[redacted]] | Raw disparity, confidence map, SGM-based refinement | ✓ | $\mathcal{L}_1^{6}$ | ○ | 7.71 | 2.27 | 3.17 | 8.74 | 2.58 | 3.61 | 7.71 | 2.27 | 3.17 | 8.74 | 2.58 | 3.61 | 68 | Nvidia GTX Titan X |
| Kuzmin *et al.* [[redacted]] | Only aggregated cost volume | ✓ | $\mathcal{L}_1^{6}$ | ○ | 10.11 | 4.81 | 5.68 | 11.35 | 5.32 | 6.32 | 10.11 | 4.82 | 5.69 | 11.35 | 5.34 | 6.34 | 0.03 | GPU @ 2.5 Ghz (C/C++) |
| Tonioni *et al.* [[redacted]] | Unsupervised adaptation - DispNetCorr1D [[redacted]] + CENSUS [[redacted]] | ✓ | $\mathcal{L}_1^{2} + \alpha\mathcal{L}_2^{4}$ | NA | – | – | – | – | – | – | – | – | – | – | – | 0.76 | – | GPU @ 2.5 Ghz (Python) |

TABLE XVI. COMPARISON OF SEMI-SUPERVISED AND UNSUPERVISED DEPTH RESULTS ON THE KITTI DATASET, FROM [186]

| Method | Year | Training pattern | Cap | Lower is better | | | | Accuracy: higher is better | | |
|---|---|---|---|---|---|---|---|---|---|---|
| | | | | Abs Rel | Sq Rel | RMSE | RMSE log | $\delta < 1.25^1$ | $\delta < 1.25^2$ | $\delta < 1.25^3$ |
| Garg *et al.* [redacted]L12 Aug8 × cap 50m | 2016 | Semi-sup | 50m | 0.169 | 1.080 | 5.104 | 0.273 | 0.740 | 0.904 | 0.962 |
| Godard *et al.* [redacted] | 2017 | Semi-sup | 80m | 0.148 | 1.344 | 5.927 | 0.247 | 0.862 | 0.960 | 0.964 |
| Kuznietsov *et al.* [redacted] | 2017 | Semi-sup | 80m | 0.113 | 0.741 | 4.621 | 0.189 | 0.803 | 0.922 | 0.986 |
| Poggi *et al.* [redacted] | 2018 | Semi-sup | 80m | 0.126 | 0.961 | 5.205 | 0.220 | 0.835 | 0.941 | 0.974 |
| Ramirez *et al.* [redacted] | 2018 | Semi-sup | 80m | 0.143 | 2.161 | 6.526 | 0.222 | 0.850 | 0.939 | 0.972 |
| Aleotti *et al.* [redacted] | 2018 | Semi-sup | 80m | 0.119 | 1.239 | 5.998 | 0.212 | 0.846 | 0.940 | 0.976 |
| Pilzer *et al.* [redacted] | 2018 | Semi-sup | 80m | 0.152 | 1.388 | 6.016 | 0.247 | 0.789 | 0.918 | 0.965 |
| Luo *et al.* [redacted] | 2018 | Semi-sup | 80m | 0.094 | **0.626** | 4.252 | 0.177 | 0.891 | 0.965 | 0.984 |
| He *et al.* [redacted] | 2018 | Semi-sup | 80m | 0.110 | 1.085 | 5.628 | 0.199 | 0.855 | 0.949 | 0.981 |
| Pilzer *et al.* [redacted] | 2019 | Semi-sup | 80m | 0.098 | 0.831 | 4.656 | 0.202 | 0.882 | 0.948 | 0.973 |
| Tosi *et al.* [redacted] | 2019 | Semi-sup | 80m | 0.111 | 0.867 | 4.714 | 0.199 | 0.864 | 0.954 | 0.979 |
| Chen *et al.* [redacted] | 2019 | Semi-sup | 80m | 0.118 | 0.905 | 5.096 | 0.211 | 0.839 | 0.945 | 0.977 |
| Feng *et al.* [redacted] | 2019 | Semi-sup | 80m | **0.065** | 0.673 | **4.003** | **0.136** | **0.944** | **0.979** | **0.991** |
| Zhou *et al.* [redacted] | 2017 | Unsup | 80m | 0.208 | 1.768 | 6.865 | 0.283 | 0.678 | 0.885 | 0.957 |
| Yang *et al.* [redacted] | 2017 | Unsup | 80m | 0.182 | 1.481 | 6.501 | 0.267 | 0.725 | 0.906 | 0.963 |
| Mahjourian *et al.* [redacted] | 2018 | Unsup | 80m | 0.163 | 1.240 | 6.221 | 0.250 | 0.762 | 0.916 | 0.968 |
| Yin *et al.* [redacted] | 2018 | Unsup | 80m | 0.155 | 1.296 | 5.857 | 0.233 | 0.793 | 0.931 | 0.973 |
| Zou *et al.* [redacted] | 2018 | Unsup | 80m | 0.150 | 1.124 | 5.507 | 0.223 | 0.806 | 0.933 | 0.973 |
| Wang *et al.* [redacted] | 2019 | Unsup | 80m | 0.158 | 1.277 | 5.858 | 0.233 | 0.785 | 0.929 | 0.973 |
| Bian *et al.* [redacted] | 2019 | Unsup | 80m | 0.137 | 1.089 | 5.439 | 0.217 | 0.830 | 0.942 | 0.975 |
| Casser *et al.* [redacted] | 2019 | Unsup | 80m | 0.109 | 0.825 | **4.750** | **0.187** | 0.874 | **0.958** | **0.983** |
| Ranjan *et al.* [redacted] | 2019 | Unsup | 80m | 0.140 | 1.070 | 5.326 | 0.217 | 0.826 | 0.941 | 0.975 |
| Chen *et al.* [redacted] | 2019 | Unsup | 80m | **0.100** | **0.811** | 4.806 | 0.189 | **0.875** | **0.958** | 0.982 |
| Li *et al.* [redacted] | 2019 | Unsup | 80m | 0.150 | 1.127 | 5.564 | 0.229 | 0.823 | 0.936 | 0.974 |
| Almalioglu *et al.* [redacted] | 2019 | Unsup | 80m | 0.150 | 1.141 | 5.448 | 0.216 | 0.808 | 0.939 | 0.975 |

TABLE XVII. COMPARISON OF DEPTH REGRESSION RESULTS ON THREE BENCHMARKS, FROM [185]

| Method | Description | Training | | Runtime | | KITTI 2012 | | | | | | | Make3D | | | | NYUDv2 | | | | | | |
|---|---|---|---|---|---|---|---|---|---|---|---|---|---|---|---|---|---|---|---|---|---|---|---|
| | | #views | 3D sup | #views | time | Performance (↓ the better) | | | | Accuracy (↑ the better) | | | Performance (↓ the better) | | | | Performance (↓ the better) | | | | Accuracy (↑ the better) | | |
| | | | | | | abs rel. | sqr rel. | RMSE (lin) | RMSE (log) | < 1.25 | $< 1.25^2$ | $< 1.25^3$ | abs rel. | sqr rel. | RMSE (lin) | RMSE (log) | abs rel. | sqr rel. | RMSE (lin) | RMSE (log) | < 1.25 | $< 1.25^2$ | $< 1.25^3$ |
| Chakrabarti *et al.* [redacted] | Probability that model confidence and ambiguities | 1 | ✓ | 1 | 24 | – | – | – | – | – | – | – | – | – | – | – | 0.149 | 0.118 | 0.620 | 0.205 | 0.806 | 0.958 | 0.987 |
| Kuznietsov *et al.* [redacted] | Supervised followed by unsupervised | 2 | 3D + Stereo | 1 | – | 0.113 | 0.741 | 4.621 | 0.189 | 0.862 | 0.960 | 0.986 | 0.157 | – | 3.97 | 0.062 | – | – | – | – | – | – | – |
| Zhan *et al.* [redacted] | visual odometry | Stereo seq. + cam. motion | ✗ | 1 | | 0.144 | 1.391 | 5.869 | 0.241 | 0.803 | 0.928 | 0.969 | – | – | – | – | – | – | – | – | – | – | – |
| Eigen [redacted] | Multi-scale | 1 | ✓ | 1 | – | 0.1904 | 1.515 | 7.156 | 0.270 | 0.702 | 0.890 | 0.958 | – | – | – | – | 0.214 | 0.204 | 0.877 | 0.283 | 0.614 | 0.888 | 0.972 |
| Eigen *et al.* [redacted] | VGG - multi-sclae CNN for depth, normals, and labeling | 1 | ✓ | 1 | 0.033 | | | | | – | | | – | – | – | – | 0.158 | 0.121 | 0.641 | 0.214 | 0.769 | 0.950 | 0.988 |
| Fu *et al.* [redacted] | ResNet - Ordinal regression | 1 | ✓ | 1 | – | 0.072 | 0.307 | 2.727 | 0.120 | 0.932 | 0.984 | 0.994 | 0.157 | – | 3.97 | 0.062 | 0.115 | – | 0.509 | 0.051 | – | – | – |
| Gan *et al.* [redacted] | Uses Affinity, Vertical Pooling, and Label Enhancement | 1 | ✓ | 1 | 0.07 | 0.098 | 0.666 | 3.933 | 0.173 | 0.890 | 0.964 | 0.985 | – | – | – | – | 0.158 | – | 0.631 | 0.066 | 0.756 | 0.934 | 0.980 |
| Garg [redacted] | variable-size input | 2 | ✗ | 1 | – | 0.177 | 1.169 | 5.285 | 0.282 | 0.727 | 0.896 | 0.958 | – | – | – | – | – | – | – | – | – | – | – |
| Godard [redacted] | Training with a stereo pair | 2 (cali-brated) | ✗ | 1 | – | 0.114 | 0.898 | 4.935 | 0.206 | 0.830 | 0.936 | 0.970 | 0.535 | 11.990 | 11.513 | 0.156 | – | – | – | – | – | – | – |
| Jiao [redacted] | 40 categories - pay more attention to distant regions | 1 | ✓ | 1 | – | – | – | – | – | – | – | – | – | – | – | – | 0.098 | | 0.329 | 0.125 | 0.917 | 0.983 | 0.996 |
| Laina [redacted] | VGG - feature map up-sampling | 1 | ✓ | 1 | 0.055 | | | | | – | | | 0.176 | – | 4.6 | 0.072 | 0.194 | – | 0.79 | 0.083 | – | – | – |
| Laina [redacted] | ResNet - feature map up-sampling | 1 | ✓ | 1 | 0.055 | | | | | – | | | | – | | | 0.127 | – | 0.573 | 0.055 | 0.811 | 0.953 | 0.988 |
| Lee [redacted] | Split and merge | 1 | ✓ | 1 | – | | | | | – | | | – | – | – | – | 0.139 | 0.096 | 0.572 | 0.193 | 0.815 | 0.963 | 0.991 |
| Li [redacted] | Multiscale patches, refinement with CRF | 1 | ✓ | 1 | – | | | | | – | | | 0.278 | – | 7.188 | – | 0.232 | – | 0.821 | – | 0.6395 | 0.9003 | 0.9741 |
| Qi [redacted] | Joint depth and normal maps | 1 | ✓ | 1 | 0.87 | – | – | – | – | – | – | – | – | – | – | – | 0.128 | | 0.569 | | 0.834 | 0.960 | 0.990 |
| Roy [redacted] | CNN + Random Forests | 1 | ✓ | 1 | – | | | | | – | | | 0.260 | – | 12.40 | 0.119 | 0.187 | – | 0.74 | – | – | – | – |
| Xian [redacted] | Training with 3K ordinal relations per image | 1 | 3K ordinal | 1 | – | | | | | – | | – | – | – | – | – | 0.155 | – | 0.660 | 0.066 | 0.781 | 0.950 | 0.987 |
| Xu [redacted] | Integration with continuous CRF | 1 | ✓ | 1 | | – | – | – | – | – | – | – | 0.184 | – | 4.386 | 0.065 | 0.121 | – | 0.586 | 0.052 | 0.706 | 0.925 | 0.981 |
| Xu [redacted] | Joint depth estimation and scene parsing | 1 | ✓ | 1 | – | | | | | – | | | – | – | – | – | 0.214 | – | 0.792 | – | 0.643 | 0.902 | 0.977 |
| Wang [redacted] | Depth and semantic prediction | 1 | ✓ | 1 | – | | | | | – | | | – | – | – | – | 0.220 | – | 0.745 | 0.262 | 0.605 | 0.890 | 0.970 |
| Zhang *et al.* [redacted] | Hierarchical guidance strategy for depth refinement | 1 | ✓ | 1 | 0.2 | 0.136 | – | 4.310 | – | 0.833 | 0.957 | 0.987 | 0.181 | – | 4.360 | – | 0.134 | – | 0.540 | – | 0.830 | 0.964 | 0.992 |
| Zhang *et al.* [redacted] | ResNet50 - Joint segmentation and depth estimation | 1 | ✓ | 1 | 0.2 | – | – | – | – | – | – | – | 0.156 | – | 0.510 | 0.187 | 0.140 | – | 0.468 | – | 0.815 | 0.962 | 0.992 |
| Zou [redacted] | Joint depth and flow | 2 | ✗ | 2 | – | 0.150 | 1.124 | 5.507 | 0.223 | 0.806 | 0.933 | 0.973 | 0.331 | 2.698– | 0.416 | 6.89 | – | – | – | – | – | – | – |
| Zhou [redacted] | Depth + pose | $\geq 2$ | ✗ | 1 | – | 0.183 | 1.595 | 6.709 | 0.270 | 0.734 | 0.902 | 0.959 | 0.383 | 5.321 | 10.47 | 0.478 | – | – | – | – | – | – | – |
| Zhou [redacted] | 3D-guided cycle consistency | 2 | ✗ | 2 | – | – | – | – | – | – | – | – | – | – | – | – | – | – | – | – | – | – | – |
| Dosovitski [redacted] | Regression from calibrated stereo images | 2 (cali-brated) | ✓ | 2 (cali-brated) | 1.05 | – | – | – | – | – | – | – | – | – | – | – | – | – | – | – | – | – | – |
| [redacted] | Regression from a pair of images | 2 | ✓ | 2 | 0.11 | – | – | – | – | – | – | – | – | – | – | – | – | – | – | – | – | – | – |
| Pang [redacted] | Cascade residual learning | 2 (cali-brated) | ✓ | 2 | – | – | – | – | – | – | – | – | – | – | – | – | – | – | – | – | – | – | – |
| Ilg [redacted] | Extension of FlowNet [redacted] | 2 | ✓ | 2 | – | – | – | – | – | – | – | – | – | – | – | – | – | – | – | – | – | – | – |
| Li [redacted] | Depth from multiscale patches, refinement with CRF | 1 | ✓ | 1 | – | – | – | – | – | – | – | – | 0.278 | – | 7.188 | – | 0.232 | – | 0.821 | – | – | – | – |
| Liu [redacted] | Refinement with continuous CRF | 1 | ✓ | 1 | – | – | – | – | – | – | – | – | 0.314 | – | 0.314 | – | 0.230 | – | 0.824 | – | 0.614 | 0.883 | 0.971 |
| Xie [redacted] | Predict one view from another using estimated depth | 2 stereo | ✗ | 1 | – | – | – | – | – | – | – | – | – | – | – | – | – | – | – | – | – | – | – |
| Chen [redacted] | Training with one ordinal relation per image | 1 | 1 ordinal | 1 | – | – | – | – | – | – | – | – | – | – | – | – | 0.34 | 0.42 | 1.10 | 0.38 | – | – | – |
| Mayer [redacted] | A dataset for training | 1 | ✓ | 0.06 | – | – | – | – | – | – | – | – | – | – | – | – | – | – | – | – | – | – | – |

TABLE XVIII. COMPARISON OF FUSION-BASED 3D OBJECT DETECTION RESULTS ON DIFFERENT BENCHMARKS, FROM [235]

| Reference | Sensors | Obj Type | Sensing Modality Representations and Processing | Network Pipeline | How to generate Region Proposals (RP) [a] | When to fuse | Fusion Operation and Method | Fusion Level[b] | Dataset(s) used |
|---|---|---|---|---|---|---|---|---|---|
| Liang *et al.*, 2019 [redacted] | LiDAR, visual camera | 3D Car, Pedestrian, Cyclist | LiDAR BEV maps, RGB image. Each processed by a ResNet with auxiliary tasks: depth estimation and ground segmentation | Faster R-CNN | Predictions with fused features | Before RP | Addition, continuous fusion layer | Middle | KITTI, self-recorded |
| Wang *et al.*, 2019 [redacted] | LiDAR, visual camera | 3D Car, Pedestrian, Cyclist, Indoor objects | LiDAR voxelized frustum (each frustum processed by the PointNet), RGB image (using a pre-trained detector). | R-CNN | Pre-trained RGB image detector | After RP | Using RP from RGB image detector to build LiDAR frustums | Late | KITTI, SUN-RGBD |
| Dou *et al.*, 2019 [redacted] | LiDAR, visual camera | 3D Car | LiDAR voxel (processed by VoxelNet), RGB image (processed by a FCN to get semantic features) | Two stage detector | Predictions with fused features | Before RP | Feature concatenation | Middle | KITTI |
| Sindagi *et al.*, 2019 [redacted] | LiDAR, visual camera | 3D Car | LiDAR voxel (processed by VoxelNet), RGB image (processed by a pre-trained 2D image detector). | One stage detector | Predictions with fused features | Before RP | Feature concatenation | **Early**, Middle | KITTI |
| Bijelic *et al.*, 2019 [redacted] | LiDAR, visual camera | 2D Car in foggy weather | Lidar front view images (depth, intensity, height), RGB image. Each processed by VGG16 | SSD | Predictions with fused features | Before RP | Feature concatenation | From early to middle layers | Self-recorded datasets focused on foggy weather, simulated foggy images from KITTI |
| Liang *et al.*, 2018 [redacted] | LiDAR, visual camera | 3D Car, Pedestrian, Cyclist | LiDAR BEV maps, RGB image. Each processed by ResNet | One stage detector | Predictions with fused features | Before RP | Addition, continuous fusion layer | Middle | KITTI, self-recorded |
| Du *et al.*, 2018 [redacted] | LiDAR, visual camera | 3D Car | LiDAR voxel (processed by RANSAC and model fitting), RGB image (processed by VGG16 and GoogLeNet) | R-CNN | Pre-trained RGB image detector produces 2D bounding boxes to crop LiDAR points, which are then clustered | Before and at RP | Ensemble: use RGB image detector to regress car dimensions for a model fitting algorithm | Late | KITTI, self-recorded data |
| Kim *et al*, 2018 [redacted] | LiDAR, visual camera | 2D Car | LiDAR front-view depth image, RGB image Each input processed by VGG16 | SSD | SSD with fused features | Before RP | Feature concatenation, Mixture of Experts | Middle | KITTI |
| Yang *et al.*, 2018 [redacted] | LiDAR, HD-map | 3D Car | LiDAR BEV maps, Road mask image from HD map. Inputs processed by PIXOR++ [redacted] with the backbone similar to FPN | One stage detector | Detector predictions | Before RP | Feature concatenation | Early | KITTI, TOR4D Dataset [redacted] |
| Pfeuffer *et al.*, 2018 [redacted] | LiDAR, visual camera | Multiple 2D objects | LiDAR spherical, and front-view sparse depth, dense depth image, RGB image. Each processed by VGG16 | Faster R-CNN | RPN from fused features | Before RP | Feature concatenation | Early, **Middle**, Late | KITTI |
| Shin *et al.*, 2018 [141] | LiDAR, visual camera | 3D Car | LiDAR point clouds, (processed by PointNet [139]); RGB image (processed by a 2D CNN) | R-CNN | A 3D object detector for RGB image | After RP | Using RP from RGB image detector to search LiDAR point clouds | Late | KITTI |
| Chen *et al.*, 2017 [redacted] | LiDAR, visual camera | 3D Car | LiDAR BEV and spherical maps, RGB image. Each processed by a base network built on VGG16 | Faster R-CNN | A RPN from LiDAR BEV map | After RP | average mean, deep fusion | Early, **Middle**, Late | KITTI |
| Asvadi *et al.*, 2017 [redacted] | LiDAR, visual camera | 2D Car | LiDAR front-view dense-depth (DM) and reflectance maps (RM), RGB image. Each processed through a YOLO net | YOLO | YOLO outputs for LiDAR DM and RM maps, and RGB image | After RP | Ensemble: feed engineered features from ensembled bounding boxes to a network to predict scores for NMS | Late | KITTI |
| Oh *et al.*, 2017 [redacted] | LiDAR, visual camera | 2D Car, Pedestrian, Cyclist | LiDAR front-view dense-depth map (for fusion: processed by VGG16), LiDAR voxel (for ROIs: segmentation and region growing), RGB image (for fusion: processed by VGG16; for ROIs: segmentation and grouping) | R-CNN | LiDAR voxel and RGB image separately | After RP | Association matrix using basic belief assignment | Late | KITTI |
| Wang *et al.*, 2017 [redacted] | LiDAR, visual camera | 3D Car, Pedestrian | LiDAR BEV map, RGB image. Each processed by a RetinaNet [redacted] | One stage detector | Fused LiDAR and RGB image features extracted from CNN | Before RP | Sparse mean manipulation | Middle | KITTI |
| Ku *et al.*, 2017 [redacted] | LiDAR, visual camera | 3D Car, Pedestrian, Cyclist | LiDAR BEV map, RGB image. Each processed by VGG16 | Faster R-CNN | Fused LiDAR and RGB image features extracted from CNN | Before and after RP | Average mean | **Early**, Middle, Late | KITTI |
| Xu *et al.*, 2017 [redacted] | LiDAR, visual camera | 3D Car, Pedestrian, Cyclist, Indoor objects | LiDAR points (processed by PointNet), RGB image (processed by ResNet) | R-CNN | Pre-trained RGB image detector | After RP | Feature concatenation for local and global features | Middle | KITTI, SUN-RGBD |
| Qi *et al.*, 2017 [redacted] | LiDAR, visual camera | 3D Car, Pedestrian, Cyclist, Indoor objects | LiDAR points (processed by PointNet), RGB image (using a pre-trained detector) | R-CNN | Pre-trained RGB image detector | After RP | Feature concatenation | Middle, **Late** | KITTI, SUN-RGBD |
| Du *et al.*, 2017 [redacted] | LiDAR, visual camera | 2D Car | LiDAR voxel (processed by RANSAC and model fitting), RGB image (processed by VGG16 and GoogLeNet) | Faster R-CNN | First clustered by LiDAR point clouds, then fine-tuned by a RPN of RGB image | Before RP | Ensemble: feed LiDAR RP to RGB image-based CNN for final prediction | Late | KITTI |
| Matti *et al.*, 2017 [redacted] | LiDAR, visual camera | 2D Pedestrian | LiDAR points (clustering with DBSCAN) and RGB image (processed by ResNet) | R-CNN | Clustered by LiDAR point clouds, then size and ratio corrected on RGB image. | Before and at RP | Ensemble: feed LiDAR RP to RGB image-based CNN for final prediction | Late | KITTI |
| Kim *et al.*, 2016 [redacted] | LiDAR, visual camera | 2D Pedestrian, Cyclist | LiDAR front-view depth image, RGB image. Each processed by Fast R-CNN network [redacted] | Fast R-CNN | Selective search for LiDAR and RGB image separately. | At RP | Ensemble: joint RP are fed to RGB image based CNN. | Late | KITTI |

Table XIX. COMPARISON OF HUMAN PREDICTION METHODS ON BENCHMARKS, FROM REFERENCE [267]

| Metric | Dataset | LSTM | Social LSTM | Social GAN | Social Attention | StarNet (Ours) |
|---|---|---|---|---|---|---|
| ADE | **ZARA-1** | 0.25 | 0.27 | **0.21** | 1.66 | 0.25 |
| | **ZARA-2** | 0.31 | 0.33 | 0.27 | 2.30 | **0.26** |
| | **UNIV** | 0.36 | 0.41 | 0.36 | 2.92 | **0.21** |
| | **ETH** | 0.70 | 0.73 | 0.61 | 2.45 | **0.31** |
| | **HOTEL** | 0.55 | 0.49 | 0.48 | 2.19 | **0.46** |
| **Average ADE** | - | 0.43 | 0.45 | 0.39 | 2.30 | **0.30** |
| **Variance of ADE** | - | 0.028 | 0.026 | 0.021 | 0.166 | **0.008** |
| FDE | **ZARA-1** | 0.53 | 0.56 | **0.42** | 2.64 | 0.47 |
| | **ZARA-2** | 0.65 | 0.70 | 0.54 | 4.75 | **0.53** |
| | **UNIV** | 0.77 | 0.84 | 0.75 | 5.95 | **0.40** |
| | **ETH** | 1.45 | 1.48 | 1.22 | 5.78 | **0.54** |
| | **HOTEL** | 1.17 | 1.01 | 0.95 | 4.94 | **0.91** |
| **Average FDE** | - | 0.91 | 0.91 | 0.78 | 4.81 | **0.57** |
| **Variance of FDE** | - | 0.118 | 0.101 | 0.802 | 1.394 | **0.031** |

Table XX. VEHICLE PREDICTION ERRORS ON KITTI AND SDD DATASETS, FROM REFERENCE [285]

| Method | 1.0 (sec) | 2.0 (sec) | 3.0 (sec) | 4.0 (sec) |
|---|---|---|---|---|
| KITTI (error in meters / miss-rate with 1 $m$ threshold) | | | | |
| *Linear* | 0.89 / 0.31 | 2.07 / 0.49 | 3.67 / 0.59 | 5.62 / 0.64 |
| *RNN ED* | 0.45 / 0.13 | 1.21 / 0.39 | 2.35 / 0.54 | 3.86 / 0.62 |
| *RNN ED-SI* | 0.56 / 0.16 | 1.40 / 0.44 | 2.65 / 0.58 | 4.29 / 0.65 |
| *CVAE 1* | 0.61 / 0.22 | 1.81 / 0.50 | 3.68 / 0.60 | 6.16 / 0.65 |
| *CVAE 10%* | 0.35 / 0.06 | 0.93 / 0.30 | 1.81 / 0.49 | 3.07 / 0.59 |
| *DESIRE-S-IT0 Best* | 0.53 / 0.17 | 1.52 / 0.45 | 3.02 / 0.58 | 4.98 / 0.64 |
| *DESIRE-S-IT0 10%* | 0.32 / 0.05 | 0.84 / 0.26 | 1.67 / 0.43 | 2.82 / 0.54 |
| *DESIRE-S-IT4 Best* | 0.51 / 0.15 | 1.46 / 0.42 | 2.89 / 0.56 | 4.71 / 0.63 |
| *DESIRE-S-IT4 10%* | **0.27** / **0.04** | **0.64** / 0.18 | **1.21** / 0.30 | 2.07 / 0.42 |
| *DESIRE-SI-IT0 Best* | 0.52 / 0.16 | 1.50 / 0.44 | 2.95 / 0.57 | 4.80 / 0.63 |
| *DESIRE-SI-IT0 10%* | 0.33 / 0.06 | 0.86 / 0.25 | 1.66 / 0.42 | 2.72 / 0.53 |
| *DESIRE-SI-IT4 Best* | 0.51 / 0.15 | 1.44 / 0.42 | 2.76 / 0.54 | 4.45 / 0.62 |
| *DESIRE-SI-IT4 10%* | 0.28 / 0.04 | 0.67 / **0.17** | 1.22 / **0.29** | **2.06** / **0.41** |
| SDD (pixel error at $1/5$ resolution) | | | | |
| *Linear* | 2.58 | 5.37 | 8.74 | 12.54 |
| *RNN ED* | 1.53 | 3.74 | 6.47 | 9.54 |
| *RNN ED-SI* | 1.51 | 3.56 | 6.04 | 8.80 |
| *CVAE 1* | 2.51 | 6.01 | 10.28 | 14.82 |
| *CVAE 10%* | 1.84 | 3.93 | 6.47 | 9.65 |
| *DESIRE-S-IT0 Best* | 2.02 | 4.47 | 7.25 | 10.29 |
| *DESIRE-S-IT0 10%* | 1.59 | 3.31 | 5.27 | 7.75 |
| *DESIRE-S-IT4 Best* | 2.11 | 4.69 | 7.58 | 10.66 |
| *DESIRE-S-IT4 10%* | 1.30 | 2.41 | 3.67 | 5.62 |
| *DESIRE-SI-IT0 Best* | 2.00 | 4.41 | 7.18 | 10.23 |
| *DESIRE-SI-IT0 10%* | 1.55 | 3.24 | 5.18 | 7.61 |
| *DESIRE-SI-IT4 Best* | 2.12 | 4.69 | 7.55 | 10.65 |
| *DESIRE-SI-IT4 10%* | **1.29** | **2.35** | **3.47** | **5.33** |